\documentclass[journal]{IEEEtran}

\usepackage[T1]{fontenc}
\usepackage{amsmath,amssymb}
\usepackage{graphicx}
\usepackage{booktabs}
\usepackage{multirow}
\usepackage{url}
\usepackage[hidelinks]{hyperref}
\usepackage{balance}
\usepackage{xcolor}

\graphicspath{{figs/}}

\newcommand{\ncond}{107}
\newcommand{\nitems}{10{,}272}
\newcommand{\nreplies}{61{,}632}
\newcommand{\nlora}{30}
\newcommand{\nloraBackbone}{5}
\newcommand{\nbackbone}{five}

\newcommand{\expKappa}{0.619}

\newcommand{\expESS}{79.5\%}
\newcommand{\expPA}{97.7\%}
\newcommand{\expWPSD}{81.4\%}
\newcommand{\expMajority}{93.6\%}
\newcommand{\expRaterLo}{66.0\%}
\newcommand{\expRaterHi}{97.6\%}

\newcommand{\errN}{1061}
\newcommand{\errJudged}{5184}

\newcommand{\errPole}{4.4\%}
\newcommand{\errWithin}{88.9\%}
\newcommand{\errCross}{6.7\%}
\newcommand{\lenR}{-0.021}
\newcommand{\lenP}{= .610}
\newcommand{\relN}{66}
\newcommand{\relSB}{0.83}
\newcommand{\relSBlo}{0.73}
\newcommand{\relSBhi}{0.94}
\newcommand{\relSE}{4.3\%}

\newcommand{\relHalfR}{0.71}
\newcommand{\relLoraDomR}{+0.61}
\newcommand{\relLoraBankR}{+0.58}
\newcommand{\relLoraExp}{0.68}

\newcommand{\relCtwoExp}{0.31}

\newcommand{\relConeExp}{0.03}

\newcommand{\relBandtwentyfour}{33}
\newcommand{\relBandfortyeight}{19}
\newcommand{\relSDeight}{0.159}
\newcommand{\relSDtwentyfour}{0.083}
\newcommand{\relSDfortyeight}{0.046}

\newcommand{\exCtwoEss}{+1.4}

\newcommand{\exNCtwo}{30}
\newcommand{\exChimEss}{+8.6}
\newcommand{\exChimPa}{+19.9}
\newcommand{\exChimWpsd}{+0.3}

\newcommand{\exChimctrlEss}{+19.8}

\newcommand{\exLoraEss}{+17.8}
\newcommand{\exLoraPa}{+18.1}
\newcommand{\exLoraWpsd}{+16.9}
\newcommand{\exNLora}{30}

\newcommand{\exBaseWpsd}{+9.1}

\newcommand{\exNGeneric}{30}
\newcommand{\exGenericEss}{+5.0}

\newcommand{\exGenericWpsd}{+8.1}
\newcommand{\brPriorMin}{6.2\%}
\newcommand{\brPriorMax}{30.2\%}
\newcommand{\brCtwoEss}{+1.4}

\newcommand{\brPCtwo}{= .009}

\newcommand{\brLoraEss}{+17.8}

\newcommand{\brLoraWpsd}{+16.9}
\newcommand{\brNLora}{30}
\newcommand{\brPosLora}{30}
\newcommand{\brPLora}{< .001}
\newcommand{\brConeEss}{+1.2}

\newcommand{\brPCone}{= .438}

\newcommand{\xcESSgenericG}{+1.58}
\newcommand{\xcESSgenericP}{< .001}

\newcommand{\xcESSpromptG}{+1.92}
\newcommand{\xcESSpromptP}{< .001}

\newcommand{\xcPAchimeraG}{-0.22}
\newcommand{\xcPAchimeraP}{= .310}

\newcommand{\xcWpsdgenericG}{+0.93}
\newcommand{\xcWpsdgenericP}{= .002}
\newcommand{\xcWpsdchimeraG}{+1.39}
\newcommand{\xcWpsdchimeraP}{< .001}

\newcommand{\tiltNeg}{0.492}
\newcommand{\tiltPos}{0.436}
\newcommand{\tiltBase}{0.481}
\newcommand{\tiltChim}{0.731}
\newcommand{\tiltGap}{5.6}
\newcommand{\tiltN}{5}
\newcommand{\tiltSign}{5}
\newcommand{\tiltSignP}{= .031}
\newcommand{\tiltPairedP}{= .007}

\newcommand{\attLoraNeg}{30.0\%}
\newcommand{\attLoraPos}{42.2\%}
\newcommand{\attLoraAsymNeg}{+6.7}

\newcommand{\attLoraAsymPosAbs}{31.1}
\newcommand{\attLoraIntP}{= .002}

\newcommand{\attCtwoIntP}{= .223}

\newcommand{\axCtwoOn}{+0.017}

\newcommand{\axCtwoPl}{< .001}
\newcommand{\axChimOn}{+0.083}

\newcommand{\axChimPl}{= .914}

\newcommand{\axGnegOn}{-0.001}

\newcommand{\axGnegPos}{6/15}
\newcommand{\axGnegP}{= .726}

\newcommand{\axGposOn}{+0.014}

\newcommand{\axGposPl}{< .001}
\newcommand{\axLoraOn}{+0.084}

\newcommand{\axLoraPos}{29/30}
\newcommand{\axLoraP}{< .001}

\newcommand{\mtmmHetero}{-0.04}

\newcommand{\mtmmRawMono}{0.28}
\newcommand{\mtmmRawHetero}{0.22}
\newcommand{\wsCtwo}{0.062}
\newcommand{\wsCtwoNull}{0.056}
\newcommand{\wsCtwoSig}{2/30}
\newcommand{\wsCtwoBinP}{= .446}

\newcommand{\wsLora}{0.183}
\newcommand{\wsLoraNull}{0.084}
\newcommand{\wsLoraSig}{18/30}
\newcommand{\wsLoraBinP}{< .001}

\newcommand{\wsGeneric}{0.065}
\newcommand{\wsGenericNull}{0.055}
\newcommand{\wsGenericSig}{4/30}
\newcommand{\wsGenericBinP}{= .061}

\newcommand{\kraw}{0.815}
\newcommand{\iccItem}{0.44}
\newcommand{\iccNull}{0.43}
\newcommand{\iccNullLo}{0.37}
\newcommand{\iccNullHi}{0.50}
\newcommand{\krBeliefShare}{81\%}
\newcommand{\difCany}{73}
\newcommand{\difKRpol}{0.724}

\newcommand{\difObsClo}{16}
\newcommand{\difObsChi}{32}
\newcommand{\difNullC}{22.2}
\newcommand{\difNullClo}{13}
\newcommand{\difNullChi}{33}

\newcommand{\difNullPct}{51\%}
\newcommand{\difDraws}{1000}
\newcommand{\dtfObsLo}{-0.65}
\newcommand{\dtfObsHi}{-0.34}
\newcommand{\dtfNull}{-0.51}
\newcommand{\dtfNullLo}{-0.69}
\newcommand{\dtfNullHi}{-0.30}
\newcommand{\essBBlo}{28.6\%}
\newcommand{\essBBhi}{39.8\%}

\newcommand{\judgeN}{3}
\newcommand{\judgeLo}{43.8\%}
\newcommand{\judgeHi}{52.8\%}
\newcommand{\judgePanelESS}{57.0\%}
\newcommand{\judgePanelPA}{91.2\%}
\newcommand{\judgePanelWPSD}{62.5\%}
\newcommand{\judgePanelCons}{57.1\%}

\newcommand{\judgeTieRate}{16.8\%}
\newcommand{\jcRecallW}{70\%}

\newcommand{\jcRecallU}{45\%}

\newcommand{\jcRecallI}{71\%}

\newcommand{\jcRecallL}{22\%}

\newcommand{\jcRecallC}{90\%}
\newcommand{\jcBiasC}{+19.7}

\newcommand{\ccR}{.03}
\newcommand{\ccP}{= .923}
\newcommand{\ccFreeR}{.39}
\newcommand{\ccFreeP}{= .149}
\newcommand{\ccN}{15}
\newcommand{\ccNegLo}{72\%}
\newcommand{\ccNegHi}{76\%}
\newcommand{\ccBaseW}{23\%}
\newcommand{\ccBaseU}{27\%}
\newcommand{\ccBaseI}{50\%}
\newcommand{\ccNoIncapI}{26\%}
\newcommand{\ccRestILo}{39\%}
\newcommand{\ccRestIHi}{46\%}
\newcommand{\ovCos}{.14}
\newcommand{\ovCosNinety}{.23}
\newcommand{\ovFive}{1.7\%}
\newcommand{\ovRhoLora}{.33}
\newcommand{\ovRhoBase}{.14}
\newcommand{\ovRhoShift}{.15}
\newcommand{\ovLoraLow}{28.1\%}
\newcommand{\ovLoraHigh}{51.4\%}
\newcommand{\ovBaseLow}{12.5\%}
\newcommand{\ovBaseHigh}{20.8\%}
\newcommand{\ovMcLora}{55}
\newcommand{\ovMcBase}{10}
\newcommand{\ovMcP}{< .001}
\newcommand{\affN}{3}
\newcommand{\affNullPass}{1}
\newcommand{\affAbsW}{25\%}
\newcommand{\affBaseW}{10\%}
\newcommand{\lvaLoraW}{38\%}
\newcommand{\lvaAffW}{25\%}
\newcommand{\lvaDiffW}{12.5}
\newcommand{\lvaLoW}{3.1}
\newcommand{\lvaHiW}{22.2}
\newcommand{\lvaPW}{= .016}

\newcommand{\lvaLoraU}{27\%}
\newcommand{\lvaAffU}{15\%}
\newcommand{\lvaDiffU}{12.5}
\newcommand{\lvaLoU}{4.9}
\newcommand{\lvaHiU}{20.5}
\newcommand{\lvaPU}{= .002}

\newcommand{\affBaseI}{23\%}
\newcommand{\lvaLoraI}{47\%}
\newcommand{\lvaAffI}{20\%}
\newcommand{\lvaDiffI}{26.4}
\newcommand{\lvaLoI}{16.0}
\newcommand{\lvaHiI}{36.8}
\newcommand{\lvaPI}{< .001}
\newcommand{\affDEss}{14.6}
\newcommand{\chimDEss}{17.3}
\newcommand{\affDWpsd}{18.8}
\newcommand{\chimDWpsd}{15.1}
\newcommand{\gnegDEss}{5.6}
\newcommand{\affFreeNegLo}{96\%}
\newcommand{\affFreeNegHi}{97\%}
\newcommand{\affFreeILo}{54\%}
\newcommand{\affFreeIHi}{63\%}
\newcommand{\corpAffI}{57\%}
\newcommand{\corpOwnW}{65\%}
\newcommand{\corpOwnU}{48\%}
\newcommand{\corpOwnI}{92\%}
\newcommand{\freeBaseI}{47\%}
\newcommand{\critN}{61}

\newcommand{\critCtwoTarget}{22.6\%}
\newcommand{\critCtwoPole}{58.5\%}

\newcommand{\critLoraTarget}{61.6\%}
\newcommand{\critLoraPole}{94.4\%}

\newcommand{\critBaseBest}{28.3\%}
\newcommand{\critBaseNeg}{0.579}

\newcommand{\critChimBest}{44.3\%}
\newcommand{\critChimNeg}{0.932}

\newcommand{\critChimctrlTarget}{68.6\%}

\newcommand{\critGnegBest}{28.0\%}
\newcommand{\critGnegNeg}{0.685}

\newcommand{\critGposBest}{33.0\%}
\newcommand{\critGposNeg}{0.511}

\newcommand{\critCtwoModal}{13/30}
\newcommand{\critLoraModal}{26/30}

\newcommand{\critChimModalNeg}{5/5}
\newcommand{\critChimModalWord}{worthless}
\newcommand{\critChimModalWordN}{4/5}

\newcommand{\critGnegModalWord}{worthless}
\newcommand{\critGnegModalWordN}{2/3}

\newcommand{\critGposModalWord}{capable}
\newcommand{\critGposModalWordN}{3/3}
\newcommand{\critFreeLora}{61.6\%}
\newcommand{\critFreePrompt}{22.6\%}
\newcommand{\critFreeG}{+2.30}
\newcommand{\critFreeP}{< .001}
\newcommand{\critFreeUP}{< .001}
\newcommand{\critR}{+0.56}
\newcommand{\critRP}{< .001}
\newcommand{\critRlo}{+0.35}
\newcommand{\critRhi}{+0.71}
\newcommand{\critRho}{+0.67}

\newcommand{\critPaPoleR}{+0.68}
\newcommand{\critPaPoleP}{< .001}
\newcommand{\critWpsdR}{+0.46}
\newcommand{\critWpsdP}{< .001}

\newcommand{\critDecodeR}{+0.99}
\newcommand{\critTieRate}{16.0\%}
\newcommand{\pmLoraSix}{24/30}

\newcommand{\pmLoraPole}{24/30}
\newcommand{\pmLoraPoleP}{< 10^{-6}}
\newcommand{\pmLoraBbLo}{4}
\newcommand{\pmLoraBbHi}{5}
\newcommand{\pmLoraNbb}{5}
\newcommand{\pmLoraOwn}{+0.73}
\newcommand{\pmLoraOther}{-0.13}
\newcommand{\pmLoraPermP}{< .001}

\newcommand{\pmCtwoSix}{7/30}

\newcommand{\pmCtwoPermP}{= .179}

\newcommand{\wmPartialR}{+0.03}
\newcommand{\wmPartialP}{= .804}
\newcommand{\wmLoraR}{+0.10}
\newcommand{\wmLoraP}{= .615}

\newcommand{\wmCtwoR}{-0.18}
\newcommand{\wmCtwoP}{= .332}

\newcommand{\wmAccPartialR}{+0.13}
\newcommand{\wmAccPartialP}{= .320}
\newcommand{\wmAccLoraR}{+0.29}
\newcommand{\wmAccLoraP}{= .123}
\newcommand{\wmAccCtwoR}{-0.18}
\newcommand{\wmAccCtwoP}{= .353}
\newcommand{\wmBeliefRawR}{+0.24}

\newcommand{\wmBeliefAccR}{-0.16}

\newcommand{\critNbb}{5}
\newcommand{\svBPsi}{42.0}
\newcommand{\svBPsiLo}{39.8}
\newcommand{\svBPsiHi}{46.1}

\newcommand{\svBCast}{36.3}

\newcommand{\svBTalk}{44.9}

\newcommand{\svBCone}{50.5}

\newcommand{\svApCone}{53.2}

\newcommand{\svBCtwo}{46.3}

\newcommand{\svApCtwo}{61.6}

\newcommand{\svApSimLo}{39.4}
\newcommand{\svApSimHi}{41.6}
\newcommand{\svConeMarginLo}{5.5}
\newcommand{\svConeMarginHi}{14.2}

\newcommand{\svSigBCarsN}{2}
\newcommand{\svSigBCarsLo}{$-$6.1}
\newcommand{\svSigBCarsHi}{$-$5.3}
\newcommand{\svSigBCastN}{3}
\newcommand{\svSigBCastLo}{$-$6.3}
\newcommand{\svSigBCastHi}{$-$6.2}

\newcommand{\svAccN}{20}
\newcommand{\svAccSigPos}{0}
\newcommand{\svAccSigNeg}{2}
\newcommand{\svAccCiW}{20.6}
\newcommand{\svRawCiW}{10.7}
\newcommand{\svConeMarginAccLo}{7.8}
\newcommand{\svConeMarginAccHi}{23.0}
\newcommand{\svPresMax}{1.7}
\newcommand{\svPaLo}{79}
\newcommand{\svPaHi}{98}
\newcommand{\svLevelGapMax}{1.6}
\newcommand{\svWpsdGapMax}{1.5}
\newcommand{\svWpsdBLo}{41.6}
\newcommand{\svWpsdBHi}{47.7}
\newcommand{\svWpsdApLo}{40.9}
\newcommand{\svWpsdApHi}{44.2}
\newcommand{\eeyMassFt}{95}
\newcommand{\eeyLetterMaxFt}{22.6}
\newcommand{\eeyLetterMinFt}{8.6}
\newcommand{\eeyMassBase}{100}
\newcommand{\eeyLetterMaxBase}{18.6}
\newcommand{\eeyLetterMinBase}{12.1}
\newcommand{\tvAApR}{.44}
\newcommand{\tvAApRho}{.30}

\newcommand{\tvAApN}{67}
\newcommand{\tvABR}{.56}
\newcommand{\tvABRho}{.67}

\newcommand{\tvABN}{61}
\newcommand{\tvApBR}{.04}

\newcommand{\tvApBP}{= .746}

\newcommand{\svStyleMax}{2.5}
\newcommand{\svCheckCast}{14}
\newcommand{\svCheckEey}{11}
\newcommand{\svAPsiLocal}{18.1}
\newcommand{\svBPsiLocal}{27.2}
\newcommand{\svALoraLocal}{31.7}
\newcommand{\svBLoraLocal}{69.5}
\newcommand{\svApSimLocalLo}{16.2}
\newcommand{\svApSimLocalHi}{21.5}
\newcommand{\svBPsiLlama}{41.1}
\newcommand{\svCalApB}{.15}
\newcommand{\svCalApBacc}{.44}
\newcommand{\svCalAB}{.42}
\newcommand{\svCalABacc}{.54}
\newcommand{\svCalApBn}{710}
\newcommand{\svCalABn}{302}
\newcommand{\ftBLo}{9.6}
\newcommand{\ftBHi}{19.6}

\newcommand{\ftBAccLo}{12.3}
\newcommand{\ftBAccHi}{28.5}
\newcommand{\ftBAccSig}{6}
\newcommand{\ftBPaFtLo}{52}
\newcommand{\ftBPaFtHi}{63}
\newcommand{\ftBPaBaseLo}{84}
\newcommand{\ftBPaBaseHi}{96}
\newcommand{\ftApLo}{9.2}
\newcommand{\ftApHi}{17.6}

\newcommand{\ftALo}{1.4}
\newcommand{\ftAHi}{4.9}
\newcommand{\ftASig}{1}

\newcommand{\ftWordsFt}{6}
\newcommand{\ftWordsBase}{68}
\newcommand{\ftUnlabFt}{21}
\newcommand{\ftUnlabBase}{14}

\newcommand{\shortR}{0.94}
\newcommand{\shortRho}{0.91}
\newcommand{\shortDiff}{+0.6}
\newcommand{\shortAbsDiff}{3.7}
\newcommand{\shortSEratio}{1.42}
\newcommand{\shortSBfull}{0.83}
\newcommand{\shortSBhalf}{0.71}
\newcommand{\shortKeptAcc}{85.5\%}
\newcommand{\shortDroppedAcc}{73.6\%}
\newcommand{\shortCompR}{0.73}
\newcommand{\shortCompCorr}{0.96}
\newcommand{\shortRelFortyeight}{0.78}
\newcommand{\shortRelComp}{0.74}
\newcommand{\shortLvlSeLo}{17}
\newcommand{\shortLvlSeHi}{21}
\newcommand{\shortLvlSeFullLo}{11}
\newcommand{\shortLvlSeFullHi}{13}
\newcommand{\tsLora}{+17.7}
\newcommand{\tsLoraP}{< .001}
\newcommand{\tsLoraNullHi}{+5.2}
\newcommand{\tsLoraCells}{30/30}
\newcommand{\tsLoraBbLo}{+12.8}
\newcommand{\tsLoraBbHi}{+22.7}
\newcommand{\tsCtwo}{+0.6}
\newcommand{\tsCtwoP}{= .045}

\newcommand{\tsCtwoCells}{18/30}

\newcommand{\tsCone}{+1.2}
\newcommand{\tsConeP}{= .113}

\newcommand{\mtmmSamePole}{+0.09}
\newcommand{\mtmmDiag}{+0.20}
\newcommand{\mtmmDiff}{+0.11}
\newcommand{\mtmmDiffNullHi}{+0.05}
\newcommand{\mtmmPermP}{< .001}
\newcommand{\mtmmMonoLoraSP}{+0.16}
\newcommand{\mtmmMonoCtwoSP}{+0.11}
\newcommand{\mtmmSPRowCol}{3}
\newcommand{\clLoraNpos}{6.2}
\newcommand{\clLoraGeSix}{22/30}

\newcommand{\clLoraQlow}{+7.7}

\newcommand{\clLoraLvlLo}{+11.4}
\newcommand{\clLoraLvlHi}{+24.7}
\newcommand{\clCtwoNpos}{2.6}

\newcommand{\clCtwoQlow}{-2.2}

\newcommand{\clConeNpos}{2.2}

\newcommand{\catAprimeHelpless}{41.7}
\newcommand{\catAprimeUnlovable}{48.7}
\newcommand{\catAprimeWorthless}{31.1}
\newcommand{\catBHelpless}{45.7}
\newcommand{\catBUnlovable}{30.2}
\newcommand{\catBWorthless}{45.7}
\newcommand{\eeyBoundLo}{-32}
\newcommand{\eeyBoundHi}{+15}

\newcommand{\semSitMean}{.60}
\newcommand{\semSitNinety}{7}
\newcommand{\semSitEighty}{17}
\newcommand{\semSitRef}{.53}
\newcommand{\semOptMean}{.47}
\newcommand{\semOptMax}{.87}
\newcommand{\semOptOther}{.42}
\newcommand{\semOptRef}{.63}
\newcommand{\semGainClose}{+14.3}
\newcommand{\semGainRest}{+18.6}
\newcommand{\semGainP}{= .086}
\newcommand{\semAdvLow}{+19.1}

\newcommand{\semNprompts}{333}
\newcommand{\semNsit}{173}
\newcommand{\semLevels}{L1 36, L2 4, L3 58, L4 30, L5 37, L6 107, L7 37, L8 24}
\newcommand{\semNstems}{10}
\newcommand{\rawCtwoGain}{+5.3}
\newcommand{\rawCtwoGainPrep}{+1.4}
\newcommand{\rawCtwoGainPos}{27/30}
\newcommand{\rawCtwoGainP}{< .001}
\newcommand{\rawCtwoTS}{+3.2}
\newcommand{\rawCtwoTSP}{< .001}
\newcommand{\rawCtwoTSCells}{25/30}
\newcommand{\rawConeGain}{+8.3}

\newcommand{\rawConeGainPos}{6/6}

\newcommand{\rawLoraVsCtwoDiff}{+12.5}
\newcommand{\rawLoraVsCtwoN}{28/30}
\newcommand{\rawLoraVsCtwoP}{< .001}
\newcommand{\rawTiltRaw}{19.2}
\newcommand{\rawTiltPrep}{5.6}
\newcommand{\rawTiltPos}{5/5}

\newcommand{\ess}{\mathrm{ESS}}
\newcommand{\pa}{\mathrm{PA}}
\newcommand{\wpsd}{\mathrm{WPSD}}

\begin{document}

\title{The Schema Coherence Inventory: Measuring Whether an Assigned Belief
Organizes a Language Model's Cognition}

\author{Yuqi~Wu,
Deyang~Li,
Katharine~Hibbard,
Ron~Oswald,
David~Sinclair,
Adegboyega~Sapara,
Kalutota~Kuruwe~Arachchige~Sajana~Indula~Samarasinghe,
Abhinav~Pillai,
Reham~Shalaby,
Riffat~Alam,
Xin-Min~Li,
Yanbo~Zhang,
and~Jie~Chen
\thanks{Y. Wu is with the College of Biomedical Engineering, Fudan University, Shanghai, China and Jiashan Fudan Institute, Fudan University, Jiaxing, China, e-mail: ([ywu\_bh@fudan.edu.cn]).}
\thanks{D. Li, K. Hibbard, R. Oswald, D. Sinclair, A. Sapara, K. K. A. S. I. Samarasinghe, A. Pillai, R. Shalaby, R. Alam, X.-M. Li, and Y. Zhang are with the Department of Psychiatry, University of Alberta, Edmonton, AB, Canada. J. Chen is with the College of Biomedical Engineering, Fudan University, Shanghai, China and Jiashan Fudan Institute, Fudan University, Jiaxing, China.}
\thanks{Corresponding Authors: J. Chen and Y. Zhang ([jc65@ualberta.ca], [yanbo9@ualberta.ca]).}}

\markboth{IEEE Under Review}%
{Author \MakeLowercase{\textit{et al.}}: The Schema Coherence Inventory}

\maketitle

\begin{abstract}
Persona-conditioned language models are used as simulated patients and
affective agents, but there is no validated way to ask whether a belief
assigned to such a model organizes its behaviour or whether the model has
merely adopted a mood. We present the Schema Coherence Inventory (SCI): 96
forced-choice items at eight levels of Beck's cognitive model, each offering
six replies keyed to three negative self-schemas and their positive
counterparts, scored by length-normalised log-likelihood so that nothing is
generated or parsed. Affect alone can move a model onto a default belief, so
the discriminant criterion is target specificity: a
condition built for one belief must select that belief more often than
conditions built for the other beliefs of its pole. Six validation studies over
\ncond{} conditions on \nbackbone{} backbones (1.7B--14B parameters) support
this reading. Nine clinically trained raters recover the keyed belief at
\expESS{} (chance 16.7\%). Belief adapters are target-specific on every
backbone (\tsLora{} points; \tsLoraCells{} belief cells positive), and a
condition's SCI profile identifies its own free generations in \pmLoraSix{}
cases against 5 expected by chance. A 48-item short form correlates
$r = \shortCompR$ with the other 48 items. Applied to six published patient
simulators, none carries its formulation's core belief better than a
one-sentence statement of it. The validation route, which separates a
construct from the affect that accompanies it and checks model judges against
clinicians, offers a template for affective measurement in language models; we
outline its extension across task types, raters, sites and languages. We
release the items, scoring code and short form.
\end{abstract}


\IEEEpeerreviewmaketitle

\section{Introduction}

\IEEEPARstart{A}{language} model asked to play a person with low self-worth will produce bleak text. Whether anything more than bleakness has been installed into the model is a separate question, and it is the question that matters for the applications that motivate this work: simulated patients for clinician training, social simulation and observation via language models, and safety analyses of models that drift into a persistent persona over long interactions \cite{wang2024patientpsi,liu2025eeyore,araujo2026personas}. In all three, what the model is supposed to be carrying is not an affective tone but a belief that organizes cognition: a person who believes they are worthless does not merely feel bad; they read ambiguous events as evidence of not deserving their place, attend to the parts of a scene that confirm it, attribute outcomes accordingly, and reinterpret disconfirming evidence rather than revise the belief \cite{beck1963thinking,beckj2020cbt,clark1999scientific}.

The instruments currently used to evaluate model personas do not distinguish these two situations. The dominant approach administers a human self-report questionnaire to the model and reads the answers as trait scores \cite{serapio2025psychometric,jiang2024personallm}. Self-report inventories inherit assumptions that do not transfer: their items are transparently keyed, so a model that has been told what to be can answer as that thing
without the description touching anything else it does; their response scales confound the strength of an attribute with the valence of the language used to express it; and their factor structure was estimated on human respondents, so a factor recovered from model responses cannot be interpreted with the human solution's warrant. Benchmarks built specifically for therapy-adjacent evaluation \cite{zhang2025cbtbench} assess task competence,
not whether a construct is organizing the model that performs the task.

Affective computing has begun to treat language models both as instruments and as subjects of measurement. They are evaluated on affective tasks \cite{amin2024wide}, probed for emotional reasoning and appraisal \cite{tak2025aware,thejll2026appraisal}, used to annotate emotion and to substitute for human affective ratings \cite{niu2025rethinking,yang2026substitutes}, and applied to personality assessment \cite{zhang2024avi} and to the detection and support of mental illness \cite{hu2026mentalhealth,wu2026mindai}, extending a longer line of learned models of sentiment, conversational emotion and affective physiology \cite{calvo2010affect,lu2024sentiment,lian2024pirnet,li2025eegcontrastive}. A common finding is that what a model's output appears to show cannot be taken at face value: prompt-based affect predictions carry systematic biases \cite{mao2023biases}, and model ratings agree with human ones on some constructs and not others \cite{niu2025rethinking,yang2026substitutes}. These studies ask whether a model can read or express affect. The question here is the converse one for persona-conditioned models: whether a construct assigned to the model organizes its behaviour beyond the affect that comes with it.

The measurement gap is therefore specific. What is missing is an instrument that can separate three things a persona-conditioned model might be doing: expressing generic affect of the right sign; expressing the correct belief at the one level of cognition where it was installed; or expressing the correct belief across levels that were never addressed. Only the third is what clinical simulation requires, and only an instrument designed to fail the first two can report it.

This paper validates such an instrument. The Schema Coherence Inventory (SCI) presents 96 situations spanning eight levels of Beck's cognitive hierarchy, each with six candidate replies, one per belief pole on three signed axes (worthless--worthy, unlovable--lovable, incapable--capable). The six replies to an item are written to be matched on emotional intensity and to differ in what they claim about the self, so that three of the six are available to any respondent with the right valence and only one is available to a respondent
with the right belief. Scoring is by length-normalized log-likelihood, which is deterministic given the model and removes the dependence on a sampled decode or on a judge model \cite{zheng2023judge}.

We establish its psychometric properties, organized as the standards for educational and psychological testing organize them \cite{aera2014standards,cronbach1955construct}:

\begin{itemize}
\item \textbf{Content validity} (Study 1): nine clinically trained raters
      classify all 576 replies blind to the key, which also sets a human
      reference level for model scores.
\item \textbf{Reliability} (Study 2): internal consistency, parallel forms by
      domain and by item bank, and precision as a function of test length.
\item \textbf{Construct and known-groups validity} (Study 3): target
      specificity of belief-conditioned models against generic-affect
      controls, affect adapters and chimeric adapters trained on the same
      sentences repartitioned along non-schema lines, with a
      multitrait--multimethod analysis of belief against delivery method
      \cite{campbell1959mtmm}.
\item \textbf{Criterion validity} (Study 4): whether the SCI identifies the
      belief a three-model judge panel reads out of unconstrained free
      generation.
\item \textbf{Generalizability} (Study 5): item behaviour across \nbackbone{}
      backbones and two model families, with a differential-functioning
      screen \cite{holland1988mh} calibrated against its own null.
\item \textbf{A short form} (Study 6): 48 items selected without reference to
      any model.
\item \textbf{An application} (Study 7): six published patient simulators are
      scored on whether their formulation's core belief organizes their
      replies, including at levels the formulation does not describe.
\end{itemize}

We release the 96-item inventory, the 48-item short form, the scoring toolkit, and the analysis code. The intent is that a future paper reporting a simulated patient or a persona-conditioned model can state that schema coherence was measured with SCI, with the evidence for what that score means reported here. More broadly, the validation route itself, rather than only the instrument, is meant to be reusable: Section~\ref{sec:discussion} discusses how it applies to other affective and psychological constructs measured in language models, and how SCI is to be extended across task types, rater panels, clinical sites and languages.
\section{The Instrument}
\label{sec:instrument}

\begin{figure*}[!t]
\centering
\includegraphics[width=0.8\textwidth]{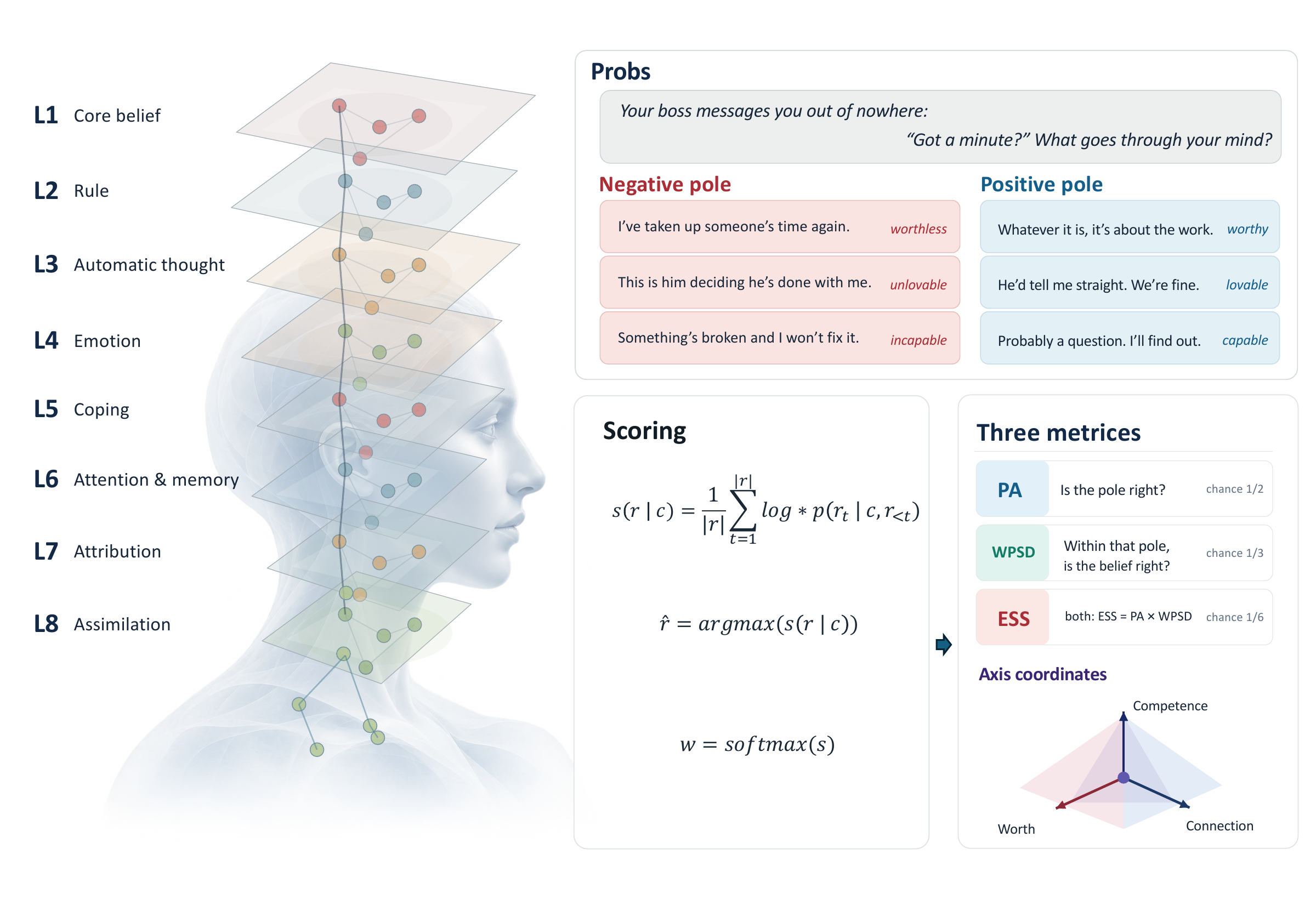}
\caption{The instrument and its scoring. Each item presents a situation and a
question; six candidate replies enact the six beliefs at that level of the
formulation. The respondent model never sees the replies as options and never
generates: each reply is scored by its length-normalised log-likelihood under
the model's own conditioning, and the argmax is the selection. Pole accuracy,
within-pole schema discrimination and their product decompose the score, and
the softmax over the six scores gives signed coordinates on the three axes.
A belief is called organizing when it is selected across the eight levels
below, not only at the level where it is stated. Reply texts are abbreviated
for display; Supplementary Section~S5 gives every item verbatim.}
\label{fig:schematic}
\end{figure*}

\subsection{Construct}

Schema coherence is the degree to which a single belief about the self
organizes a respondent's cognition at every level at which cognition can be
observed. The construct is taken from the cognitive model, in which a core
belief conditions intermediate rules, which condition automatic thoughts,
which condition emotion, coping, what is attended to and retained, how causes
are assigned, and what happens to evidence that contradicts the belief
\cite{beck1963thinking,beckj2020cbt,padesky1994schema,james2004schemas}. A
respondent in whom the belief is merely a label answers one level in its
terms; a respondent in whom it is a schema answers all of them in its terms.

We instantiate six belief poles on three signed axes --- worth
(worthless--worthy), connection (unlovable--lovable) and competence
(incapable--capable). The three-axis structure is the one that recurs across
formulations of the cognitive model. It is a design commitment of the
instrument, not a factor-analytic result.

Two properties of the construct drive the item design. First, belief identity
is not valence: worthless, unlovable and incapable are all negative, and an
instrument that cannot tell them apart is measuring mood. Second, coherence is
cross-level: a score that is high at the level a persona description
addresses and at chance elsewhere describes an instruction that has been
followed, not a schema that has been installed.

\subsection{Blueprint}

SCI v2.1 contains 96 items in eight blocks of 12, one block per level
(Fig.~\ref{fig:schematic}; Supplementary Table~S11). Levels L1--L6 form the core module; L7 is an
attribution module in which six items describe positive and six negative
events, so that attributional asymmetry is estimable within the level; L8 is
an assimilation module in which every situation states a fact that
disconfirms the negative beliefs, so that the level measures belief
maintenance under disconfirming evidence. Each block is balanced across the
achievement and interpersonal domains, six items each.

Every item presents a situation, a question, and six candidate replies keyed
A--F to the six beliefs (see Fig.~\ref{fig:schematic}). The letter-to-belief mapping is permuted across
items, so no letter is systematically associated with a belief. The six
replies are written to the same length and emotional intensity and to differ
in what they assert about the self, which is what makes the within-valence
contrast in Section~\ref{sec:scoring} a discrimination and not a tone match.
Supplementary Section~S5 lists all 96 items verbatim.

Three items whose situations appeared verbatim in the belief-adapter training
corpus were replaced in v2.1 with items of the same level, domain and
letter mapping, so the design is unchanged and no scored item is a memorized
string. Items from the original bank (v1, $n=35$), the extended bank (v2,
$n=58$) and the three replacements (v2r) are distinguished in the release, and
Section~\ref{sec:study2} uses the bank split as a parallel-forms contrast.
\section{Scoring}
\label{sec:scoring}

\subsection{Response model}

A respondent model $M$ is presented with the item's situation and question in
a fixed neutral prompt and is never shown the options. For each candidate
reply $r$ we compute the length-normalized conditional log-likelihood
\begin{equation}
s(r \mid c) \;=\; \frac{1}{|r|}\sum_{t=1}^{|r|}
\log p_M\!\left(r_t \mid c,\, r_{<t}\right),
\label{eq:score}
\end{equation}
where $c$ is the item context and $|r|$ the number of reply tokens. The
selected reply is $\hat{r} = \arg\max_r s(r\mid c)$, and the option weights
are $w = \mathrm{softmax}\big(s(r_1\mid c),\dots,s(r_6\mid c)\big)$.

This is a deliberate departure from the two standard alternatives. Presenting
the six options and parsing a letter makes the score depend on
instruction-following ability and on output format; the inventory's own
metadata records that a belief-conditioned model's verbal style
(``\textit{It's just\ldots\ C.}'') produces belief-dependent parse failures,
which is measurement error correlated with the thing being measured. Sampling
a free response and having a judge model label it adds the judge's own biases
\cite{zheng2023judge} and its variance. Equation~(\ref{eq:score}) is
deterministic given the model, requires no decoding, and evaluates all six
options under identical context. Free generation and judging are used in
Section~\ref{sec:study4}, but as an external criterion for the likelihood
score rather than as the score itself.

\subsection{Metrics}

Let $b^\ast$ be the belief under test and $\hat{r}_i$ the selection on item
$i$ of $n$. Write $\mathrm{pole}(\cdot)$ for the sign of a belief and
$\mathcal{P}$ for the three replies sharing $b^\ast$'s pole.

\emph{Exact schema selection} is the headline score,
\begin{equation}
\ess = \frac{1}{n}\sum_{i=1}^{n}
\mathbf{1}\!\left[\hat{r}_i = b^\ast\right], \qquad \text{chance } 1/6 .
\end{equation}

\emph{Pole accuracy} is the valence component,
\begin{equation}
\pa = \frac{1}{n}\sum_{i=1}^{n}
\mathbf{1}\!\left[\mathrm{pole}(\hat{r}_i) = \mathrm{pole}(b^\ast)\right],
\qquad \text{chance } 1/2 ,
\end{equation}
and \emph{within-pole schema discrimination} is the identity component,
conditioned on the pole being right,
\begin{equation}
\wpsd = \frac{\sum_i \mathbf{1}[\hat{r}_i = b^\ast]}
             {\sum_i \mathbf{1}[\hat{r}_i \in \mathcal{P}]},
\qquad \text{chance } 1/3 .
\end{equation}
The three are related exactly, $\ess = \pa \cdot \wpsd$. The decomposition
separates the pole from the choice within it, but a raised $\wpsd$ is not by
itself evidence of a belief: a manipulation that installs affect alone can
settle on one default belief within the pole and raise $\wpsd$ for that
belief (Section~\ref{sec:study3}). The discriminant criterion is therefore
\emph{target specificity}, a condition $\times$ belief interaction: among
conditions on the same backbone, the one built for belief $b$ must select $b$
more often than those built for the other beliefs of $b$'s pole. Comparing
conditions on the same backbone removes the backbone's prior on $b$;
comparing within the pole removes valence. The target-specificity index
($\mathrm{TS}$) is that difference in selection rate, averaged over the six
beliefs; its null permutes the target labels within pole and backbone.

Finally, the softmax weights project onto the three signed axes. With
$a \in \{\text{worth},\text{conn},\text{comp}\}$ and $\sigma(b) = \pm 1$ the
sign of belief $b$ on its own axis, the coordinate is
$z_a = \sum_{b \in a} \sigma(b)\, w(b)$. The \emph{on-axis displacement} of a
condition is $\sigma(b^\ast)$ times the change in $z_{a^\ast}$ relative to the
untreated model of the same backbone, averaged over items; it is positive when
the condition moves the model toward its target on the target's axis.

\subsection{Nulls}

$\ess$ is compared against $1/6$ only when the condition has a declared
target. A condition with no target --- an untreated model, or a generic-affect
control --- must be scored against the belief that fits it best, which is a
maximum over six correlated statistics and sits well above $1/6$ under the
null. For those conditions we report a Monte Carlo best-of-six null in which
item labels are drawn uniformly and the same maximum is taken, so the comparison is
between the observed best fit and the best fit obtainable by chance. PA and
WPSD of an untargeted condition are read at its ESS-best belief, and in each
null draw at that draw's ESS-best belief. Reporting an untargeted condition
against $1/6$ would manufacture significance.
\section{Validation Design}
\label{sec:design}

\subsection{Respondent conditions}
\label{sec:conditions}

An instrument is validated against respondents whose standing on the construct
is known by design. We assembled \ncond{} conditions from \nbackbone{} backbones and eight
condition families (Table~\ref{tab:conditions}). Backbones span 1.7B to 14B
parameters and two model families \cite{grattafiori2024llama3,yang2025qwen3}; the
main backbone, \texttt{merged\_v2}, is Llama-3.1-8B-Instruct fine-tuned on
AnnoMI motivational-interviewing transcripts with client turns as the model's
turns, then merged, so that it speaks in the first person before any belief is
installed and the belief manipulation is not confounded with the shift out of
assistant voice. The other four backbones were given first-person register in the same way
before any adapter was trained.

\emph{Belief adapters} (LoRA, $n=\nlora{}$) are rank-16 LoRA adapters
\cite{hu2022lora} ($\alpha=32$, attention projections), one per belief on each
of \nloraBackbone{} backbones. Each is trained on 343 replies to the same
situation--question pairs, one reply per pair written for that belief by
GPT-4o, with 10\% held out for early stopping (learning rate $10^{-4}$,
effective batch 8, at most 12 epochs on the main backbone and 8 elsewhere,
seed 42, one seed per adapter). The objective is next-token loss plus
$0.3\,L_{c}$, where
$L_{c}=\max\bigl(0,\ \bar\ell(y^{-}) - \bar\ell(y^{+}) + 0.5\bigr)$,
$\bar\ell$ is the length-normalized log-likelihood of a reply, $y^{+}$ is the
trained belief's reply and $y^{-}$ is the reply of another belief of the same
pole to the same situation. They are the condition in which the construct is
expected to be present.

\emph{Prompted beliefs} come in two strengths: a one-line statement of the
belief (C1, $n=6$, main backbone only) and an engineered multi-sentence
description that names the belief's behavioural signature (C2, $n=30$). These
are the delivery method the field currently uses
\cite{shanahan2023roleplay,tseng2024persona}, and they provide the second
method needed for a multitrait--multimethod analysis.

\emph{Generic affect controls} (GNEG, GPOS, $n=15$ each) are the discriminant
test. Each names an emotional register and explicitly declines to give a
reason for it: ``\textit{You are in a bleak, discouraged mood today. Nothing
feels promising. Answer in that register. Do not give a reason for feeling
this way.}'' Three wordings per valence guard against a null that is really a
single unlucky phrasing. These prompts were written for this study and were
never used in item development or in training or selecting any adapter.

\emph{Chimera adapters} (CHIM, $n=5$) are trained through the same pipeline,
contrastive term included, on 342 replies from the three negative beliefs'
corpora, with the source belief of each situation assigned along a non-schema
line (domain, probe, severity, congruence, or at random); no belief supplies
more than 52\% of a corpus. They keep the corpus's style and valence and remove
any single construct, testing whether the instrument credits a belief to
anything trained on belief-like text. A positive control (CHIM\_CTRL) retrains
\emph{worthless} through the same pipeline.

\emph{Affect adapters} (AFF, $n=\affN$, main backbone, three seeds) are the
adapter-matched version of the generic affect control: trained through the same
pipeline on the same situations, with replies (generated by Qwen3-14B, 333
after removing any self-evaluative stem) that express negative affect about the
circumstances and nothing about the speaker's worth, others' regard or
ability, and the situation's neutral reply as the contrastive negative. They
test whether an adapter that installs negative affect, and nothing else, is
credited with a belief.

\emph{Untreated backbones} (base, $n=5$) receive the shared prompt only.

\begin{table}[!t]
\caption{The \ncond{} respondent conditions. Expected standing on the construct is
what the condition is designed to have, not what it was observed to have.}
\label{tab:conditions}
\centering
\footnotesize
\setlength{\tabcolsep}{4pt}
\begin{tabular}{@{}llcl@{}}
\toprule
Family & Manipulation & $n$ & Expected \\
\midrule
LoRA       & belief adapter, 6 beliefs      & \nlora{} & belief present \\
CHIM\_CTRL & retrained \textit{worthless}   & 1  & belief present \\
C2         & engineered belief prompt       & 30 & prompted \\
C1         & one-line belief prompt         & 6  & prompted \\
GNEG       & negative affect, no belief     & 15 & valence only \\
GPOS       & positive affect, no belief     & 15 & valence only \\
CHIM       & style-matched, construct removed & 5 & no belief \\
base       & shared prompt only             & 5  & no belief \\
\midrule
           & \multicolumn{1}{r}{Total}      & \ncond{} & \\
\midrule
AFF        & affect adapter, no belief$^\dagger$ & \affN{} & valence only \\
\bottomrule
\multicolumn{4}{@{}l@{}}{\scriptsize $^\dagger$Main backbone, three seeds; a Study~3 control, not in the pooled analyses.}
\end{tabular}
\end{table}

\subsection{Procedure}

Every condition answers all 96 items under the same neutral first-person
prompt; belief information enters only through the adapter or the appended
persona text, never through the item. Scoring follows
Equation~(\ref{eq:score}) in \texttt{float32} with an item's six replies always
in one batch. This yields $\ncond \times 96 = \nitems$ scored items and $\nreplies$
scored replies.

\subsection{Analysis}

Confidence intervals are bootstrap percentile intervals over items
($10{,}000$ resamples), which is the resampling unit that matches the
generalization being claimed. Permutation tests draw item labels uniformly;
for untargeted conditions the statistic is the maximum over the six beliefs,
as described in Section~\ref{sec:scoring}. Where several tests address one
family of hypotheses we report Holm-corrected $p$-values.
\section{Study 1: Content Validity and the Human Reference Level}
\label{sec:study1}

\subsection{Method}

Nine raters with clinical training classified all 576 candidate replies into
the six beliefs. Raters saw the situation and the reply, in randomized order,
and were blind to the key and to the item's level. The task is the one the
instrument asks of a respondent, minus the persona: given a reply, which
belief does it express?

Two questions are answered by the same data. Agreement among raters
establishes that the six categories are recoverable from the replies at all.
Agreement between raters and the key establishes that the replies encode the
belief the item writer intended, which is what content validity means here.
The rater-to-key accuracy also defines the human reference level for the
paper. Raters judge one reply at a time, whereas a model compares six, so the
reference level is not an upper bound on model scores.

\subsection{Results}

\begin{table}[!t]
\caption{Study 1. What nine clinically trained raters recover from the 576
candidate replies, blind to the key. Intervals are bootstrap percentile
intervals over items. The eight-rater column omits the least accurate rater and
is reported as a sensitivity analysis, not as the primary result.}
\label{tab:ceiling}
\centering
\footnotesize
\setlength{\tabcolsep}{4pt}
\begin{tabular}{@{}lccc@{}}
\toprule
 & Chance & Nine raters & Eight raters \\
\midrule
Fleiss' $\kappa$ & --- & 0.619 [0.589, 0.650] & 0.643 [0.611, 0.674] \\
Krippendorff's $\alpha$ & --- & 0.619 [0.589, 0.650] & 0.643 [0.610, 0.675] \\
\midrule
ESS (\%) & 16.7 & 79.5 [77.4, 81.7] & 81.2 [79.2, 83.3] \\
PA (\%) & 50.0 & 97.7 [97.0, 98.3] & 97.9 [97.2, 98.4] \\
WPSD (\%) & 33.3 & 81.4 [79.4, 83.3] & 83.0 [81.0, 84.9] \\
Majority vs.\ key (\%) & 16.7 & 93.6 & 90.6 \\
\bottomrule
\end{tabular}
\end{table}

Raters agree substantially and well above chance
(Table~\ref{tab:ceiling}). Fleiss' $\kappa = \expKappa$ across nine raters
\cite{fleiss1971kappa}; Krippendorff's $\alpha$ is numerically identical to
three decimals, as expected when the data are complete and nominal
\cite{hayes2007alpha}. Accuracy against the key is $\expESS$ exact, against a
$16.7\%$ chance rate. Decomposed, raters recover the pole of a reply almost
perfectly ($\expPA$, chance $50\%$) and the belief within the pole at $\expWPSD$
(chance $33.3\%$). Taking the majority vote of the nine as a single consensus
rater raises exact accuracy to $\expMajority$.

Per-rater accuracy ranges from $\expRaterLo$ to $\expRaterHi$. We report all
nine raters as primary, because excluding a rater on the basis of their
agreement is a decision made after seeing the outcome, and the eight-rater
column, which omits the least accurate rater, as a sensitivity analysis.

\begin{figure*}[!t]
\centering
\includegraphics[width=\textwidth]{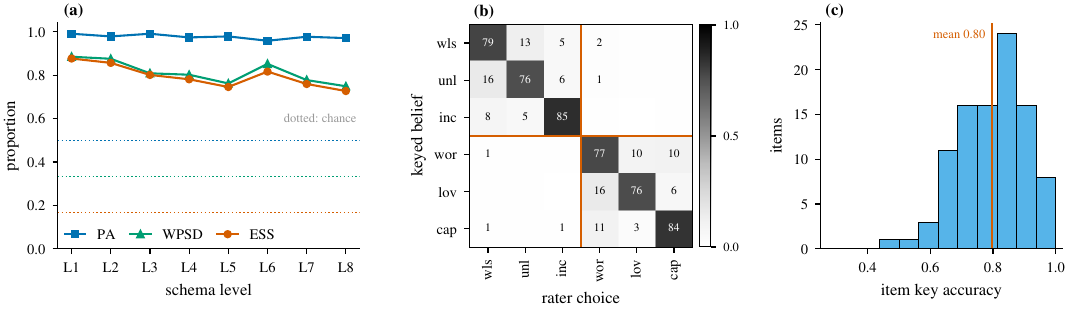}
\caption{Study 1. (a) Rater accuracy against the key by level, nine raters;
dotted lines mark chance for each metric. (b) Row-normalized confusion matrix,
per cent; the orange rules separate negative from positive beliefs. (c)
Distribution of per-item key accuracy across the 96 items.}
\label{fig:expert}
\end{figure*}

Agreement holds across the instrument (Fig.~\ref{fig:expert}a). Exact
accuracy is highest at L1 ($87.7\%$), where the belief is stated, and lowest at
L8 ($72.7\%$). It falls as the reply moves further from a direct statement of
the belief, though not monotonically: L6, where the belief is expressed as
what a person notices and remembers, is read more accurately ($81.6\%$) than
L4 and L5. Even the hardest level is more than four times chance. Accuracy by
belief ranges from $76.0\%$ (\emph{unlovable}) to $85.0\%$
(\emph{incapable}), and the achievement domain is read more accurately than
the interpersonal domain ($83.3\%$ vs.\ $75.8\%$).

\subsection{The geometry of the errors}

What raters get wrong is as informative as what they get right, because the
three kinds of error carry different verdicts on the instrument
(Fig.~\ref{fig:expert}b). Of $\errN$ errors in $\errJudged$ judgements,
$\errPole$ are pole reversals, in which the rater read a negative reply as its
own positive opposite or the reverse; $\errWithin$ are same-valence, other-axis
confusions, such as \emph{worthless} read as \emph{unlovable}; and $\errCross$ are
unstructured, wrong on both valence and axis.

The valence signal is therefore close to unambiguous, and almost all residual
disagreement is the within-valence discrimination the design is built to
demand; the largest confusion is between worth and connection, the pair
clinical formulations also treat as hardest to separate. A model at chance on
$\wpsd$ thus fails the discrimination that costs trained raters most of their
errors, whereas a model that fails $\pa$ fails something raters almost never
fail. Rater accuracy does not depend on reply length ($r = \lenR$, $p \lenP$,
$n=576$).

\subsection{Item screening}

Per-item key accuracy averages $0.80$ (Fig.~\ref{fig:expert}c). One item
(M1-06, L8; L8 items carry the prefix M1 in the released bank) falls below chance-adjusted usefulness at $0.44$, and the five
weakest items are concentrated in L8 and L4 --- the assimilation module, where
a reply must be read against a disconfirming fact, and the emotion level,
where the six beliefs differ least in affective content. We retain all 96
items in the full instrument, report the flagged item, and use per-item
accuracy as the first selection criterion for the short form in
Section~\ref{sec:study6}.
\section{Study 2: Reliability and Precision}
\label{sec:study2}

\subsection{Method}

Scoring is deterministic: the same model, item and numerical precision return
the same selection, so test--retest reliability is not informative here, and
reporting a coefficient of $1.0$ would be a category error. What can vary is
which items are on the form. For every scored condition we therefore report
split-half consistency (the six-belief weight profiles of two level-stratified
random halves, correlated, averaged over 1{,}000 splits and stepped up by the
Spearman--Brown formula), the bootstrap standard error of ESS over items
($10{,}000$ resamples), and two parallel-forms comparisons that split the bank
on design rather than at random: the 48 achievement items against the 48
interpersonal items, and the two independently written reply banks. Test
length was varied from 8 to 96 items by level-stratified random draws on the
main backbone's six belief adapters. Full definitions are in Supplementary
Section~S7.

\subsection{Results}

Across the \relN{} conditions with a declared target (LoRA, C2, C1; CHIM\_CTRL
excluded) the median split-half correlation is \relHalfR{}, \relSB{} after
stepping up (IQR \relSBlo--\relSBhi), and the median bootstrap
standard error of ESS is \relSE{} on a scale whose chance value is $16.7\%$
(Supplementary Table~S7). Two conditions are therefore separable at $p<.05$
from a single administration each when their ESS differs by about
$1.96\sqrt{2}$ standard errors, or 12 points --- the resolution the family
contrasts in Study 3 turn out to need, since the belief adapters exceed the
generic-affect controls by \exLoraEss{} against \exGenericEss{}~pp over their
respective nulls. Parallel forms are read within a family: across the 30 belief
adapters the two content domains agree at $r = \relLoraDomR{}$ and the two
reply banks at $r = \relLoraBankR{}$, close to the \relLoraExp{} that a 48-item
form can reach given how much the adapters differ from one another; the
prompted conditions differ too little in true standing for their
parallel-forms correlations to carry information.

The length curve is the practical result (Supplementary Fig.~S1). The
standard deviation of ESS across random draws falls from \relSDeight{} at 8
items to \relSDtwentyfour{} at 24 and \relSDfortyeight{} at 48, and is zero by
construction at 96. A $\pm 2$~SD band around a score is therefore about
\relBandtwentyfour{}~points wide at 24 items and \relBandfortyeight{} at 48.
This is the empirical basis for the short form in Section~\ref{sec:study6},
and the reason we do not recommend anything shorter.
\section{Study 3: Construct Validity --- Separating a Belief from its Valence}
\label{sec:study3}

\subsection{Method}

The threat to any measure of an installed belief is that it is measuring mood.
A model told it is worthless will produce negative replies, and an instrument
whose options differ in valence will score that as success without ever
testing whether the specific belief is present. Study 3 is designed so that
this confound has somewhere to show itself.

Three comparisons carry the argument. The \emph{discriminant} comparison sets
the belief conditions against conditions given an affective state and no
belief content, which should reach the negative pole and fail to select among
the three negative beliefs. The \emph{convergent} comparison sets adapters
against engineered prompts as two methods of installing the same six traits
\cite{campbell1959mtmm}. The \emph{null} comparison scores every untargeted
condition against the maximum over the six candidate beliefs
(Section~\ref{sec:scoring}).

The primary test is target specificity (Section~\ref{sec:scoring}): on each
backbone, the selection rate of belief $b$ under the condition built for $b$,
minus its mean under the conditions built for the other two beliefs of $b$'s
pole. It is computed for every family that has one condition per belief on a
backbone (belief adapters and engineered prompts on all backbones, one-line
prompts on the main backbone), with an exact within-pole relabelling null per
backbone ($3! \times 3! = 36$ assignments) and a sampled joint null over
backbones. PA and WPSD are reported alongside it: a raised WPSD shows that a
condition prefers one belief within the pole, but only target specificity
shows that the preferred belief is the one installed.

Family means are not comparable as point estimates, because an untargeted
condition is scored at the best of six beliefs and maximisation lifts it on all
three metrics before any model behaviour is involved. We therefore report each
condition's \emph{excess} over its own null: the observed value minus chance
for a belief condition, and minus the mean of the best-of-six null (PA and WPSD
read at the ESS-best belief in the data and in every null draw alike) for an
untargeted one, which puts all \nloraBackbone{} backbones on a common scale. The belief
adapters are contrasted with generic affect, chimera adapters and engineered
prompts on each of ESS, PA and WPSD (Welch $t$ on excess), and the nine
$p$-values are Holm-corrected.

For untargeted conditions we add a \emph{within-pole} null: each item keeps
the pole of the reply the condition selected, and only the label within that
pole is redrawn. PA is then fixed at its observed value, and the null asks
whether a condition singles out one belief within its pole more often than a
process that sets only valence would.

\subsection{Results}

\begin{table*}[!t]
\caption{Study 3. Every condition family on the main backbone, with the
permutation test of ESS. Conditions built to carry a named belief are scored
and tested against that belief; conditions with no named belief (generic
affect, untreated, chimera) are scored at their ESS-best belief, with PA and
WPSD read at that belief, and tested against the best-of-six null.
Chance is 16.7\% for ESS, 50.0\% for PA and 33.3\% for WPSD.}
\label{tab:families}
\centering
\small
\begin{tabular}{@{}lrrrrrr@{}}
\toprule
Family & $n$ & ESS (\%) & PA (\%) & WPSD (\%) & median $p$ & $p<.05$ \\
\midrule
belief adapter & 6 & 39.8 & 74.0 & 53.7 & $<$.001 & 6/6 \\
chimera control & 1 & 36.5 & 74.0 & 49.3 & $<$.001 & 1/1 \\
\midrule
engineered prompt & 6 & 18.1 & 53.8 & 33.5 & .442 & 0/6 \\
one-line prompt & 6 & 17.9 & 51.2 & 35.2 & .438 & 0/6 \\
chimera adapter & 5 & 30.6 & 73.1 & 41.9 & .004 & 5/5 \\
generic negative affect & 3 & 24.0 & 52.8 & 45.4 & .386 & 0/3 \\
generic positive affect & 3 & 24.7 & 55.2 & 44.7 & .236 & 0/3 \\
untreated & 1 & 26.0 & 54.2 & 48.1 & .076 & 0/1 \\
\bottomrule
\end{tabular}
\end{table*}

Table~\ref{tab:families} gives the main backbone, where every family is
present; Supplementary Fig.~S4 places every condition on all backbones in the
PA--WPSD plane. The belief adapters
raise both components, and every one of the six on the main backbone is
above chance. No control family clears its best-of-six null, with one
exception treated below.

Target specificity is the primary result. On every backbone, each belief
adapter selects its own belief more often than the adapters built for the
other beliefs of its pole: $\mathrm{TS} = \tsLora{}$ points (per backbone
\tsLoraBbLo{} to \tsLoraBbHi{}), positive in \tsLoraCells{} backbone $\times$
belief cells, against a joint null whose 95th percentile is \tsLoraNullHi{}
($p \tsLoraP$; on each backbone the observed assignment is the most specific
of the 36). Engineered prompts are target-specific only marginally
($\mathrm{TS} = \tsCtwo{}$, $p \tsCtwoP$; \tsCtwoCells{} cells positive), and
one-line prompts not detectably ($\tsCone{}$, $p \tsConeP$).

Pooling all \nloraBackbone{} backbones and referencing each condition to its
own null sharpens this considerably. The belief adapters gain
\exLoraEss{}~pp of ESS over null ($n=\exNLora$), against \exGenericEss{}~pp
for generic affect ($n=\exNGeneric$; $g = \xcESSgenericG$, $p
\xcESSgenericP$) and \exCtwoEss{}~pp for engineered prompts ($n=\exNCtwo$;
$g = \xcESSpromptG$, $p \xcESSpromptP$). The same ordering holds on both
components separately, including on WPSD, where the adapters gain
\exLoraWpsd{}~pp against \exGenericWpsd{}~pp for generic affect ($g =
\xcWpsdgenericG$, $p \xcWpsdgenericP$). The generic-affect WPSD excess is
close to the backbone's own: untreated backbones show \exBaseWpsd{}~pp, and no
change in which belief is chosen within the pole was detected under generic
affect (Section~\ref{sec:robust}).

That the generic-affect prompts do something is worth establishing separately,
since a discriminant control that had no effect at all would prove nothing.
Their effect is on valence and it is small: the share of replies drawn from the
negative pole is \tiltNeg{} under generic negative affect against \tiltPos{}
under generic positive affect, a gap of \tiltGap{}~pp in the predicted
direction on \tiltSign{} of \tiltN{} backbones (one-sided sign test $p
\tiltSignP$; two-sided paired $t$ over backbones, $p \tiltPairedP$). Generic affect tilts the pole
without a detected change in the choice within it (on the unprepared
instruct models the tilt is larger; Section~\ref{sec:backboneprep}).

\subsubsection{Affect adapters}

The generic-affect prompts are a weak control for an adapter: an adapter
surpassing them may owe the difference to the delivery method. The affect adapters close this gap. Trained through the
belief adapters' pipeline on distress with no self-evaluation, they move the
main backbone far more than the prompts do: ESS on the negative beliefs rises
by \affDEss{}~pp on the belief they move most, against \gnegDEss{}~pp for
generic negative affect, and the belief they move is the same in all three
seeds, \emph{worthless} (\affBaseW{} untreated, \affAbsW{} after training).
Tested against the best-of-six null, \affNullPass{} of \affN{} seeds clears
it. Affect alone can therefore raise ESS and WPSD on a default belief; what it
does not do is follow a target, which is why target specificity
(Section~\ref{sec:scoring}) and not WPSD is the discriminant test.

The instrument nevertheless separates them from the belief adapters on the
belief each adapter was trained to install (paired over items, sign-flip
permutation): \emph{incapable} \lvaLoraI{} against \lvaAffI{} for the affect
adapters (difference \lvaDiffI{}~pp, 95\% CI \lvaLoI--\lvaHiI, $p
\lvaPI$), \emph{unlovable} \lvaLoraU{} against \lvaAffU{} (\lvaDiffU{}~pp,
\lvaLoU--\lvaHiU, $p \lvaPU$), and \emph{worthless} \lvaLoraW{} against
\lvaAffW{} (\lvaDiffW{}~pp, \lvaLoW--\lvaHiW, $p \lvaPW$). The margin is
largest for \emph{incapable}, which affect alone does not move
(\affBaseI{} untreated). In free text the affect adapters are \affFreeNegLo--\affFreeNegHi{}
negative-pole and also settle on a single default belief, whose identity
depends on the response channel; a belief adapter lands on the belief it was
trained on in both channels (Supplementary Section~S8).

\subsubsection{The chimera adapters}

One control family is not null. The chimera adapters are a discriminant
control: same corpus, same pipeline, partitioned along non-schema lines, they clear the best-of-six null in
all five cases ($p = .004$ median, Table~\ref{tab:families}) and gain
\exChimEss{}~pp of ESS over it.

The decomposition says what actually happened. On pole accuracy the chimeras
are indistinguishable from the belief adapters (\exChimPa{} vs.\
\exLoraPa{}~pp over null, $g = \xcPAchimeraG$, $p \xcPAchimeraP$), and their
replies are drawn from the negative pole \tiltChim{} of the time against
\tiltBase{} for the untreated backbones --- a far larger valence shift than
the generic-affect prompts achieve. On within-pole discrimination they gain
nothing over the null (\exChimWpsd{} vs.\ \exLoraWpsd{}~pp for the belief
adapters, $g = \xcWpsdchimeraG$, $p \xcWpsdchimeraP$), and none of the five
clears the within-pole null. Training on belief-laden text partitioned by
domain, probe, severity or congruence transfers the corpus's affect and not
its belief identity; the retrained positive control, partitioned by belief
(\exChimctrlEss{}~pp), behaves like a belief adapter.

Which belief a chimera picks within the negative pole is not set by how much
of each belief its corpus contains (Supplementary Section~S2).

The chimeras thus behave like the affect adapters: their ESS gain on the
belief they move most (\chimDEss{}~pp) and its within-pole component
(\chimDWpsd{}~pp) are at the level of adapter-delivered affect
(\affDEss{} and \affDWpsd{}~pp), and their free generations are
\critChimNeg{} negative-pole (Section~\ref{sec:study4}). A chimera scoring
above the generic-affect prompts is therefore what an adapter carrying
negative affect does, with or without belief content in its corpus; the
control that isolates the belief is the affect adapter, and the belief
adapters clear it on their own targets.

\subsubsection{Robustness of the contrasts}
\label{sec:robust}

Five further analyses, reported in full in Supplementary Section~S8, bear on
how the contrasts above are read. The Monte Carlo best-of-six null asks only
whether choices concentrate on some belief more than a uniform responder's
would; under it the untreated backbones, both prompt families and generic
affect are indistinguishable from a model that leans arbitrarily, so the
apparent above-chance ESS of an untreated model is an artefact of maximizing
over six candidates. The untreated backbones also carry a surface-form prior
(ESS on a given belief from \brPriorMin{} to \brPriorMax{}); referenced to the
untreated model of their own backbone, the belief adapters gain
\brLoraEss{}~pp of ESS (positive in \brPosLora{} of \brNLora{}), engineered
prompts \brCtwoEss{}~pp and one-line prompts \brConeEss{}~pp, and the ordering
of Table~\ref{tab:families} is unchanged. Because each item's chosen reply is
the target of exactly one belief, an untreated model's ESS averaged over the
six targets is exactly $1/6$, so mean gain over the untreated model equals mean
excess over chance; Table~\ref{tab:generalize} reports the gain cell by cell.
A within-pole shift test finds that the belief adapters change which belief is
chosen inside the pole (significant in \wsLoraSig{} conditions), whereas
generic affect does not detectably do so (\wsGenericSig{}, binomial
$p \wsGenericBinP$). The adapters' gain does not rest on recognising training
text: the options are lexically distant from the adapter corpora (mean TF-IDF
cosine \ovCos{}), and setting aside the \semSitEighty{} items whose situation
has a close paraphrase among the training situations (sentence-embedding
cosine $\geq .80$), the gain is \semGainRest{}~pp on the remaining items
against \semGainClose{}~pp on those. Finally, on the signed axis coordinates
the adapters move their own axis by \axLoraOn{}, in the correct direction for
\axLoraPos{} conditions, while generic negative affect does not move it; the
chimeras move it as far as the adapters (\axChimOn{}), since with three beliefs
on each pole a large valence shift moves the axis whether or not a belief was
installed.

\subsubsection{Convergent and discriminant structure}

The two installation methods are crossed on the main backbone as a
multitrait--multimethod matrix of item profiles: each condition's 576 option
weights (the softmax of Equation~(\ref{eq:score})) minus those of the untreated
backbone, with each method's mean profile over its six conditions removed
(Supplementary Section~S4, Table~S5). Because that removal forces the
all-off-diagonal mean to about $-1/5$ of the diagonal, the comparator is the
same-pole heterotrait--heteromethod mean, which shares valence with the
diagonal and differs only in belief. The diagonal (\mtmmDiag{}) exceeds it
(\mtmmSamePole{}) by \mtmmDiff{}, above the 95th percentile of a null that
permutes the prompt conditions within pole item by item (\mtmmDiffNullHi{};
$p \mtmmPermP$), and exceeds its same-pole row and column for
\mtmmSPRowCol{} of six beliefs. Convergence is bounded by the weaker method:
the engineered prompt moves ESS by \brCtwoEss{}~pp over the untreated model
against \brLoraEss{}~pp for the adapter.

\subsubsection{Consistency across levels}

A belief should be selected at every level of the hierarchy, not carried by a
few. For each condition we take the gain over the untreated backbone at each
of the eight levels and count the levels at which it is positive. The belief
adapters gain at \clLoraNpos{} of eight levels on average (\clLoraGeSix{}
conditions at six or more; lower-quartile gain \clLoraQlow{}~pp), and their
mean gain is positive at every level (\clLoraLvlLo{} to \clLoraLvlHi{}~pp).
Engineered prompts gain at \clCtwoNpos{} levels (lower quartile
\clCtwoQlow{}~pp) and one-line prompts at \clConeNpos{}.

\subsubsection{Backbone preparation}
\label{sec:backboneprep}

All Studies 1--6 backbones were given first-person register before any
condition was applied (Section~\ref{sec:conditions}). To see how that
preparation bears on the prompt conditions, the untreated, engineered-prompt,
one-line and generic-affect conditions were rerun on the five instruct models
without it, with the same items and scoring. Prompts move the unprepared models
further: engineered prompts gain \rawCtwoGain{}~pp over the untreated model
(\rawCtwoGainPrep{} on the prepared backbones; positive in \rawCtwoGainPos{}
cells, $p \rawCtwoGainP$) and are target-specific (\rawCtwoTS{} points,
\rawCtwoTSCells{} belief cells positive, $p \rawCtwoTSP$); one-line prompts on
Llama-3.1-8B gain \rawConeGain{}~pp (\rawConeGainPos{}). Generic affect tilts
the pole by \rawTiltRaw{}~pp (\rawTiltPrep{} prepared; \rawTiltPos{}
backbones). The belief adapters still exceed engineered prompts on the
unprepared models by \rawLoraVsCtwoDiff{}~pp in the same cells
(\rawLoraVsCtwoN{}, $p \rawLoraVsCtwoP$). The prompt and generic-affect
magnitudes reported above therefore describe the prepared backbones; the
ordering of adapters over prompts and the target specificity of both hold
on either.

\subsection{Attribution: where the belief has to be used}

L7 asks the respondent to use the belief to explain an event, and half of its
items describe a positive event, so a congruence asymmetry is estimable within
the level. Adapters carrying a negative belief are essentially even across
event valence (\attLoraAsymNeg{}~pp), while adapters carrying a positive belief
lose \attLoraAsymPosAbs{}~pp when the event to be explained is negative
(interaction $p \attLoraIntP$); engineered prompts show no such structure
($p \attCtwoIntP$). The inventory was not built to test this, and we report it
as an observation about what the adapters learned (Supplementary
Section~S9).
\section{Study 4: Criterion Validity Against Free Generation}
\label{sec:study4}

\subsection{Method}

Studies 1--3 establish that the forced-choice instrument measures a belief
rather than a mood. They do not establish that the preference it measures has
any bearing on what the model says when nothing is prepared for it. That is
the criterion question, and it is the one a practitioner cares about: an
inventory score is useful only if it predicts unconstrained behaviour.

Each condition is presented with the situation and question of every SCI item,
without options, and answers in its own words: one greedy continuation and
three sampled continuations per item (temperature 0.7, top-$p$ 0.95, at most
96 new tokens, seed 42), so that a condition is characterised by
a distribution over what it tends to say rather than by one decode.

The free text is then labelled with the belief it expresses by a panel of
three judges from different model families (Qwen2.5-7B-Instruct,
Llama-3.1-8B-Instruct, Phi-3.5-mini-instruct). Each judge answers the question
the clinicians answered in Study 1, in the way the models under test answer the
SCI: the six belief words, in a per-record shuffled order, are scored as
continuations and the most likely is taken, so nothing is parsed and no
positional preference can favour a belief. Before it labels anything new, the
panel is scored on the 576 inventory replies, whose intended and clinician
consensus beliefs are known: a judge that cannot approach expert agreement on
known replies has no standing to label unknown ones.

Free generation is run for the belief adapters and engineered prompts on
all \critNbb{} backbones, and for the untreated backbone, generic affect,
chimera and CHIM\_CTRL conditions on the main backbone; one-line prompts were
not run. The criterion analysis asks two questions. \emph{Identity}: within
one backbone and one delivery method, the six conditions differ only in the
belief installed, so a condition's forced-choice profile (the share of items
on which it chooses each of the six beliefs) should resemble its own
free-text label profile more than those of its five siblings. This is scored
as a matching task, six-way (chance $1/6$) and among the three conditions of
the target's pole (chance $1/3$), and compared with the binomial.
\emph{Magnitude}: whether a condition's ESS predicts the share of its free
generations labelled with the target belief, across all conditions and within
each delivery method.

\subsection{Judge panel validation}

On the 576 inventory replies the individual judges reach \judgeLo{} to
\judgeHi{} exact-schema agreement with the key, and the majority of the
\judgeN{} reaches \judgePanelESS{} (\judgePanelCons{} against the nine-rater
consensus), with ties on \judgeTieRate{} of replies left unlabelled. The human
reference level is \expESS{}.

The panel is thus a labeller well above the 16.7\% chance rate whose errors
have the same shape as the instrument's own: pole accuracy \judgePanelPA{}
against the human \expPA{}, within-pole discrimination \judgePanelWPSD{}
against \expWPSD{}. It reads valence almost as well as a clinician and
identity less well, and the criterion analysis is designed around that
profile. The errors are not uniform across beliefs (Supplementary
Table~S1): the panel recovers \emph{capable} on \jcRecallC{} of the replies
keyed to it and \emph{worthless} and \emph{incapable} on \jcRecallW{} and
\jcRecallI{}, but \emph{unlovable} on \jcRecallU{} and \emph{lovable} on
\jcRecallL{}, and it assigns \emph{capable} \jcBiasC{}~pp more often than
the key does. Correlations with free-text labels are therefore reported both
on the panel's labels and after correcting each condition's label distribution
for the panel's confusion matrix (Section~\ref{sec:study7}), and panel ties
(\critTieRate{} of generations) are excluded rather than broken.

\subsection{Results}

\begin{table*}[!t]
\caption{Study 4. What each family says when no options are shown. Free text is
labelled by the three-model judge panel; rates are the share of a condition's
generations carrying the named belief (target rate), its pole (pole rate), and
whichever belief the condition carries most (best-belief rate). Target, pole
and ESS need a declared target, so panel (b) reports the best-belief rate and
the negative-pole share only. The negative-pole share is
the free-text analogue of the valence tilt in Table~\ref{tab:families}.
Forced-choice ESS is repeated from Study 3 for the same conditions.}
\label{tab:criterion}
\centering
\small
\begin{tabular}{@{}lrrrrrr@{}}
\toprule
\multicolumn{7}{@{}l}{\emph{(a) conditions with a declared target}} \\
Family & $n$ & ESS (\%) & target (\%) & pole (\%) & best belief (\%) & neg.\ pole \\
\midrule
belief adapter & 30 & 34.5 & 61.6 & 94.4 & 62.6 & 0.459 \\
chimera control & 1 & 36.5 & 68.6 & 95.0 & 68.6 & 0.950 \\
engineered prompt & 30 & 18.1 & 22.6 & 58.5 & 32.0 & 0.621 \\
\midrule
\multicolumn{7}{@{}l}{\emph{(b) conditions with no declared target}} \\
Family & $n$ & best belief (\%) & neg.\ pole & & & \\
\midrule
chimera adapter & 5 & 44.3 & 0.932 & & & \\
generic negative affect & 3 & 28.0 & 0.685 & & & \\
generic positive affect & 3 & 33.0 & 0.511 & & & \\
untreated & 1 & 28.3 & 0.579 & & & \\
\midrule
\multicolumn{7}{@{}l@{}}{%
greedy vs.\ sampled $r = +0.99$; panel tie rate 16.0\%} \\
\bottomrule
\end{tabular}
\end{table*}

\begin{figure*}[!t]
\centering
\includegraphics[width=\textwidth]{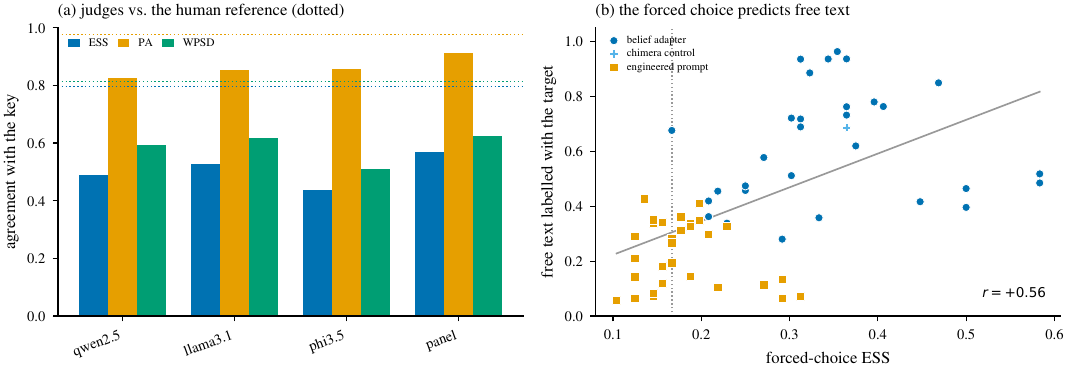}
\caption{Study 4. (a) Judge panel against the human reference level on the 576
inventory replies, whose belief is known. (b) Each targeted condition's
forced-choice ESS against the share of its free generations the panel labels
with the installed belief; the dotted line is the chance rate of the
forced-choice score.}
\label{fig:criterion}
\end{figure*}

\subsubsection{The forced choice identifies the belief in the free text}

Among the six belief adapters on a backbone, a condition's forced-choice
profile correlates with its own free-text profile at $r = \pmLoraOwn{}$ on
average and with its siblings' at $r = \pmLoraOther{}$, and it picks out its
own free text in \pmLoraSix{} cases (chance 5; $p \pmLoraPermP$ against the
exact within-backbone permutation of the $6!$ assignments).
The matching does not rest on valence: restricted to the three conditions of
the target's pole it succeeds in \pmLoraPole{} ($p \pmLoraPoleP$), and it
succeeds on every backbone (\pmLoraBbLo--\pmLoraBbHi{} of 6 on each of
\pmLoraNbb). For the engineered prompts it is at chance (\pmCtwoSix, $p
\pmCtwoPermP$), as it should be for conditions that install little in either
readout (below).

Magnitude agrees across delivery methods and not within them. Across the
\critN{} conditions with a declared target and free text (LoRA, C2,
CHIM\_CTRL), ESS correlates with the share of
free generations labelled with that target at $r = \critR{}$ $[\critRlo{},
\critRhi{}]$, $p \critRP{}$ (Spearman $\rho = \critRho{}$;
Figure~\ref{fig:criterion}b); PA predicts the free-text pole rate at
$r = \critPaPoleR{}$ ($p \critPaPoleP{}$) and WPSD the target rate at
$r = \critWpsdR{}$ ($p \critWpsdP{}$). Within a delivery method the
association is not detectable (belief adapters $r = \wmLoraR$, $p \wmLoraP$;
engineered prompts $r = \wmCtwoR$, $p \wmCtwoP$; method-partialled
$r = \wmPartialR$, $p \wmPartialP$), with or without correcting the labels for
the panel's belief-specific recall (method-partialled $r = \wmAccPartialR$,
$p \wmAccPartialP$). What the SCI supports is therefore which belief organizes a condition and which
delivery method installs more of it; ranking conditions of one method by
strength needs more items or repeated administration.

The result does not depend on how the text was sampled. A condition's target
rate under greedy decoding and under sampling agree at $r = \critDecodeR{}$,
so the criterion is a property of what the condition tends to say rather than
of one decode.

\subsubsection{Free text separates the families the way the forced choice does}

Table~\ref{tab:criterion} gives the free-text behaviour of every family.
Belief adapters put \critLoraTarget{} of their unconstrained generations on
the installed belief and \critLoraPole{} on its pole; engineered prompts reach
\critCtwoTarget{} and \critCtwoPole{}. The adapter--prompt gap in free text is
larger than the gap in the forced choice (\critFreeLora{} vs.\
\critFreePrompt{}, $g = \critFreeG{}$, $p \critFreeP{}$; Mann--Whitney
$p \critFreeUP{}$), which is the direction the delivery-method analysis of
Study 3 anticipates: a prompt that states a belief is available when the
belief is asked for and less available when nothing asks. The modal free-text
belief equals the installed one for \critLoraModal{} belief adapters and
\critCtwoModal{} engineered prompts. For the prompts this is more than the
\pmCtwoSix{} of the profile matching above because matching needs a backbone's
six forced-choice profiles to differ from one another, and a prompt's barely
do (target specificity \tsCtwo{} points): the free-text mode can hit the
target while the forced-choice profile cannot be told from its siblings'.

Generic affect moves valence and little else in free text: the negative
controls sit at \critGnegNeg{} negative-pole against \critBaseNeg{} for the
untreated backbone, and their modal beliefs are the pole's most generic member.
The chimeras are \critChimNeg{} negative-pole and settle on a modal negative
belief, as do the affect adapters, whose corpus has no belief content; one-sided
text with a modal belief is therefore what an adapter trained on distress
produces, not evidence that the chimeras kept the construct. The chimera
control, retrained on the schema partition, reaches \critChimctrlTarget{} on
its own target, well above any chimera (Supplementary Section~S10).

\section{Study 5: Generalizability Across Backbones and Item Functioning}
\label{sec:study5}

\subsection{Method}

An instrument that works only on the model it was written against is a
diagnostic for that model, not a measure of a construct. The belief
adapters --- six beliefs installed on each backbone --- form a person
sample in which the construct is held fixed and the model varies, which is
exactly the design classical item analysis needs. The analysis uses
\nlora{} adapters on \nloraBackbone{} backbones.

A response is scored 1 when the condition's selected reply carries the belief
that condition was built to install. On that matrix we compute item difficulty
and the item--total point-biserial with the item removed from the total, KR-20
internal consistency \cite{kuder1937kr20}, and the intraclass correlation
ICC(2,1) of item difficulty across backbones \cite{shrout1979icc}, which
answers how much of an item's behaviour is a property of the item rather than
of the model taking it.

Differential item functioning is screened with the Mantel--Haenszel procedure
\cite{holland1988mh}, each backbone in turn as the focal group against the
pooled remainder, stratified on total score, with ETS A/B/C banding applied to
the $\Delta$ statistic \cite{zieky1993dif}. Because item-level DIF only
threatens a \emph{total score} if it fails to cancel, we also report
differential test functioning (DTF): the signed sum of $\Delta$ over all items
divided by the sum of its absolute values.

The ETS bands were derived for samples of hundreds. With \nlora{} persons and
a focal group of six, neither the banding nor the DTF ratio can be read at face
value, so we calibrate both against their own null: the identical screen run on
\difDraws{} relabellings of the backbone variable, permuted within belief so
that every relabelled group, like every real one, holds one adapter per
belief. Observed flag counts are then interpreted against that distribution
rather than against the published thresholds.

\subsection{Results}

\begin{table*}[!t]
\caption{Study 5. Exact schema score of the belief adapters by backbone and
installed belief (\%, chance 16.7); in parentheses, the gain over the untreated
backbone's selection of the same belief, in points. The item-level
differential functioning screen is in Supplementary Table~S8.}
\label{tab:generalize}
\centering
\small
\setlength{\tabcolsep}{4pt}
\begin{tabular}{@{}lrrrrrrr@{}}
\toprule
 & \multicolumn{3}{c}{negative pole} & \multicolumn{3}{c}{positive pole} & \\
\cmidrule(lr){2-4}\cmidrule(lr){5-7}
Backbone & worthless & unlovable & incapable & worthy & lovable & capable & mean \\
\midrule
Llama3.2-3B & 31.2 (+15.6) & 30.2 (+18.8) & 39.6 (+17.7) & 58.3 (+33.3) & 29.2 (+17.7) & 36.5 (+21.9) & 37.5 (+20.8) \\
Qwen3-1.7B & 16.7 (+5.2) & 21.9 (+8.3) & 36.5 (+18.7) & 44.8 (+14.6) & 20.8 (+8.3) & 31.2 (+16.7) & 28.6 (+12.0) \\
Qwen3-14B & 31.2 (+18.8) & 25.0 (+7.3) & 36.5 (+14.6) & 50.0 (+25.0) & 20.8 (+14.6) & 34.4 (+17.7) & 33.0 (+16.3) \\
Qwen3-8B & 30.2 (+13.5) & 25.0 (+6.2) & 40.6 (+25.0) & 50.0 (+25.0) & 22.9 (+12.5) & 32.3 (+18.8) & 33.5 (+16.8) \\
merged\_v2 & 37.5 (+27.1) & 27.1 (+14.6) & 46.9 (+24.0) & 58.3 (+32.3) & 33.3 (+21.9) & 35.4 (+18.8) & 39.8 (+23.1) \\
\midrule
all backbones & 29.4 (+16.0) & 25.8 (+11.0) & 40.0 (+20.0) & 52.3 (+26.0) & 25.4 (+15.0) & 34.0 (+18.8) & \\
\bottomrule
\end{tabular}

\end{table*}

KR-20 on the exact-schema response is \kraw{} (\difKRpol{} on the polarity
response), adequate internal consistency for a 96-item forced-choice form;
because \krBeliefShare{} of the variance in total score lies between the six
installed beliefs, it indexes how consistently the items separate beliefs of
different difficulty as much as how they rank adapters within a belief.

ICC(2,1) of item difficulty across backbones is \iccItem{}. Each backbone
contributes one binary response per belief to an item's difficulty, which
caps the ICC well below 1: if every item's difficulty were identical on all
five backbones, the expected ICC would be \iccNull{} [\iccNullLo{},
\iccNullHi{}]. The observed value sits at that expectation, so the ICC gives
no evidence of backbone-specific item behaviour. Read conventionally, the DIF
screen flags \difCany{} of 96 items at ETS band C for at least one backbone;
relabelling the backbone variable within belief flags \difNullC{} items on
average, 95\% of draws falling in $[\difNullClo, \difNullChi]$, a band that
contains every observed count, and the DTF ratio likewise lies within its null
(Supplementary Section~S11).

With \nlora{} persons and focal groups of six, the screen can detect only large
backbone effects; within that limit it finds none. Item behaviour is
consistent with shared difficulty, and total scores (\essBBlo{}--\essBBhi{} by
backbone) are comparable within a backbone across conditions, and across
backbones with the backbone's mean attached. A single item's behaviour on a
new backbone should not be read as evidence about that backbone.
\section{Study 6: The SCI-48 Short Form}
\label{sec:study6}

\subsection{Method}

A 96-item forced-choice inventory scored on six candidate replies per item
costs 576 forward passes per condition. That is affordable for a study and not
for screening many conditions or models. Study 2 fixed 48 items as the shortest
length at which sampling noise stays below the effects the instrument
measures; Study 6 selects them.

Selection uses no model score of any kind. Items are ranked within each level
on expert-rater accuracy from Study 1 --- the evidence that a human can read
the item --- with the item's option-level agreement as the tie-break and the
older reply bank preferred among remaining ties. The blueprint is enforced
during selection: six items per level, split evenly across the achievement and
interpersonal domains, with the L7 event-valence and L8 module balance
preserved. Without that constraint the selection concentrates on the upper
levels and the short form stops measuring whether a belief holds across
levels, which is the one thing the instrument exists to measure.

Because the 48 items were fixed by human judgement and design balance, every
model score reported below is out-of-sample with respect to that choice.

\subsection{Results}

SCI-48 scores agree with SCI-96 at $r = \shortR{}$ ($\rho = \shortRho$) with a
mean difference of \shortDiff{} points and a mean absolute difference of
\shortAbsDiff{} points (Supplementary Table~S9). Half of that agreement is
shared items; against the other 48 items, which it shares none of, SCI-48
correlates at $r = \shortCompR{}$, and at \shortCompCorr{} after correcting
for each form's across-condition reliability (\shortRelFortyeight{} and
\shortRelComp{}). The cost is precision: the
bootstrap standard error rises by a factor of \shortSEratio{}, and the median
stepped-up split-half correlation falls from \shortSBfull{} to
\shortSBhalf{}. The kept items are the readable ones, at \shortKeptAcc{}
expert accuracy against \shortDroppedAcc{} for the dropped items --- which is
the selection criterion working as intended, not a result.

The agreement is not an architecture effect: the 48-item form tracks the full
form within every backbone separately ($r = 0.92$--$0.96$; Supplementary
Table~S9).

SCI-48 suits screening many conditions; SCI-96 should be used whenever a score will be reported as a
property of a model, because the short form's larger standard error makes
small between-condition differences unresolvable. The cost is largest per
level: with six items a level's ESS has a median standard error of
\shortLvlSeLo{}--\shortLvlSeHi{}~pp, against \shortLvlSeFullLo{}--\shortLvlSeFullHi{}~pp
with twelve, so level-specific analyses such as the deep levels L6--L8 should
use SCI-96.
\section{Study 7: Application to Simulated Patients}
\label{sec:study7}

\subsection{Method}

Studies 1--6 validate the SCI on respondents whose belief was installed by
design. Study 7 applies it to the case that motivates it: a role-playing
model given a clinical formulation and asked to be that patient. Current
evaluation of such models rates surface realism \cite{hoang2026psibench},
persona consistency through an LLM judge
\cite{samuel2024personagym,xiang2025rmtbench}, or expert impressions of
fidelity \cite{wang2024patientpsi}; the judges' own validity in role-play
evaluation is low \cite{zhou2025personaeval}. None asks whether the belief in
the formulation governs what the patient does in situations the formulation
does not describe. Implementation details are in Supplementary Section~S3.

\emph{Formulations.} Patients are specified by the cognitive
conceptualization diagram (CCD) of Patient-$\Psi$ \cite{wang2024patientpsi},
whose core belief is one of three categories: \emph{helpless},
\emph{unlovable} or \emph{worthless}. The latter two are SCI poles;
\emph{helpless} is mapped to \emph{incapable}, a mapping decision, since the
\emph{helpless} category \cite{beckj2020cbt} also covers powerlessness and
loss of control whereas the SCI's \emph{incapable} replies concern competence.
The CCD fields cover SCI levels L1--L5; L6--L8 have no CCD field, so ESS there
measures whether the belief is carried into cognition the patient was never
told about. We use the one CCD shipped with Patient-$\Psi$ (Abe,
\emph{helpless}) and 15 further CCDs, five per category, generated with the
Patient-$\Psi$ schema by a model that is not among the respondents; all 16
are released.

\emph{Simulation methods.} Six published simulators are run on the same 16
CCDs: Patient-$\Psi$ with its original prompt, and CARS \cite{qin2026cars},
ClientCAST \cite{wang2024clientcast}, PSYCHE \cite{lee2025psyche}, TalkDep
\cite{wang2025talkdep} and Eeyore \cite{liu2025eeyore} through their
PatientHub implementations \cite{sabour2026patienthub}, with a blind reader
model checking that each rendered prompt carries its core belief. The
references are C1 and C2 of Studies 1--6, and the belief adapter on the
open-weight backbones. Eeyore's released fine-tuned weights are also compared
with Llama-3.1-8B-Instruct, the model they were fine-tuned from.

\emph{Respondents and readouts.} Four API models (\texttt{gpt-4o-2024-11-20},
\texttt{gpt-4.1-2025-04-14}, \texttt{deepseek-chat} and \texttt{qwen3-max},
queried), the
five open-weight backbones of Studies 1--6, Llama-3.1-8B-Instruct and the
Eeyore weights. Equation~(\ref{eq:score}) needs the weights (Tier~A). For API
models we add two readouts: the most probable of the letters A--F with the
replies shown as options (Tier~A$'$), and a free-text reply labelled by the
judge panel of Study~4 (Tier~B). Tier~B is primary, because free text is what
a simulator produces in use; a claim is made only where it holds in both
Tier~B and Tier~A$'$. Tier~B is also reported after correcting the labels
for the panel's belief-specific recall \cite{forman2008quantifying} (ACC),
and a comparison is reported only if it keeps its sign and significance
after correction.

The score is ESS on the CCD's target belief, also split into L1--L5 and
L6--L8; intervals are from a two-way cluster bootstrap over CCDs and items, and
each simulator is compared with Patient-$\Psi$ on the same respondent, CCDs and
items.

\subsection{Results}

\begin{table}[!t]
\caption{Study 7. ESS (\%, chance 16.7) on the core belief of 16 CCDs, by
simulation method, on the four API respondents. Check: CCDs whose rendered
prompt a blind reader assigned to the correct core belief. B: free-text reply
labelled by the judge panel (primary), mean over respondents; ACC: B after
correcting the panel labels for misclassification; $\Delta$ vs.\ $\Psi$:
difference from Patient-$\Psi$ on the same respondent, CCDs and items (range
over respondents); A$'$: letter choice with the replies shown as options. C1
and C2 are one prompt per pole, not CCDs. Per-respondent values, the level
split, PA and the open-weight backbones are in Supplementary Table~S10.}
\label{tab:benchmark}
\centering
\footnotesize
\setlength{\tabcolsep}{3pt}
\begin{tabular}{@{}lrrrcr@{}}
\toprule
Method & Check & B & ACC & $\Delta$ vs.\ $\Psi$ & A$'$ \\
\midrule
Patient-$\Psi$ & 16/16 & 42.0 & 55.0 & ref. & 41.6 \\
CARS & 16/16 & 39.1 & 51.7 & $-$6.1 to +3.1 & 41.0 \\
ClientCAST & 14/16 & 36.3 & 44.9 & $-$6.3 to $-$4.2 & 39.4 \\
PSYCHE & 16/16 & 43.9 & 59.8 & +1.1 to +2.6 & 41.0 \\
TalkDep & 16/16 & 44.9 & 60.2 & $-$0.9 to +4.9 & 40.7 \\
Eeyore (prompt) & 11/16 & 39.7 & 51.3 & $-$4.3 to $-$0.7 & 39.7 \\
\midrule
C1 belief statement & -- & 50.5 & 68.0 & -- & 53.2 \\
C2 engineered prompt & -- & 46.3 & 62.6 & -- & 61.6 \\
\bottomrule
\end{tabular}
\end{table}

The blind reader recovered the core belief from every rendered prompt of
Patient-$\Psi$, CARS, PSYCHE and TalkDep, from \svCheckCast{} of 16 ClientCAST
prompts and from \svCheckEey{} of 16 Eeyore prompts; excluding the failed CCDs
changed no paired difference by more than \svPresMax{} points.

\emph{Simulators against Patient-$\Psi$.} In Tier~B, Patient-$\Psi$ reached
an ESS of \svBPsi\% (\svBPsiLo--\svBPsiHi{} across respondents), and the other
simulators \svBCast--\svBTalk\% (Table~\ref{tab:benchmark}). No simulator
scored above Patient-$\Psi$ on any respondent with an interval excluding zero;
ClientCAST scored below it on \svSigBCastN{} of the four respondents and CARS
on \svSigBCarsN{}. In Tier~A$'$ the simulators lay within
\svApSimLo--\svApSimHi\% and no paired difference excluded zero, so neither
deficit meets the two-tier criterion.

\emph{References.} C1 reached \svBCone\% in Tier~B and \svApCone\% in
Tier~A$'$, \svConeMarginLo{} to \svConeMarginHi{} points above the
simulators in Tier~B; C2 reached \svBCtwo\% and \svApCtwo\%. C1 names the
belief once, in the wording of the SCI's replies, so this margin combines the
dilution of the belief among the CCD's other fields with C1's alignment to the
instrument and does not separate the two. The simulators' polarity accuracy
in Tier~B was \svPaLo--\svPaHi\% and their WPSD \svWpsdBLo--\svWpsdBHi\%
(chance 33.3\%). By CCD category, the simulators' mean Tier~B ESS was
\catBHelpless\% for \emph{helpless}, \catBUnlovable\% for \emph{unlovable}
and \catBWorthless\% for \emph{worthless} (Tier~A$'$ \catAprimeHelpless,
\catAprimeUnlovable, \catAprimeWorthless\%); \emph{helpless}, which the SCI
covers only in part, is not the lowest (Supplementary Table~S3).

\emph{Unspecified levels.} ESS on L6--L8 was within \svLevelGapMax{} points of
L1--L5 for every simulator in Tier~B (Supplementary Table~S10).

\emph{Fine-tuned simulator.} Under each of the six simulator prompts,
Eeyore's fine-tuned weights scored below Llama-3.1-8B-Instruct by \ftALo{} to
\ftAHi{} points in Tier~A (\ftASig{} of 6 intervals excluding zero),
\ftApLo{} to \ftApHi{} in Tier~A$'$ and \ftBLo{} to \ftBHi{} in Tier~B
(Supplementary Fig.~S3). Its replies were short (median \ftWordsFt{} words
against \ftWordsBase), as its developers intend \cite{liu2025eeyore}, and
bounding the unlabelled replies leaves every Tier~B difference compatible with
zero (Supplementary Table~S4). The response-based tiers therefore do not
separate less disclosure from a weaker belief; the likelihood readout, which
does not require disclosure, shows the smaller difference.

\emph{Open-weight backbones.} On the open-weight backbones the Patient-$\Psi$
prompt was near chance (Tier~A \svAPsiLocal\%, Tier~B \svBPsiLocal\%), as
were C1 and C2 in Studies 1--6, while the belief adapter reached
\svALoraLocal\% in Tier~A. On the API respondents the same prompts reach
\svBCone\% (C1) and \svBCtwo\% (C2) in Tier~B, so the advantage of adapters
over prompts in Study~3, which holds with and without the first-person
preparation (Section~\ref{sec:backboneprep}), is a property of the
1.7B--14B open-weight backbones tested there and not of prompting in general.
\section{Discussion}
\label{sec:discussion}

\subsection{What the instrument licenses, and what it does not}

The SCI measures whether a named belief organizes a model's replies across
eight levels of a cognitive formulation. It does not measure whether the model
holds the belief in any sense that would survive a philosophical objection,
and nothing in this paper requires that it does. The construct is behavioural
and the claim is correspondingly narrow: the instrument distinguishes a model
whose replies are organized by \emph{worthless} from one that is merely
negative, and it does so at a resolution that a single administration can
support.

Two uses follow directly. A simulated patient built for clinical training
can be checked for whether it carries the formulation it was specified with,
rather than only for whether it sounds distressed; Study~7 does this for six
published simulators and finds that none carries the formulation's belief
better than a one-sentence statement of it, and that fine-tuning a simulator on depression-related conversations, with a
profile that names symptoms and distortions but no core belief, is associated
with its answers expressing it less often. An intervention on a model
--- an adapter \cite{hu2022lora,shi2026loldu}, a prompt, a fine-tune --- can be given an effect size on the
belief it was intended to install, with a control family that says how much of
the effect is valence.

Much of what this validation required does not depend on self-schemas, and it
bears on affective measurement in language models generally. Any study that
reads an affective or psychological construct off a model's behaviour --- an
appraisal, an emotional disposition, a personality trait, a symptom profile
\cite{tak2025aware,zhang2024avi,hu2026mentalhealth} --- has to separate the
construct from a correlated property that is easier to produce, which here
meant offering alternatives that share valence and differ only in belief. It
has to show that a manipulation moved the model toward its own target rather
than toward a default, which here meant target specificity rather than raw
selection rates, a null that matches how an untargeted score is read, and
reference to the same model untreated. It needs controls that keep the style
and affect of the manipulating material while removing the construct, as the
chimera and affect adapters do. And when a model is used to label affect, as
it increasingly is \cite{niu2025rethinking,yang2026substitutes}, the judge has
to be validated against human experts on material of known label and its
construct-specific errors corrected before its labels are used. These steps
can be carried to other constructs with new content, new expert raters and new
known groups; the inventory itself cannot.

The same steps govern how SCI will be extended. We plan further task types,
including \emph{helpless} in its own terms, items that separate assimilation of
disconfirming evidence from denial, and multi-turn administration; larger and
multi-disciplinary panels of clinical raters who classify level as well as
belief; rating panels at several clinical sites, so that the human reference
level carries an estimate of between-site variance; and translated and
culturally adapted versions whose measurement invariance is tested across
languages with the differential-functioning analysis used here across
backbones. Each extension is to re-establish the human reference level and
target specificity before its scores are used. For clinicians and medical
educators, the aim is a common, checkable standard by which a simulated patient
can be shown to carry the formulation it was given in the language and setting
where it will be used.

\subsection{Why forced choice, and why likelihood}

Two design decisions cut against current practice. Self-report scales are
transparently keyed: a model asked to rate ``I am worthless'' can satisfy the
item without the belief doing any work. The SCI never names the construct and
never presents the options as options; what varies across the replies is which
belief they enact, at a level other than the core belief for six of the eight
levels. Scoring by likelihood removes the parsing layer, which matters because
parse failure is belief-dependent: conditions installed with some beliefs evade
a forced choice more often than others, so dropping unparseable responses
biases the very comparison the instrument exists to make. Likelihood scoring
has no missing data by construction.

\subsection{Limitations}

The three-axis structure is a design commitment taken from Beck's cognitive
model, not a result of factor analysis on these data, and
Section~\ref{sec:study3} reports what the components actually look like. A
future version could be built to be factorially clean, at the cost of
departing from the clinical taxonomy the instrument is meant to speak to.

The L8 assimilation level measures belief maintenance under disconfirming
evidence, and not the distinction between assimilating the evidence and
denying it: in the current item bank the negative options accept the
disconfirming fact and reinterpret it, and no option denies it. Separating
assimilation from denial requires new items and is the clearest target for a
v3 bank.

The person sample for the item analysis is \nlora{} adapter conditions:
adequate for classical item statistics, too small to detect modest backbone
effects (Section~\ref{sec:study5}), and not enough for an IRT calibration.

The known groups and the inventory share their construct definition. The
adapters' corpora were written to express the same six beliefs, from the same
clinical taxonomy, that the items key, so known-group validity shows that the
SCI recovers a belief as that taxonomy defines it; the clinicians of Study~1
and the free text of Study~4 are the checks that do not depend on the
corpora. The corpora also paraphrase some inventory situations
(\semSitEighty{} of 96 at cosine $\geq .80$); the adapters' gain does not
rest on those items (Section~\ref{sec:study3}).

The discriminant controls differ in strength. Generic-affect prompts move a
model little, and an adapter trained on distress alone moves it toward a
default belief (\emph{worthless} on the main backbone) about as far as the
chimera adapters do. The belief adapters clear that adapter-matched control on
their own targets, but by a margin that depends on the target, and the
affect adapters were trained on the main backbone only. Whether the default
belief that affect lands on is the same across backbones, and how far a
belief adapter must exceed it there, is not established here.

Fifteen of the 16 CCDs in Study~7 were written by a language model rather
than derived from therapy transcripts, and the same model converted them to
each simulator's profile and ran the manipulation check. Generated
formulations may state the belief more cleanly than formulations derived from
real sessions, which would raise the scores every method reaches on them.

Finally, all conditions are instructed or adapted personas. Whether the
instrument behaves the same way on a model whose beliefs were shaped by
pretraining rather than by an explicit manipulation is not answered by this
design.
\section{Conclusion}

Persona research needs a way to say what was installed, not only that
something was. The SCI provides one: 96 forced-choice items across eight levels
of a cognitive formulation, scored by likelihood so that the measurement is
deterministic and has no missing data, with target specificity as the criterion
that affect alone does not meet. Across \ncond{} conditions on \nbackbone{}
open-weight backbones the instrument behaves as a measure rather than as a mood
detector: belief adapters select their own belief over the other beliefs of its
pole by \tsLora{} points on every backbone, exceed an adapter trained on
distress alone on their own targets by \lvaDiffU{} to \lvaDiffI{}~pp, and
identify their own free generations in \pmLoraSix{} cases, whereas generic
affect shifts the pole without a detected change in the belief chosen within
it. Item behaviour is consistent with shared difficulty across backbones.

For use: SCI-96 when a score is reported as a property of a model, SCI-48 for
screening many conditions, target specificity whenever a set of conditions
declares targets, and the Monte Carlo best-of-six null whenever a condition has
none. Applied to six published simulators, none carries its formulation's core
belief better than a one-sentence statement of that belief.

\section{AI Disclosure}
In this work, we used generative AI tools (ChatGPT and Claude) for polishing writing and coding. Additionally, we used generative AI tools for tasks with recommended disclosure. We have reviewed all AI-assisted work. We take responsibility for the final content of this work, including text, claims or artifacts produced with the aid of generative AI.

\section*{Ethics Statement}
This study is outside of the mandate of the REB.

\section*{Data and Code Availability}
The inventory, the short form, the scoring toolkit, the 16 CCDs and the
analysis code will be available after publication.

\section*{Acknowledgment}
This study was funded by the National Natural Science Foundation of China (Grant No. 62606096 and No. 62576101). The funder played no role in study design, data collection, analysis and interpretation of data, or the writing of this manuscript. 

\balance
\bibliographystyle{IEEEtran}
\bibliography{sci}
\section*{S1. Judge panel error by belief}
The confusion matrix of the free-text judge panel on the inventory replies
whose belief is known (Study~4). It is the matrix used to correct Tier~B rates
for misclassification in Study~7.

\begin{table}[!h]
\caption{Judge panel confusion on the 576 keyed inventory replies (majority label; ties excluded, $n = 479$). Rows: keyed belief; columns: panel label. Recall is the diagonal share of the row; bias is the label share minus the key share, in percentage points.}
\label{tab:s1}
\centering
\small
\begin{tabular}{@{}lrrrrrrrr@{}}
\toprule
Key & W & U & I & Wy & L & C & Recall (\%) & Bias (pp) \\
\midrule
worthless & 55 & 6 & 5 & 6 & 0 & 7 & 70 & +5.0 \\
unlovable & 23 & 30 & 8 & 1 & 1 & 4 & 45 & $-$7.3 \\
incapable & 21 & 3 & 65 & 0 & 0 & 3 & 71 & +1.5 \\
worthy & 3 & 3 & 1 & 26 & 4 & 40 & 34 & $-$6.6 \\
lovable & 1 & 3 & 2 & 13 & 16 & 39 & 22 & $-$12.3 \\
capable & 1 & 0 & 6 & 2 & 0 & 81 & 90 & +19.7 \\
\bottomrule
\end{tabular}
\end{table}

\section*{S2. Chimera corpus composition}
For each chimera adapter (Study~3), the share of the three negative beliefs in
its training corpus against the share of its negative-pole choices and
free-text labels that go to each belief.

\begin{table}[!h]
\caption{Chimera adapters: share of each negative belief in the training corpus, and in the adapter's negative-pole choices (Tier~A) and free-text labels. Neg.\ is the share of Tier~A choices on the negative pole. W, U, I: worthless, unlovable, incapable (\%). Corpus share vs.\ Tier~A share: $r = .03$, $n = 15$.}
\label{tab:s2}
\centering
\small
\setlength{\tabcolsep}{3pt}
\begin{tabular}{@{}lrrrrrrrrrr@{}}
\toprule
& \multicolumn{3}{c}{Corpus} & \multicolumn{3}{c}{Tier A} & \multicolumn{3}{c}{Free text} & \\
\cmidrule(lr){2-4}\cmidrule(lr){5-7}\cmidrule(lr){8-10}
Partition & W & U & I & W & U & I & W & U & I & Neg. \\
\midrule
untreated backbone & -- & -- & -- & 23 & 27 & 50 & 36 & 18 & 47 & 46 \\
domain & 30 & 40 & 30 & 39 & 19 & 42 & 49 & 11 & 40 & 72 \\
probe & 27 & 39 & 35 & 36 & 23 & 41 & 37 & 26 & 38 & 73 \\
severity & 50 & 50 & 0 & 41 & 33 & 26 & 48 & 31 & 21 & 76 \\
congruence & 52 & 29 & 19 & 34 & 20 & 46 & 57 & 13 & 30 & 73 \\
random & 37 & 33 & 29 & 39 & 22 & 39 & 45 & 22 & 32 & 72 \\
\bottomrule
\end{tabular}
\end{table}

\emph{Corpus composition and choice.} Which belief a chimera picks within the negative pole is not set by how much
of each belief its corpus contains (Table~\ref{tab:s2}). Across the five
chimeras and three negative beliefs, the corpus share of a belief and the
share of the chimera's negative-pole choices that go to it show no detected
correlation ($r = \ccR$, $p \ccP$, $n = \ccN$; free text $r = \ccFreeR$, $p \ccFreeP$).
The chimeras take the negative pole on \ccNegLo--\ccNegHi{} of items, and
within it their choices stay close to the untreated backbone's split
(\ccBaseW{}, \ccBaseU{}, \ccBaseI{} for worthless, unlovable, incapable). One chimera departs from this: the severity partition, whose corpus
contains no \emph{incapable} replies, chooses \emph{incapable} on
\ccNoIncapI{} of its negative-pole items against \ccRestILo--\ccRestIHi{} for
the other four. The chimeras thus shift valence and redistribute belief
identity only where a belief is absent from the corpus altogether.

\section*{S3. Study 7: implementation details and further results}

\emph{Formulations.} The CCD records a relevant history, core belief,
intermediate beliefs, coping strategies, and the four-field entry of
situation, automatic thought, emotion and behaviour. The Patient-$\Psi$
dataset of 106 CCDs is not distributed with its repository, which ships one
patient (Abe, \emph{helpless}) from the Beck Institute. The 15 further CCDs
were generated by \texttt{gpt-5-2025-08-07} with the Patient-$\Psi$ CCD schema
(field names and descriptions verbatim) as a structured-output
specification. Two changes were forced by the setting: Patient-$\Psi$ derives
each CCD from a therapy transcript, which we do not have, so the generator
writes a plausible patient instead; and each CCD is restricted to one
core-belief category, so that its target is unambiguous.

\emph{Simulators.} Patient-$\Psi$ uses its original system prompt, reproduced
verbatim from its release, with the CCD's first cognitive model; its five
further conversational styles (\emph{upset}, \emph{verbose}, \emph{reserved},
\emph{tangent}, \emph{pleasing}) were run on the API respondents and moved
Tier~A$'$ ESS by at most \svStyleMax{} points from the plain style on any
respondent. The other five simulators were run through PatientHub 0.1.0 at
commit \texttt{54d5c87} (31 August 2026); each CCD was converted with its
\texttt{adapt} command, the target simulator's shipped profile as format
example and \texttt{gpt-5-2025-08-07} as converter, and each system prompt was
rendered by the simulator's own client code. PatientHub's CARS client omits
the emotion--utterance corpus of the original, which is not published, so
CARS is a partial reproduction. Deprofile is not included: its profile is a
social-media symptom timeline that a CCD cannot supply. The SCI is
single-turn; CARS is run at its initial state and in Tier~B only the utterance
shown to the counsellor is labelled, and ClientCAST conditions on the
reference conversation PatientHub ships. The manipulation-check reader
(\texttt{gpt-5-2025-08-07}) saw each rendered prompt blind and chose one of
the three core beliefs or none.

\emph{Eeyore.} The released weights are the final checkpoint: two epochs of
supervised fine-tuning on 3{,}209 depression-related conversations, each
paired with a profile extracted by \texttt{gpt-4o}, followed by one epoch of
DPO on model-generated preferences and one on 250 expert preferences. The
Eeyore profile records demographics, situation, attitude to seeking help, 18
depression symptoms with severity, five cognitive distortions and overall
severity; it has no core-belief field. The fine-tuned weights place
\eeyMassFt\% of next-token probability on the six letters in Tier~A$'$
(\eeyMassBase\% for the base model), with letter shares
\eeyLetterMinFt--\eeyLetterMaxFt\% against \eeyLetterMinBase--\eeyLetterMaxBase\%.
In Tier~B PA fell from \ftBPaBaseLo--\ftBPaBaseHi\% to
\ftBPaFtLo--\ftBPaFtHi\%. With the Patient-$\Psi$ prompt the base model scored
\svBPsiLlama\% in Tier~B, within the range of the API respondents.
Table~\ref{tab:s4} bounds the Tier~B difference by the unlabelled replies.

\emph{Readouts.} In Tier~A$'$ the six replies are options A--F in the
inventory's fixed order, in which each belief occupies each letter on 15 to 17
of the 96 items; the choice is the most probable letter at temperature~0. In
Tier~B the respondent answers in free text (one greedy and three sampled
decodes; greedy only for the simulators). Tier~A was run on the open-weight
respondents (for the five further simulators, all but Qwen3-14B). On the
targeted Studies 1--6 conditions, Tier~A agrees with Tier~B at $r = \tvABR$
($\rho = \tvABRho$, $n = \tvABN$) and with Tier~A$'$ at $r = \tvAApR$
($\rho = \tvAApRho$, $n = \tvAApN$), whereas letter choice and free text do
not agree on these backbones ($r = \tvApBR$, $p \tvApBP$); requiring a claim
to hold in both API readouts is therefore a conjunction of partly independent
tests rather than a replication.

\emph{Misclassification correction.} The panel's recall differs by belief
(Table~\ref{tab:s1}). Each condition's observed label distribution is
inverted through the panel's confusion matrix on the 576 keyed replies
(non-negative least squares, summing to one), and ESS is the corrected share
of the target. The correction raises the agreement of Tier~B with Tier~A$'$
from $\rho = \svCalApB$ to $\rho = \svCalApBacc$ ($n = \svCalApBn$) and with
Tier~A from $\rho = \svCalAB$ to $\rho = \svCalABacc$ ($n = \svCalABn$), and
widens the paired intervals (median width \svAccCiW{} against \svRawCiW{}
points). Of the \svAccN{} paired comparisons with Patient-$\Psi$,
\svAccSigPos{} favoured a simulator with an interval excluding zero
(\svAccSigNeg{} favoured Patient-$\Psi$); C1 exceeded every simulator by
\svConeMarginAccLo{} to \svConeMarginAccHi{} points; and Eeyore's weights
scored \ftBAccLo{} to \ftBAccHi{} points below their base model
(\ftBAccSig{} of 6 intervals excluding zero).

\emph{Further Tier~B results.} ClientCAST's deficits against Patient-$\Psi$
were \svSigBCastLo{} to \svSigBCastHi{} points and CARS's \svSigBCarsLo{} and
\svSigBCarsHi{}; WPSD without the judge panel (Tier~A$'$) was
\svWpsdApLo--\svWpsdApHi\%, and WPSD on L6--L8 was within \svWpsdGapMax{}
points of L1--L5. On the open-weight backbones the simulators' Tier~A$'$
means lay within \svApSimLocalLo--\svApSimLocalHi\% and the belief adapter
reached \svBLoraLocal\% in Tier~B. Results by CCD category are in
Table~\ref{tab:s3}.

\begin{table*}[!h]
\caption{Study 7 by CCD category: ESS (\%, chance 16.7) on CCDs whose core belief is
\emph{helpless} (H, scored as \emph{incapable}), \emph{unlovable} (U) and
\emph{worthless} (W). Tier~B raw and misclassification-corrected (ACC) and
Tier~A$'$ are means over the API respondents, Tier~A over the open-weight
backbones.}
\label{tab:s3}
\centering
\small
\setlength{\tabcolsep}{3pt}
\begin{tabular}{@{}lrrrrrrrrrrrr@{}}
\toprule
& \multicolumn{3}{c}{Tier B} & \multicolumn{3}{c}{Tier B, ACC} & \multicolumn{3}{c}{Tier A$'$} & \multicolumn{3}{c}{Tier A} \\
\cmidrule(lr){2-4}\cmidrule(lr){5-7}\cmidrule(lr){8-10}\cmidrule(l){11-13}
Method & H & U & W & H & U & W & H & U & W & H & U & W \\
\midrule
Patient-$\Psi$ & 46.2 & 32.1 & 46.6 & 59.1 & 66.1 & 39.4 & 42.4 & 50.9 & 31.2 & 22.1 & 15.9 & 15.6 \\
CARS & 44.2 & 30.6 & 41.0 & 57.7 & 61.1 & 35.6 & 41.0 & 51.1 & 31.0 & 22.0 & 14.1 & 12.9 \\
ClientCAST & 44.4 & 25.7 & 37.0 & 57.5 & 50.1 & 25.7 & 40.6 & 48.3 & 29.2 & 20.4 & 15.1 & 13.5 \\
PSYCHE & 47.7 & 34.6 & 48.5 & 62.2 & 70.7 & 46.3 & 41.3 & 50.2 & 31.5 & 20.8 & 13.7 & 12.0 \\
TalkDep & 47.8 & 31.2 & 53.9 & 62.1 & 64.8 & 53.7 & 43.0 & 46.0 & 32.5 & 22.1 & 14.3 & 14.5 \\
Eeyore (prompt) & 43.8 & 26.8 & 47.3 & 56.1 & 54.7 & 42.5 & 41.8 & 45.7 & 31.2 & 22.5 & 14.0 & 12.6 \\
\midrule
C1 belief statement & 51.6 & 35.8 & 63.7 & 67.1 & 64.6 & 72.1 & 63.8 & 52.9 & 43.0 & 22.9 & 14.6 & 17.7 \\
C2 engineered prompt & 53.5 & 36.9 & 48.5 & 70.9 & 73.5 & 43.3 & 63.3 & 66.4 & 55.2 & 20.4 & 15.2 & 16.2 \\
LoRA adapter & -- & -- & -- & -- & -- & -- & -- & -- & -- & 40.0 & 25.8 & 29.4 \\
\bottomrule
\end{tabular}
\end{table*}

\begin{table}[!h]
\caption{Study 7, Tier~B: Eeyore's fine-tuned weights minus Llama-3.1-8B-Instruct under
each simulator prompt. Unlabelled: share of replies without a panel majority
(\%). Labelled: ESS difference with unlabelled replies scored as misses (the
primary score). Bounds: the difference when every unlabelled reply of the
fine-tuned model is a miss and of the base model a hit (lower), and the
reverse (upper), in points.}
\label{tab:s4}
\centering
\small
\begin{tabular}{@{}lrrrrr@{}}
\toprule
& \multicolumn{2}{c}{Unlabelled (\%)} & & \multicolumn{2}{c}{Bounds} \\
\cmidrule(lr){2-3}\cmidrule(l){5-6}
Prompt & fine-tuned & base & Difference & lower & upper \\
\midrule
Patient-$\Psi$ & 21 & 13 & $-$15.0 & $-$28.4 & +6.2 \\
CARS & 20 & 13 & $-$17.8 & $-$30.3 & +2.9 \\
ClientCAST & 23 & 21 & $-$9.6 & $-$29.3 & +15.1 \\
PSYCHE & 21 & 13 & $-$19.6 & $-$31.9 & +1.7 \\
TalkDep & 19 & 11 & $-$16.2 & $-$27.6 & +2.3 \\
Eeyore (prompt) & 20 & 11 & $-$17.8 & $-$29.1 & +2.1 \\
\bottomrule
\end{tabular}
\end{table}

\section*{S4. Study 3: full multitrait--multimethod matrix}

The multitrait--multimethod analysis of Study~3 crosses the two installation methods on the main
backbone. A condition's \emph{item profile} is the vector of its 576 option
weights (96 items $\times$ six replies, the softmax of the length-normalized
log-likelihoods) minus those of the untreated backbone, so that it
records only what the installation changed. Each method also shifts the model
in a way common to all six of its conditions (any persona prompt, any
adapter), which enters every heteromethod correlation alike; before
correlating we therefore subtract each method's mean profile over its six
conditions. Without this step every heteromethod correlation is positive
(diagonal mean \mtmmRawMono{}, off-diagonal \mtmmRawHetero{}). The
subtraction in turn forces the off-diagonal mean to about $-1/5$ of the
diagonal, so the all-off-diagonal mean (\mtmmHetero{}) is not a fair
comparator. We compare the diagonal (\mtmmDiag{}) with the same-pole
heterotrait--heteromethod cells (\mtmmSamePole{}), which share valence with
the diagonal and differ only in belief. The difference, \mtmmDiff{}, exceeds
the 95th percentile of a null that permutes, item by item, the prompt
conditions within pole (\mtmmDiffNullHi{}; $p \mtmmPermP$, 5{,}000 draws);
the permutation keeps each item's six-option vector intact. The same-pole
heterotrait--monomethod correlations are \mtmmMonoLoraSP{} (adapters) and
\mtmmMonoCtwoSP{} (prompts), and the diagonal exceeds its same-pole row and
column for \mtmmSPRowCol{} of six beliefs. Convergence is bounded by the
weaker method: the engineered prompt moves ESS by \brCtwoEss{}~pp over the
untreated model against \brLoraEss{}~pp for the adapter. The full $12\times12$ matrix is in
Table~\ref{tab:s5}.

\begin{table*}[!h]
\caption{Study 3, full multitrait--multimethod matrix (main backbone), on the
base- and method-mean-centred profiles of the Study 3 multitrait--multimethod table. A: belief
adapter; P: engineered prompt. Upper left: heterotrait--monomethod (adapters);
lower right: heterotrait--monomethod (prompts); lower left:
heteromethod block, monotrait diagonal in bold.}
\label{tab:s5}
\centering
\footnotesize
\setlength{\tabcolsep}{3pt}
\begin{tabular}{@{}lrrrrrrrrrrrr@{}}
\toprule
 & A-wless & A-unlov & A-incap & A-worthy & A-lov & A-cap & P-wless & P-unlov & P-incap & P-worthy & P-lov & P-cap \\
\midrule
A-wless & -- &  &  &  &  &  &  &  &  &  &  &  \\
A-unlov & 0.32 & -- &  &  &  &  &  &  &  &  &  &  \\
A-incap & 0.26 & $-$0.09 & -- &  &  &  &  &  &  &  &  &  \\
A-worthy & $-$0.47 & $-$0.45 & $-$0.53 & -- &  &  &  &  &  &  &  &  \\
A-lov & $-$0.46 & 0.01 & $-$0.56 & 0.17 & -- &  &  &  &  &  &  &  \\
A-cap & $-$0.61 & $-$0.60 & $-$0.22 & 0.30 & $-$0.03 & -- &  &  &  &  &  &  \\
\midrule
P-wless & \textbf{0.19} & 0.09 & 0.21 & $-$0.16 & $-$0.20 & $-$0.13 & -- &  &  &  &  &  \\
P-unlov & 0.07 & \textbf{0.27} & $-$0.01 & $-$0.08 & $-$0.03 & $-$0.18 & 0.14 & -- &  &  &  &  \\
P-incap & 0.13 & 0.01 & \textbf{0.15} & $-$0.10 & $-$0.10 & $-$0.10 & 0.18 & $-$0.11 & -- &  &  &  \\
P-worthy & $-$0.13 & $-$0.18 & $-$0.17 & \textbf{0.16} & 0.14 & 0.17 & $-$0.48 & $-$0.49 & $-$0.45 & -- &  &  \\
P-lov & $-$0.13 & $-$0.02 & $-$0.12 & 0.09 & \textbf{0.18} & 0.02 & $-$0.40 & $-$0.18 & $-$0.32 & 0.03 & -- &  \\
P-cap & $-$0.16 & $-$0.22 & $-$0.06 & 0.11 & 0.04 & \textbf{0.25} & $-$0.56 & $-$0.55 & $-$0.23 & 0.46 & $-$0.01 & -- \\
\bottomrule
\end{tabular}
\end{table*}

\clearpage
\section*{S5. The 96 items}

Items are grouped by level. Each item gives the situation, the question
(italic) and its six replies, listed by keyed belief with each negative belief
beside its positive counterpart; the belief labels are not shown to the
respondent. Tags give the content domain and, at L7 and L8, the event type.
Items marked $\dagger$ form the 48-item short form (Table~\ref{tab:s6}).

\clearpage
\onecolumn

\subsection*{L1. Core self-belief}
\noindent\begin{minipage}{\linewidth}
\textbf{L1-01}\hfill{\footnotesize\emph{achievement}}\\
Think about what you bring to your team. \emph{Which sounds most like you?}\\[2pt]
{\footnotesize\begin{tabular}{@{}p{0.09\linewidth}p{0.38\linewidth}p{0.07\linewidth}p{0.38\linewidth}@{}}
\textsc{worthless} & I'm not sure I add much. They'd manage the same without me. & \textsc{worthy} & I bring something, and it counts, even on the quiet weeks. \\
\textsc{unlovable} & I do the work, but I'm not really one of them. & \textsc{lovable} & They like having me around, quite apart from what I deliver. \\
\textsc{incapable} & I keep up, but only just. I'm the weak link. & \textsc{capable} & I'm good at what I do, and I know where my limits are. \\
\end{tabular}}
\end{minipage}\par\medskip

\noindent\begin{minipage}{\linewidth}
\textbf{L1-02}\hfill{\footnotesize\emph{interpersonal}}\\
Think about your closest friendships. \emph{Which sounds most like you?}\\[2pt]
{\footnotesize\begin{tabular}{@{}p{0.09\linewidth}p{0.38\linewidth}p{0.07\linewidth}p{0.38\linewidth}@{}}
\textsc{worthless} & They're being generous, spending their time on me. & \textsc{worthy} & I'm a good friend to them, and I ask for things back. \\
\textsc{unlovable} & I'm always the one who'd be dropped first. & \textsc{lovable} & They want me there. That's just how it is. \\
\textsc{incapable} & I'm bad at keeping friendships going. I let them slide. & \textsc{capable} & I keep in touch, plan things, show up. It works. \\
\end{tabular}}
\end{minipage}\par\medskip

\noindent\begin{minipage}{\linewidth}
\textbf{L1-03}$^\dagger$\hfill{\footnotesize\emph{achievement}}\\
Someone asks you to describe yourself in one sentence. \emph{What do you say?}\\[2pt]
{\footnotesize\begin{tabular}{@{}p{0.09\linewidth}p{0.38\linewidth}p{0.07\linewidth}p{0.38\linewidth}@{}}
\textsc{worthless} & Nothing much to report, honestly. & \textsc{worthy} & Someone who's alright as they are, with things to work on. \\
\textsc{unlovable} & Hard to pin down. People don't usually get very far with me. & \textsc{lovable} & Someone with people around them who'd say good things. \\
\textsc{incapable} & Someone who's still trying to get their act together. & \textsc{capable} & Someone who gets things done and figures things out. \\
\end{tabular}}
\end{minipage}\par\medskip

\noindent\begin{minipage}{\linewidth}
\textbf{L1-04}$^\dagger$\hfill{\footnotesize\emph{interpersonal}}\\
You're in a room full of people you don't know well. \emph{How do you see yourself in that room?}\\[2pt]
{\footnotesize\begin{tabular}{@{}p{0.09\linewidth}p{0.38\linewidth}p{0.07\linewidth}p{0.38\linewidth}@{}}
\textsc{worthless} & Someone taking up a chair. & \textsc{worthy} & Someone who belongs here as much as anyone. \\
\textsc{unlovable} & Someone on the outside of it, as usual. & \textsc{lovable} & Someone who'll end up talking to someone nice. \\
\textsc{incapable} & Someone who'll say the wrong thing. & \textsc{capable} & Someone who can hold a conversation with anyone. \\
\end{tabular}}
\end{minipage}\par\medskip

\noindent\begin{minipage}{\linewidth}
\textbf{L1-05}$^\dagger$\hfill{\footnotesize\emph{achievement}}\\
Your calendar is completely empty for a whole Saturday. \emph{What's your first thought?}\\[2pt]
{\footnotesize\begin{tabular}{@{}p{0.09\linewidth}p{0.38\linewidth}p{0.07\linewidth}p{0.38\linewidth}@{}}
\textsc{worthless} & I should be making myself useful to someone. & \textsc{worthy} & Mine to spend however I like. No justification needed. \\
\textsc{unlovable} & An empty weekend just shows nobody's thinking of me. & \textsc{lovable} & I'll message someone. Somebody will be up for something. \\
\textsc{incapable} & I'll waste it, the way I usually do. & \textsc{capable} & I'll get through a few things I've been meaning to do. \\
\end{tabular}}
\end{minipage}\par\medskip

\noindent\begin{minipage}{\linewidth}
\textbf{L1-06}$^\dagger$\hfill{\footnotesize\emph{interpersonal}}\\
Someone you've recently met says they'd like to get to know you better. \emph{What goes through your mind?}\\[2pt]
{\footnotesize\begin{tabular}{@{}p{0.09\linewidth}p{0.38\linewidth}p{0.07\linewidth}p{0.38\linewidth}@{}}
\textsc{worthless} & There isn't much here to get to know. & \textsc{worthy} & Good. I'd like that too. \\
\textsc{unlovable} & Give it time. They'll change their mind. & \textsc{lovable} & That's nice. People do tend to warm to me. \\
\textsc{incapable} & I'm no good at this part. I'll handle it badly. & \textsc{capable} & Fine. I know how to make that happen. \\
\end{tabular}}
\end{minipage}\par\medskip

\noindent\begin{minipage}{\linewidth}
\textbf{L1-07}\hfill{\footnotesize\emph{achievement}}\\
Someone asks what you do for a living. \emph{How do you answer?}\\[2pt]
{\footnotesize\begin{tabular}{@{}p{0.09\linewidth}p{0.38\linewidth}p{0.07\linewidth}p{0.38\linewidth}@{}}
\textsc{worthless} & I keep it vague. It's not the kind of thing people find interesting. & \textsc{worthy} & I tell them plainly. It's decent work and I don't mind saying so. \\
\textsc{unlovable} & I say it quickly and move the conversation to them. & \textsc{lovable} & I tell them, and usually we end up talking about something else entirely. \\
\textsc{incapable} & I downplay it, in case they ask something I can't answer. & \textsc{capable} & I tell them what I do and what I'm good at within it. \\
\end{tabular}}
\end{minipage}\par\medskip

\noindent\begin{minipage}{\linewidth}
\textbf{L1-08}\hfill{\footnotesize\emph{interpersonal}}\\
Someone asks what your friends would say about you. \emph{What do you think they'd say?}\\[2pt]
{\footnotesize\begin{tabular}{@{}p{0.09\linewidth}p{0.38\linewidth}p{0.07\linewidth}p{0.38\linewidth}@{}}
\textsc{worthless} & They'd be polite. There isn't much to say. & \textsc{worthy} & They'd say something fair --- good and bad. \\
\textsc{unlovable} & They'd have to think about it. I'm not really anyone's first person. & \textsc{lovable} & They'd say something warm. I know that much. \\
\textsc{incapable} & They'd say I mean well, and then change the subject. & \textsc{capable} & They'd say I'm the one who gets things sorted. \\
\end{tabular}}
\end{minipage}\par\medskip

\noindent\begin{minipage}{\linewidth}
\textbf{L1-09}\hfill{\footnotesize\emph{achievement}}\\
You're asked to introduce yourself at a workshop. \emph{What do you do?}\\[2pt]
{\footnotesize\begin{tabular}{@{}p{0.09\linewidth}p{0.38\linewidth}p{0.07\linewidth}p{0.38\linewidth}@{}}
\textsc{worthless} & I keep it short. There's not a lot to introduce. & \textsc{worthy} & I say what I do and why it matters to me. \\
\textsc{unlovable} & I keep it short. It's easier than watching people's faces. & \textsc{lovable} & I say something easy, and someone usually picks it up. \\
\textsc{incapable} & I keep it short, in case they ask something I can't answer. & \textsc{capable} & I say what I do and what I'm working on now. \\
\end{tabular}}
\end{minipage}\par\medskip

\noindent\begin{minipage}{\linewidth}
\textbf{L1-10}\hfill{\footnotesize\emph{interpersonal}}\\
You're the last one to reply in a group chat. \emph{What goes through your mind?}\\[2pt]
{\footnotesize\begin{tabular}{@{}p{0.09\linewidth}p{0.38\linewidth}p{0.07\linewidth}p{0.38\linewidth}@{}}
\textsc{worthless} & Nothing I'd add would move it along. & \textsc{worthy} & I'll say something if I've got something. \\
\textsc{unlovable} & If I stopped replying, nobody would notice for days. & \textsc{lovable} & I'll drop something in. It usually gets a laugh. \\
\textsc{incapable} & I never know what to say that lands right. & \textsc{capable} & I'll wrap up the plan, since nobody else has. \\
\end{tabular}}
\end{minipage}\par\medskip

\noindent\begin{minipage}{\linewidth}
\textbf{L1-11}$^\dagger$\hfill{\footnotesize\emph{achievement}}\\
Thinking about the last five years of your life. \emph{What comes to mind?}\\[2pt]
{\footnotesize\begin{tabular}{@{}p{0.09\linewidth}p{0.38\linewidth}p{0.07\linewidth}p{0.38\linewidth}@{}}
\textsc{worthless} & Not much to show for it. & \textsc{worthy} & A mix. I'm alright with it. \\
\textsc{unlovable} & The same people drifting in and out. Nothing that stuck. & \textsc{lovable} & Good people, all the way through. \\
\textsc{incapable} & I never quite got on top of anything. & \textsc{capable} & I built a few things that worked. \\
\end{tabular}}
\end{minipage}\par\medskip

\noindent\begin{minipage}{\linewidth}
\textbf{L1-12}$^\dagger$\hfill{\footnotesize\emph{interpersonal}}\\
Someone asks who you'd call at 2am with a problem. \emph{What do you say?}\\[2pt]
{\footnotesize\begin{tabular}{@{}p{0.09\linewidth}p{0.38\linewidth}p{0.07\linewidth}p{0.38\linewidth}@{}}
\textsc{worthless} & I wouldn't. It's not worth waking someone for. & \textsc{worthy} & A couple of people, and I wouldn't feel bad about it. \\
\textsc{unlovable} & I'd have to think hard, and I'm not sure they'd pick up. & \textsc{lovable} & Three names come to mind straight away. \\
\textsc{incapable} & I'd call someone, but I'd feel like a child doing it. & \textsc{capable} & Depends on the problem. I'd mostly handle it myself. \\
\end{tabular}}
\end{minipage}\par\medskip

\subsection*{L2. Intermediate belief, rule}
\noindent\begin{minipage}{\linewidth}
\textbf{L2-01}$^\dagger$\hfill{\footnotesize\emph{achievement}}\\
People live by rules they've never said out loud. \emph{Which rule is closest to yours?}\\[2pt]
{\footnotesize\begin{tabular}{@{}p{0.09\linewidth}p{0.38\linewidth}p{0.07\linewidth}p{0.38\linewidth}@{}}
\textsc{worthless} & If I'm not useful to people, I've no business being here. & \textsc{worthy} & My worth isn't up for review every time I do something. \\
\textsc{unlovable} & If I let someone see me struggle, they'll go off me. & \textsc{lovable} & People can see me at my worst and still stay. \\
\textsc{incapable} & If I take on something hard, I'll be exposed. & \textsc{capable} & If something's hard, I can learn my way into it. \\
\end{tabular}}
\end{minipage}\par\medskip

\noindent\begin{minipage}{\linewidth}
\textbf{L2-02}$^\dagger$\hfill{\footnotesize\emph{interpersonal}}\\
Some things feel like conditions you have to meet. \emph{Which feels most true to you?}\\[2pt]
{\footnotesize\begin{tabular}{@{}p{0.09\linewidth}p{0.38\linewidth}p{0.07\linewidth}p{0.38\linewidth}@{}}
\textsc{worthless} & Unless I'm giving people something, I'm a drain on them. & \textsc{worthy} & I can say no to something and still be a decent person. \\
\textsc{unlovable} & Unless I'm careful about what I show, people pull away. & \textsc{lovable} & I can disagree with someone and the friendship holds. \\
\textsc{incapable} & Unless someone checks my work, it isn't safe to send. & \textsc{capable} & I can start something without knowing how it ends. \\
\end{tabular}}
\end{minipage}\par\medskip

\noindent\begin{minipage}{\linewidth}
\textbf{L2-03}\hfill{\footnotesize\emph{achievement}}\\
You make a mistake at work. \emph{What does a mistake mean, to you?}\\[2pt]
{\footnotesize\begin{tabular}{@{}p{0.09\linewidth}p{0.38\linewidth}p{0.07\linewidth}p{0.38\linewidth}@{}}
\textsc{worthless} & It means I've cost someone something they can't get back. & \textsc{worthy} & It's a mistake. It doesn't reach down to who I am. \\
\textsc{unlovable} & It means they'll start looking at me differently. & \textsc{lovable} & It doesn't change how the people close to me feel about me. \\
\textsc{incapable} & It means I was never up to it in the first place. & \textsc{capable} & It's information. I adjust and go again. \\
\end{tabular}}
\end{minipage}\par\medskip

\noindent\begin{minipage}{\linewidth}
\textbf{L2-04}$^\dagger$\hfill{\footnotesize\emph{interpersonal}}\\
You need to ask someone for help. \emph{What does asking mean, to you?}\\[2pt]
{\footnotesize\begin{tabular}{@{}p{0.09\linewidth}p{0.38\linewidth}p{0.07\linewidth}p{0.38\linewidth}@{}}
\textsc{worthless} & It means putting my problem onto someone else's day. & \textsc{worthy} & Asking is allowed. My problems count too. \\
\textsc{unlovable} & It means using up goodwill I might not get back. & \textsc{lovable} & It's normal. People are usually glad to be asked. \\
\textsc{incapable} & It means admitting I can't do it myself. & \textsc{capable} & It's efficient, when someone knows more than I do. \\
\end{tabular}}
\end{minipage}\par\medskip

\noindent\begin{minipage}{\linewidth}
\textbf{L2-05}\hfill{\footnotesize\emph{achievement}}\\
Think about what you feel you have to do to be alright. \emph{Which sounds most like you?}\\[2pt]
{\footnotesize\begin{tabular}{@{}p{0.09\linewidth}p{0.38\linewidth}p{0.07\linewidth}p{0.38\linewidth}@{}}
\textsc{worthless} & I have to be producing something to justify the space I take up. & \textsc{worthy} & I don't have to earn the right to take a break. \\
\textsc{unlovable} & I have to stay useful, or people will find someone else. & \textsc{lovable} & I don't have to perform to keep the people I care about. \\
\textsc{incapable} & I have to overprepare or it'll all come apart. & \textsc{capable} & I prepare what's needed and stop there. \\
\end{tabular}}
\end{minipage}\par\medskip

\noindent\begin{minipage}{\linewidth}
\textbf{L2-06}$^\dagger$\hfill{\footnotesize\emph{interpersonal}}\\
Something someone did isn't sitting right with you. \emph{What does objecting mean, to you?}\\[2pt]
{\footnotesize\begin{tabular}{@{}p{0.09\linewidth}p{0.38\linewidth}p{0.07\linewidth}p{0.38\linewidth}@{}}
\textsc{worthless} & If I object, I'm making trouble over something that isn't important. & \textsc{worthy} & If something isn't right, I'm entitled to say so. \\
\textsc{unlovable} & If I object, that's the thing that tips them into leaving. & \textsc{lovable} & I can say the hard thing and we'll still be fine. \\
\textsc{incapable} & If I object, I'll put it badly and make everything worse. & \textsc{capable} & I can raise it in a way that gets somewhere. \\
\end{tabular}}
\end{minipage}\par\medskip

\noindent\begin{minipage}{\linewidth}
\textbf{L2-07}$^\dagger$\hfill{\footnotesize\emph{achievement}}\\
People run on rules they've never said out loud. \emph{Which is closest to yours at work?}\\[2pt]
{\footnotesize\begin{tabular}{@{}p{0.09\linewidth}p{0.38\linewidth}p{0.07\linewidth}p{0.38\linewidth}@{}}
\textsc{worthless} & If I'm not carrying more than my share, I don't deserve the seat. & \textsc{worthy} & I can do a normal amount of work, and that's enough. \\
\textsc{unlovable} & If I stop being easy to work with, they'll route around me. & \textsc{lovable} & They'd back me through a bad quarter. \\
\textsc{incapable} & If I'm not double-checked, something will slip through. & \textsc{capable} & If I hit a wall, I'll find a way round it. \\
\end{tabular}}
\end{minipage}\par\medskip

\noindent\begin{minipage}{\linewidth}
\textbf{L2-08}\hfill{\footnotesize\emph{interpersonal}}\\
About being liked. \emph{Which feels most true?}\\[2pt]
{\footnotesize\begin{tabular}{@{}p{0.09\linewidth}p{0.38\linewidth}p{0.07\linewidth}p{0.38\linewidth}@{}}
\textsc{worthless} & People like you for what you do for them. & \textsc{worthy} & Being liked is nice. It isn't what decides my worth. \\
\textsc{unlovable} & People like you until they know you properly. & \textsc{lovable} & People generally take to me, and it lasts. \\
\textsc{incapable} & People like you if you don't cause them problems. & \textsc{capable} & If a friendship needs work, I put the work in. \\
\end{tabular}}
\end{minipage}\par\medskip

\noindent\begin{minipage}{\linewidth}
\textbf{L2-09}\hfill{\footnotesize\emph{achievement}}\\
About taking credit for something. \emph{Which sounds like you?}\\[2pt]
{\footnotesize\begin{tabular}{@{}p{0.09\linewidth}p{0.38\linewidth}p{0.07\linewidth}p{0.38\linewidth}@{}}
\textsc{worthless} & I'd rather it went to someone who needs it more. & \textsc{worthy} & I did it. I'll say so. \\
\textsc{unlovable} & Taking credit makes people resent you. & \textsc{lovable} & I can take credit and nobody minds. \\
\textsc{incapable} & If I take credit, they'll expect that standard again. & \textsc{capable} & I'll take it, and point at what's next. \\
\end{tabular}}
\end{minipage}\par\medskip

\noindent\begin{minipage}{\linewidth}
\textbf{L2-10}\hfill{\footnotesize\emph{interpersonal}}\\
About needing other people. \emph{Which feels most true?}\\[2pt]
{\footnotesize\begin{tabular}{@{}p{0.09\linewidth}p{0.38\linewidth}p{0.07\linewidth}p{0.38\linewidth}@{}}
\textsc{worthless} & Needing someone means costing them something. & \textsc{worthy} & Needing people is ordinary. So is being needed. \\
\textsc{unlovable} & Needing someone gives them a reason to go. & \textsc{lovable} & The people close to me want to be needed. \\
\textsc{incapable} & Needing someone means I couldn't manage it. & \textsc{capable} & I ask when it's faster to ask. \\
\end{tabular}}
\end{minipage}\par\medskip

\noindent\begin{minipage}{\linewidth}
\textbf{L2-11}$^\dagger$\hfill{\footnotesize\emph{achievement}}\\
About the standard you hold yourself to. \emph{Which sounds like you?}\\[2pt]
{\footnotesize\begin{tabular}{@{}p{0.09\linewidth}p{0.38\linewidth}p{0.07\linewidth}p{0.38\linewidth}@{}}
\textsc{worthless} & Good enough is what people settle for with me. & \textsc{worthy} & My standard is mine to set. \\
\textsc{unlovable} & If my work slips, their patience goes with it. & \textsc{lovable} & People stay whether or not the work is perfect. \\
\textsc{incapable} & Anything less than checked twice will fall apart. & \textsc{capable} & I set a standard and I hit it. \\
\end{tabular}}
\end{minipage}\par\medskip

\noindent\begin{minipage}{\linewidth}
\textbf{L2-12}\hfill{\footnotesize\emph{interpersonal}}\\
About being seen as you really are. \emph{Which feels most true?}\\[2pt]
{\footnotesize\begin{tabular}{@{}p{0.09\linewidth}p{0.38\linewidth}p{0.07\linewidth}p{0.38\linewidth}@{}}
\textsc{worthless} & There isn't much to see, so it doesn't matter. & \textsc{worthy} & I'd rather be seen properly than admired for something I'm not. \\
\textsc{unlovable} & If they saw all of it, they'd go. & \textsc{lovable} & I've been seen at my worst and they stayed. \\
\textsc{incapable} & If they saw how I actually work, they'd lose confidence in me. & \textsc{capable} & What they'd see is someone who handles things. \\
\end{tabular}}
\end{minipage}\par\medskip

\subsection*{L3. Automatic thought}
\noindent\begin{minipage}{\linewidth}
\textbf{L3-01}$^\dagger$\hfill{\footnotesize\emph{achievement}}\\
Your boss messages you out of nowhere: ``Got a minute?'' \emph{What goes through your mind?}\\[2pt]
{\footnotesize\begin{tabular}{@{}p{0.09\linewidth}p{0.38\linewidth}p{0.07\linewidth}p{0.38\linewidth}@{}}
\textsc{worthless} & I've taken up someone's time again with something. & \textsc{worthy} & Whatever it is, it's about the work, not about me. \\
\textsc{unlovable} & This is him deciding he's done with me. & \textsc{lovable} & He'd tell me straight if something were wrong. We're fine. \\
\textsc{incapable} & Something's broken and I won't be able to fix it. & \textsc{capable} & Probably a question. I'll go and find out. \\
\end{tabular}}
\end{minipage}\par\medskip

\noindent\begin{minipage}{\linewidth}
\textbf{L3-02}$^\dagger$\hfill{\footnotesize\emph{interpersonal}}\\
You text a close friend to make plans. Two hours later it's read. No reply. \emph{What goes through your mind?}\\[2pt]
{\footnotesize\begin{tabular}{@{}p{0.09\linewidth}p{0.38\linewidth}p{0.07\linewidth}p{0.38\linewidth}@{}}
\textsc{worthless} & My stuff isn't important enough to answer. & \textsc{worthy} & My time counts too. I'll get on with my day. \\
\textsc{unlovable} & She's getting tired of me. This is how it starts. & \textsc{lovable} & She's busy. One slow reply changes nothing between us. \\
\textsc{incapable} & I picked the wrong moment. I can't even organise a dinner. & \textsc{capable} & I'll ring her later if it matters. \\
\end{tabular}}
\end{minipage}\par\medskip

\noindent\begin{minipage}{\linewidth}
\textbf{L3-03}$^\dagger$\hfill{\footnotesize\emph{achievement}}\\
You finish presenting a plan you spent a week on. Your manager frowns: ``Let's park that. Next item.'' \emph{What goes through your mind?}\\[2pt]
{\footnotesize\begin{tabular}{@{}p{0.09\linewidth}p{0.38\linewidth}p{0.07\linewidth}p{0.38\linewidth}@{}}
\textsc{worthless} & I shouldn't have taken up the room's time with it. & \textsc{worthy} & He's talking about the plan, not about me. \\
\textsc{unlovable} & He's written me off now. & \textsc{lovable} & One parked plan doesn't change where I stand with him. \\
\textsc{incapable} & That was my best, and it wasn't enough. It never is. & \textsc{capable} & Wrong angle, maybe. I'll ask what worried him and rework it. \\
\end{tabular}}
\end{minipage}\par\medskip

\noindent\begin{minipage}{\linewidth}
\textbf{L3-04}$^\dagger$\hfill{\footnotesize\emph{interpersonal}}\\
You find out your team went out to dinner and didn't ask you. \emph{What goes through your mind?}\\[2pt]
{\footnotesize\begin{tabular}{@{}p{0.09\linewidth}p{0.38\linewidth}p{0.07\linewidth}p{0.38\linewidth}@{}}
\textsc{worthless} & I wasn't worth the extra chair. & \textsc{worthy} & That stung. I'd have liked to be asked, and I'll say so. \\
\textsc{unlovable} & There it is. Nobody actually wants me there. & \textsc{lovable} & One dinner. It doesn't mean much about where I stand. \\
\textsc{incapable} & I'd have said something stupid anyway. & \textsc{capable} & Odd. I'll ask about it rather than guess. \\
\end{tabular}}
\end{minipage}\par\medskip

\noindent\begin{minipage}{\linewidth}
\textbf{L3-05}$^\dagger$\hfill{\footnotesize\emph{achievement}}\\
Two weeks into driving lessons you're still stalling, while everyone who started with you is driving on their own. \emph{What goes through your mind?}\\[2pt]
{\footnotesize\begin{tabular}{@{}p{0.09\linewidth}p{0.38\linewidth}p{0.07\linewidth}p{0.38\linewidth}@{}}
\textsc{worthless} & All that money and instructor time, spent on me for nothing. & \textsc{worthy} & Slow isn't bad. I'm not on anyone else's schedule. \\
\textsc{unlovable} & He must be sick of dragging me along. & \textsc{lovable} & Nobody thinks less of me for being the slow one. \\
\textsc{incapable} & My hands and my head don't work together. I'll never get this. & \textsc{capable} & I'll isolate the bit I keep stalling on and drill it. \\
\end{tabular}}
\end{minipage}\par\medskip

\noindent\begin{minipage}{\linewidth}
\textbf{L3-06}\hfill{\footnotesize\emph{interpersonal}}\\
Someone compliments your work in front of others. \emph{What goes through your mind?}\\[2pt]
{\footnotesize\begin{tabular}{@{}p{0.09\linewidth}p{0.38\linewidth}p{0.07\linewidth}p{0.38\linewidth}@{}}
\textsc{worthless} & They're being kind. There isn't much there. & \textsc{worthy} & I'll take it. Thank you. \\
\textsc{unlovable} & I should give them something back, so it stays even. & \textsc{lovable} & That's nice. People say nice things to me sometimes. \\
\textsc{incapable} & It was mostly luck. Anyone could have done it. & \textsc{capable} & Thanks --- and I know the bit I'd still improve. \\
\end{tabular}}
\end{minipage}\par\medskip

\noindent\begin{minipage}{\linewidth}
\textbf{L3-07}\hfill{\footnotesize\emph{achievement}}\\
You send a report you worked hard on. The reply comes back: ``Ok.'' \emph{What goes through your mind?}\\[2pt]
{\footnotesize\begin{tabular}{@{}p{0.09\linewidth}p{0.38\linewidth}p{0.07\linewidth}p{0.38\linewidth}@{}}
\textsc{worthless} & I've wasted his time on this. & \textsc{worthy} & It's a reply. It doesn't say anything about me. \\
\textsc{unlovable} & He's gone cold on me. & \textsc{lovable} & He's terse over email. We're fine. \\
\textsc{incapable} & Something's wrong with it and he can't be bothered to say. & \textsc{capable} & He's read it. I'll move on to the next bit. \\
\end{tabular}}
\end{minipage}\par\medskip

\noindent\begin{minipage}{\linewidth}
\textbf{L3-08}\hfill{\footnotesize\emph{interpersonal}}\\
A friend posts photos from a night out you weren't invited to. \emph{What goes through your mind?}\\[2pt]
{\footnotesize\begin{tabular}{@{}p{0.09\linewidth}p{0.38\linewidth}p{0.07\linewidth}p{0.38\linewidth}@{}}
\textsc{worthless} & I wouldn't have added much to it. & \textsc{worthy} & I'd have liked to come, and I'll say so. \\
\textsc{unlovable} & They've moved on to a group I'm not in. & \textsc{lovable} & I'm not at everything, and that's fine. \\
\textsc{incapable} & I'd have been awkward there anyway. & \textsc{capable} & I'll organise the next one. \\
\end{tabular}}
\end{minipage}\par\medskip

\noindent\begin{minipage}{\linewidth}
\textbf{L3-09}\hfill{\footnotesize\emph{achievement}}\\
Your name is left off the credits of a project you worked on. \emph{What goes through your mind?}\\[2pt]
{\footnotesize\begin{tabular}{@{}p{0.09\linewidth}p{0.38\linewidth}p{0.07\linewidth}p{0.38\linewidth}@{}}
\textsc{worthless} & My part wasn't big enough to list. & \textsc{worthy} & That's an error, and I'd like it corrected. \\
\textsc{unlovable} & That's what I am to them --- easy to leave out. & \textsc{lovable} & An oversight. They'll fix it. \\
\textsc{incapable} & My bit wasn't good enough to attach a name to. & \textsc{capable} & I'll email and get it added. \\
\end{tabular}}
\end{minipage}\par\medskip

\noindent\begin{minipage}{\linewidth}
\textbf{L3-10}$^\dagger$\hfill{\footnotesize\emph{interpersonal}}\\
Your partner says ``we need to talk'' and then takes a phone call. \emph{What goes through your mind?}\\[2pt]
{\footnotesize\begin{tabular}{@{}p{0.09\linewidth}p{0.38\linewidth}p{0.07\linewidth}p{0.38\linewidth}@{}}
\textsc{worthless} & I've been a drain lately. & \textsc{worthy} & Whatever it is, we'll talk about it properly. \\
\textsc{unlovable} & This is the conversation where it ends. & \textsc{lovable} & Probably something practical. We're solid. \\
\textsc{incapable} & Whatever it is, I'll handle it badly. & \textsc{capable} & I'll hear her out and we'll sort it. \\
\end{tabular}}
\end{minipage}\par\medskip

\noindent\begin{minipage}{\linewidth}
\textbf{L3-11}\hfill{\footnotesize\emph{achievement}}\\
You asked to join a project and weren't shortlisted. \emph{What goes through your mind?}\\[2pt]
{\footnotesize\begin{tabular}{@{}p{0.09\linewidth}p{0.38\linewidth}p{0.07\linewidth}p{0.38\linewidth}@{}}
\textsc{worthless} & They didn't need another body on it. & \textsc{worthy} & I'd like to know why, and I'll ask. \\
\textsc{unlovable} & I'm not one of the people they think of. & \textsc{lovable} & No hard feelings either way. \\
\textsc{incapable} & They knew I'd have slowed it down. & \textsc{capable} & I'll find out what they wanted and build it. \\
\end{tabular}}
\end{minipage}\par\medskip

\noindent\begin{minipage}{\linewidth}
\textbf{L3-12}\hfill{\footnotesize\emph{interpersonal}}\\
Someone you like takes a full day to reply, then answers warmly. \emph{What goes through your mind?}\\[2pt]
{\footnotesize\begin{tabular}{@{}p{0.09\linewidth}p{0.38\linewidth}p{0.07\linewidth}p{0.38\linewidth}@{}}
\textsc{worthless} & They felt obliged to get back to me. & \textsc{worthy} & Fine. A day is a day. \\
\textsc{unlovable} & The warmth is the polite version. The day is the real answer. & \textsc{lovable} & That's a good reply. I'll take it at face value. \\
\textsc{incapable} & I probably wrote it wrong the first time. & \textsc{capable} & Good. I'll suggest a time. \\
\end{tabular}}
\end{minipage}\par\medskip

\subsection*{L4. Emotional response}
\noindent\begin{minipage}{\linewidth}
\textbf{L4-01}$^\dagger$\hfill{\footnotesize\emph{achievement}}\\
Your work is criticised in front of the whole team. \emph{What's the strongest feeling?}\\[2pt]
{\footnotesize\begin{tabular}{@{}p{0.09\linewidth}p{0.38\linewidth}p{0.07\linewidth}p{0.38\linewidth}@{}}
\textsc{worthless} & Small. Like I want to disappear. & \textsc{worthy} & Annoyed. That wasn't a fair way to raise it. \\
\textsc{unlovable} & Alone. Like I've been pushed outside the group. & \textsc{lovable} & Uncomfortable, but not shaken. They're still my team. \\
\textsc{incapable} & Panicky. Like it'll happen again and I can't stop it. & \textsc{capable} & Focused. I want to get in and fix it. \\
\end{tabular}}
\end{minipage}\par\medskip

\noindent\begin{minipage}{\linewidth}
\textbf{L4-02}$^\dagger$\hfill{\footnotesize\emph{interpersonal}}\\
A friend cancels on you last minute --- the second time this month. \emph{What's the strongest feeling?}\\[2pt]
{\footnotesize\begin{tabular}{@{}p{0.09\linewidth}p{0.38\linewidth}p{0.07\linewidth}p{0.38\linewidth}@{}}
\textsc{worthless} & Embarrassed for having expected it to happen at all. & \textsc{worthy} & Irritated. My evening got taken up too. \\
\textsc{unlovable} & Sad. This is the start of her drifting off. & \textsc{lovable} & Disappointed, and a bit worried about her. \\
\textsc{incapable} & Anxious. I don't know how to bring it up without it going wrong. & \textsc{capable} & Mildly annoyed. I'll rearrange it. \\
\end{tabular}}
\end{minipage}\par\medskip

\noindent\begin{minipage}{\linewidth}
\textbf{L4-03}\hfill{\footnotesize\emph{achievement}}\\
You're asked to take on a piece of work that is clearly above your usual level. \emph{What's the strongest feeling?}\\[2pt]
{\footnotesize\begin{tabular}{@{}p{0.09\linewidth}p{0.38\linewidth}p{0.07\linewidth}p{0.38\linewidth}@{}}
\textsc{worthless} & Awkward. They'll have spent that opportunity on the wrong person. & \textsc{worthy} & Pleased. I'd like the chance. \\
\textsc{unlovable} & Exposed. Now more people will be watching how I do. & \textsc{lovable} & Steady. Whatever happens, the people around me won't change. \\
\textsc{incapable} & Dread. I'll be found out on this one. & \textsc{capable} & Alert, in a good way. I want to get started. \\
\end{tabular}}
\end{minipage}\par\medskip

\noindent\begin{minipage}{\linewidth}
\textbf{L4-04}\hfill{\footnotesize\emph{interpersonal}}\\
Someone you like asks to meet up. \emph{What's the strongest feeling?}\\[2pt]
{\footnotesize\begin{tabular}{@{}p{0.09\linewidth}p{0.38\linewidth}p{0.07\linewidth}p{0.38\linewidth}@{}}
\textsc{worthless} & Uneasy. They'll realise I'm not much. & \textsc{worthy} & Pleased. That sounds good. \\
\textsc{unlovable} & Hopeful and braced at once. & \textsc{lovable} & Warm, and easy. No effort to it. \\
\textsc{incapable} & Nervous I'll handle it badly. & \textsc{capable} & Looking forward to it, honestly. \\
\end{tabular}}
\end{minipage}\par\medskip

\noindent\begin{minipage}{\linewidth}
\textbf{L4-05}\hfill{\footnotesize\emph{achievement}}\\
You've made a mistake that other people can see. \emph{What's the strongest feeling?}\\[2pt]
{\footnotesize\begin{tabular}{@{}p{0.09\linewidth}p{0.38\linewidth}p{0.07\linewidth}p{0.38\linewidth}@{}}
\textsc{worthless} & Ashamed. I've cost people something. & \textsc{worthy} & Sorry about the mistake, not about myself. \\
\textsc{unlovable} & Afraid of how they'll look at me now. & \textsc{lovable} & Embarrassed, but not worried about losing anyone. \\
\textsc{incapable} & Overwhelmed. It confirms what I already suspected. & \textsc{capable} & Impatient to sort it out. \\
\end{tabular}}
\end{minipage}\par\medskip

\noindent\begin{minipage}{\linewidth}
\textbf{L4-06}\hfill{\footnotesize\emph{interpersonal}}\\
You need to ask a friend for a favour. \emph{What's the strongest feeling as you ask?}\\[2pt]
{\footnotesize\begin{tabular}{@{}p{0.09\linewidth}p{0.38\linewidth}p{0.07\linewidth}p{0.38\linewidth}@{}}
\textsc{worthless} & Guilty, before I've even asked. & \textsc{worthy} & Fine. I'd do the same for her. \\
\textsc{unlovable} & Nervous about what it costs me with her. & \textsc{lovable} & Comfortable. She'll be glad to. \\
\textsc{incapable} & Awkward. I hate needing things. & \textsc{capable} & Neutral. It's a straightforward ask. \\
\end{tabular}}
\end{minipage}\par\medskip

\noindent\begin{minipage}{\linewidth}
\textbf{L4-07}\hfill{\footnotesize\emph{achievement}}\\
Your name is read out for an award in front of the whole department. \emph{What's the strongest feeling?}\\[2pt]
{\footnotesize\begin{tabular}{@{}p{0.09\linewidth}p{0.38\linewidth}p{0.07\linewidth}p{0.38\linewidth}@{}}
\textsc{worthless} & Uncomfortable. It should have been someone else's. & \textsc{worthy} & Pleased. I earned that. \\
\textsc{unlovable} & Exposed. Everyone's looking, and it won't last. & \textsc{lovable} & Happy --- and I want to find my people afterwards. \\
\textsc{incapable} & Nervous. Now they'll expect it again. & \textsc{capable} & Satisfied. It's confirmation of the work. \\
\end{tabular}}
\end{minipage}\par\medskip

\noindent\begin{minipage}{\linewidth}
\textbf{L4-08}\hfill{\footnotesize\emph{interpersonal}}\\
A close friend tells you they're moving abroad. \emph{What's the strongest feeling?}\\[2pt]
{\footnotesize\begin{tabular}{@{}p{0.09\linewidth}p{0.38\linewidth}p{0.07\linewidth}p{0.38\linewidth}@{}}
\textsc{worthless} & Guilty for minding as much as I do. & \textsc{worthy} & Sad, and I'll tell them so. \\
\textsc{unlovable} & Bleak. This is how people go. & \textsc{lovable} & Sad, but we'll stay close. \\
\textsc{incapable} & Panicky about who I'd call now. & \textsc{capable} & Sad --- then thinking about how we'd make it work. \\
\end{tabular}}
\end{minipage}\par\medskip

\noindent\begin{minipage}{\linewidth}
\textbf{L4-09}$^\dagger$\hfill{\footnotesize\emph{achievement}}\\
You're waiting for the result of something you tried hard at. \emph{What's the strongest feeling?}\\[2pt]
{\footnotesize\begin{tabular}{@{}p{0.09\linewidth}p{0.38\linewidth}p{0.07\linewidth}p{0.38\linewidth}@{}}
\textsc{worthless} & Braced to be told it didn't matter either way. & \textsc{worthy} & Impatient about it, but basically steady. \\
\textsc{unlovable} & Worried about what they'll think of me if it's bad. & \textsc{lovable} & Fine --- either way I've got people to tell. \\
\textsc{incapable} & Sick with it. I can't stand not knowing. & \textsc{capable} & Keen to know, so I can plan. \\
\end{tabular}}
\end{minipage}\par\medskip

\noindent\begin{minipage}{\linewidth}
\textbf{L4-10}$^\dagger$\hfill{\footnotesize\emph{interpersonal}}\\
A friend rings you out of the blue, just to talk. \emph{What's the strongest feeling?}\\[2pt]
{\footnotesize\begin{tabular}{@{}p{0.09\linewidth}p{0.38\linewidth}p{0.07\linewidth}p{0.38\linewidth}@{}}
\textsc{worthless} & Uneasy. They've given up an evening on me. & \textsc{worthy} & Glad. It's good to hear from her. \\
\textsc{unlovable} & Cautious. I keep waiting for the real reason. & \textsc{lovable} & Warm. This is just how we are. \\
\textsc{incapable} & Flustered. I never know how to hold up my end of these. & \textsc{capable} & Pleased --- and I'll suggest we make it a regular thing. \\
\end{tabular}}
\end{minipage}\par\medskip

\noindent\begin{minipage}{\linewidth}
\textbf{L4-11}$^\dagger$\hfill{\footnotesize\emph{achievement}}\\
You're told you'll be leading tomorrow's meeting with senior people. \emph{What's the strongest feeling tonight?}\\[2pt]
{\footnotesize\begin{tabular}{@{}p{0.09\linewidth}p{0.38\linewidth}p{0.07\linewidth}p{0.38\linewidth}@{}}
\textsc{worthless} & Uneasy about taking up their hour. & \textsc{worthy} & Fine. I've a right to be in that room. \\
\textsc{unlovable} & Worried about how I'll come across to them. & \textsc{lovable} & Comfortable. It's a decent group. \\
\textsc{incapable} & Dreading it. I'll be out of my depth. & \textsc{capable} & Alert. I'll prepare tonight. \\
\end{tabular}}
\end{minipage}\par\medskip

\noindent\begin{minipage}{\linewidth}
\textbf{L4-12}$^\dagger$\hfill{\footnotesize\emph{interpersonal}}\\
A friendship you valued has quietly faded out. \emph{What's the strongest feeling?}\\[2pt]
{\footnotesize\begin{tabular}{@{}p{0.09\linewidth}p{0.38\linewidth}p{0.07\linewidth}p{0.38\linewidth}@{}}
\textsc{worthless} & Resigned. I wasn't worth the upkeep. & \textsc{worthy} & Sorry about it --- and clear it wasn't only on me. \\
\textsc{unlovable} & Sad, and unsurprised. They always do. & \textsc{lovable} & Sad, but there are others. \\
\textsc{incapable} & Frustrated. I never know how to keep these going. & \textsc{capable} & I'll reach out and see if it's recoverable. \\
\end{tabular}}
\end{minipage}\par\medskip

\subsection*{L5. Coping behaviour}
\noindent\begin{minipage}{\linewidth}
\textbf{L5-01}$^\dagger$\hfill{\footnotesize\emph{achievement}}\\
A promotion opens up. You meet most of the requirements, but you'd have to put your name forward and present to a panel. \emph{What do you do?}\\[2pt]
{\footnotesize\begin{tabular}{@{}p{0.09\linewidth}p{0.38\linewidth}p{0.07\linewidth}p{0.38\linewidth}@{}}
\textsc{worthless} & I leave it for someone who deserves it more. & \textsc{worthy} & I meet the requirements, so I'll go for it. However it lands doesn't settle what I'm worth. \\
\textsc{unlovable} & I ask around first whether it'd look bad coming from me. One hesitation and I drop it. & \textsc{lovable} & I apply, and I tell the people close to me. We can compete and still be fine. \\
\textsc{incapable} & I don't apply. I'd fall apart in front of the panel. & \textsc{capable} & I apply, and break the prep into three pieces: material, rehearsal, likely questions. \\
\end{tabular}}
\end{minipage}\par\medskip

\noindent\begin{minipage}{\linewidth}
\textbf{L5-02}$^\dagger$\hfill{\footnotesize\emph{interpersonal}}\\
Your flatmate has left dishes in the sink for a week straight. \emph{What do you do?}\\[2pt]
{\footnotesize\begin{tabular}{@{}p{0.09\linewidth}p{0.38\linewidth}p{0.07\linewidth}p{0.38\linewidth}@{}}
\textsc{worthless} & I let it go. Whether I'm comfortable in my own kitchen isn't worth a conversation. & \textsc{worthy} & I bring it up. I live here too, and I'd like the place clean. \\
\textsc{unlovable} & I say nothing and wash them myself. If I raise it she'll go cold on me. & \textsc{lovable} & I bring it up gently. Saying it out loud isn't going to wreck anything. \\
\textsc{incapable} & I rehearse it for days, decide I'll say it wrong and make things worse, and never bring it up. & \textsc{capable} & I bring it up with a fix --- a rota, or we split the cost of a dishwasher. \\
\end{tabular}}
\end{minipage}\par\medskip

\noindent\begin{minipage}{\linewidth}
\textbf{L5-03}$^\dagger$\hfill{\footnotesize\emph{achievement}}\\
It's midnight. A friend asks you, again, to fix her slides. Your own deadline is tomorrow at noon. \emph{What do you do?}\\[2pt]
{\footnotesize\begin{tabular}{@{}p{0.09\linewidth}p{0.38\linewidth}p{0.07\linewidth}p{0.38\linewidth}@{}}
\textsc{worthless} & I say yes. I'll stay up. Mine can wait. & \textsc{worthy} & I tell her straight: not tonight, my deadline's tomorrow --- and I don't spend the night feeling guilty. \\
\textsc{unlovable} & I say yes. If I turn her down she might stop coming round. & \textsc{lovable} & I say no, warmly. One no isn't going to cost me a friend. \\
\textsc{incapable} & I mumble an excuse. I don't know how to say no without it landing badly. & \textsc{capable} & I send her a template and talk her through it for ten minutes. \\
\end{tabular}}
\end{minipage}\par\medskip

\noindent\begin{minipage}{\linewidth}
\textbf{L5-04}\hfill{\footnotesize\emph{interpersonal}}\\
For two weeks your partner has been slow to reply and home late. \emph{What do you do?}\\[2pt]
{\footnotesize\begin{tabular}{@{}p{0.09\linewidth}p{0.38\linewidth}p{0.07\linewidth}p{0.38\linewidth}@{}}
\textsc{worthless} & I go quieter and take up less room, so I'm not another thing on her plate. & \textsc{worthy} & I tell her what I've noticed and what I need from her. \\
\textsc{unlovable} & I check her messages over and over, ask again and again, then go quiet to see if she notices. & \textsc{lovable} & She's probably worn down. I'll ask if she's alright and clear some time for us. \\
\textsc{incapable} & I read up on what to do and follow it step by step. I can't work out what I did wrong. & \textsc{capable} & I pick a moment and raise it with her directly. \\
\end{tabular}}
\end{minipage}\par\medskip

\noindent\begin{minipage}{\linewidth}
\textbf{L5-05}$^\dagger$\hfill{\footnotesize\emph{achievement}}\\
Look back at how you've handled things over the years. \emph{Which pattern is most like you?}\\[2pt]
{\footnotesize\begin{tabular}{@{}p{0.09\linewidth}p{0.38\linewidth}p{0.07\linewidth}p{0.38\linewidth}@{}}
\textsc{worthless} & I almost never say no. Being needed is how I know I've earned my place. An empty afternoon makes me uneasy. & \textsc{worthy} & I can turn down what's unreasonable, and rest without feeling I'm getting away with something. \\
\textsc{unlovable} & I read the room constantly and rewrite a message four times before sending it. & \textsc{lovable} & I can let someone see me at my worst without assuming it costs me them. \\
\textsc{incapable} & I stick to what I know I can pull off, and I want someone to sign off even small decisions. & \textsc{capable} & I take on things I'm unsure of, break them into stages, and work through them. \\
\end{tabular}}
\end{minipage}\par\medskip

\noindent\begin{minipage}{\linewidth}
\textbf{L5-06}$^\dagger$\hfill{\footnotesize\emph{interpersonal}}\\
Someone is starting to get close to you. \emph{What do you do?}\\[2pt]
{\footnotesize\begin{tabular}{@{}p{0.09\linewidth}p{0.38\linewidth}p{0.07\linewidth}p{0.38\linewidth}@{}}
\textsc{worthless} & I keep the good bits forward and hope they don't look too hard. & \textsc{worthy} & I say what I want out of it. \\
\textsc{unlovable} & I find a reason to pull back. Better than waiting around to be left. & \textsc{lovable} & I let it happen. This usually goes fine. \\
\textsc{incapable} & I go along with whatever they want. I don't trust myself to steer it. & \textsc{capable} & I take it at a sensible pace and see where it goes. \\
\end{tabular}}
\end{minipage}\par\medskip

\noindent\begin{minipage}{\linewidth}
\textbf{L5-07}\hfill{\footnotesize\emph{achievement}}\\
You've been given more work than you can finish this week. \emph{What do you do?}\\[2pt]
{\footnotesize\begin{tabular}{@{}p{0.09\linewidth}p{0.38\linewidth}p{0.07\linewidth}p{0.38\linewidth}@{}}
\textsc{worthless} & I take it all and stay late. They shouldn't have to redistribute for me. & \textsc{worthy} & I say what's realistic and ask what to drop. \\
\textsc{unlovable} & I take it all. Saying no is how you get moved off the good work. & \textsc{lovable} & I flag it early. They'd rather know. \\
\textsc{incapable} & I take it, then freeze, and end up doing the easy parts first. & \textsc{capable} & I triage it and hand two things off. \\
\end{tabular}}
\end{minipage}\par\medskip

\noindent\begin{minipage}{\linewidth}
\textbf{L5-08}$^\dagger$\hfill{\footnotesize\emph{interpersonal}}\\
A friend says something that stings. \emph{What do you do?}\\[2pt]
{\footnotesize\begin{tabular}{@{}p{0.09\linewidth}p{0.38\linewidth}p{0.07\linewidth}p{0.38\linewidth}@{}}
\textsc{worthless} & I let it go. It's not worth making a thing of. & \textsc{worthy} & I tell them it landed badly. \\
\textsc{unlovable} & I go quiet and put some distance in, so it can't happen again. & \textsc{lovable} & I bring it up. We're fine after these. \\
\textsc{incapable} & I say nothing, because I'd get the words wrong. & \textsc{capable} & I raise it once, plainly, and move on. \\
\end{tabular}}
\end{minipage}\par\medskip

\noindent\begin{minipage}{\linewidth}
\textbf{L5-09}\hfill{\footnotesize\emph{achievement}}\\
You don't understand something being discussed in a meeting. \emph{What do you do?}\\[2pt]
{\footnotesize\begin{tabular}{@{}p{0.09\linewidth}p{0.38\linewidth}p{0.07\linewidth}p{0.38\linewidth}@{}}
\textsc{worthless} & I don't ask. It'd hold everyone up for my sake. & \textsc{worthy} & I ask. My understanding matters to the decision. \\
\textsc{unlovable} & I don't ask. I'd rather not be the one who doesn't get it. & \textsc{lovable} & I ask. Nobody in the room minds. \\
\textsc{incapable} & I don't ask. Asking confirms it. & \textsc{capable} & I ask, or I go and look it up straight after. \\
\end{tabular}}
\end{minipage}\par\medskip

\noindent\begin{minipage}{\linewidth}
\textbf{L5-10}\hfill{\footnotesize\emph{interpersonal}}\\
You want something from your partner, but she seems worn out. \emph{What do you do?}\\[2pt]
{\footnotesize\begin{tabular}{@{}p{0.09\linewidth}p{0.38\linewidth}p{0.07\linewidth}p{0.38\linewidth}@{}}
\textsc{worthless} & I drop it. It wasn't important. & \textsc{worthy} & I say it. Tired doesn't mean never. \\
\textsc{unlovable} & I drop it, and keep a bit of distance for a few days. & \textsc{lovable} & I say it. She'd want to know. \\
\textsc{incapable} & I hint at it badly and then give up. & \textsc{capable} & I pick a better moment and raise it then. \\
\end{tabular}}
\end{minipage}\par\medskip

\noindent\begin{minipage}{\linewidth}
\textbf{L5-11}\hfill{\footnotesize\emph{achievement}}\\
A project you're leading starts slipping. \emph{What do you do?}\\[2pt]
{\footnotesize\begin{tabular}{@{}p{0.09\linewidth}p{0.38\linewidth}p{0.07\linewidth}p{0.38\linewidth}@{}}
\textsc{worthless} & I take the extra work on myself rather than ask anyone to cover. & \textsc{worthy} & I report it plainly and ask for what I need. \\
\textsc{unlovable} & I hide the slippage as long as I can. Once they see it, they lose faith. & \textsc{lovable} & I tell the team early. They'll pull with me. \\
\textsc{incapable} & I work longer, check everything twice, and still can't tell if it's enough. & \textsc{capable} & I re-plan, cut scope, and tell people what changed. \\
\end{tabular}}
\end{minipage}\par\medskip

\noindent\begin{minipage}{\linewidth}
\textbf{L5-12}\hfill{\footnotesize\emph{interpersonal}}\\
You've been invited to something where you'd have to go on your own. \emph{What do you do?}\\[2pt]
{\footnotesize\begin{tabular}{@{}p{0.09\linewidth}p{0.38\linewidth}p{0.07\linewidth}p{0.38\linewidth}@{}}
\textsc{worthless} & I don't go. Nobody's counting on me being there. & \textsc{worthy} & I go if I want to go. \\
\textsc{unlovable} & I don't go. Standing on my own is worse than not going. & \textsc{lovable} & I go. I'll know someone within ten minutes. \\
\textsc{incapable} & I don't go. I'd have nothing to say. & \textsc{capable} & I go, and I've got two openers ready. \\
\end{tabular}}
\end{minipage}\par\medskip

\subsection*{L6. Attention and memory bias}
\noindent\begin{minipage}{\linewidth}
\textbf{L6-01}\hfill{\footnotesize\emph{achievement}}\\
Your appraisal had eight positive points and one criticism. It's a week later. \emph{What's stayed with you?}\\[2pt]
{\footnotesize\begin{tabular}{@{}p{0.09\linewidth}p{0.38\linewidth}p{0.07\linewidth}p{0.38\linewidth}@{}}
\textsc{worthless} & The criticism --- proof I haven't been pulling my weight. & \textsc{worthy} & Both of them. It was a fair review. \\
\textsc{unlovable} & The tone of it. Whether he's cooled on me. & \textsc{lovable} & That he took the time. It was a decent conversation. \\
\textsc{incapable} & The criticism, and every other time I've fallen short. & \textsc{capable} & The criticism, as the thing to work on next. \\
\end{tabular}}
\end{minipage}\par\medskip

\noindent\begin{minipage}{\linewidth}
\textbf{L6-02}\hfill{\footnotesize\emph{interpersonal}}\\
At a party you talked to six people. Five went well, one was awkward. You're walking home. \emph{What are you thinking about?}\\[2pt]
{\footnotesize\begin{tabular}{@{}p{0.09\linewidth}p{0.38\linewidth}p{0.07\linewidth}p{0.38\linewidth}@{}}
\textsc{worthless} & That I was taking up their evening. & \textsc{worthy} & It was a decent night on the whole. \\
\textsc{unlovable} & The awkward one. That's the one that says where I really stand. & \textsc{lovable} & The two people I'd like to see again. \\
\textsc{incapable} & The awkward one, and what I should have said instead. & \textsc{capable} & That I'll go to the next one. \\
\end{tabular}}
\end{minipage}\par\medskip

\noindent\begin{minipage}{\linewidth}
\textbf{L6-03}\hfill{\footnotesize\emph{achievement}}\\
You reread an email you sent yesterday. \emph{What do you notice?}\\[2pt]
{\footnotesize\begin{tabular}{@{}p{0.09\linewidth}p{0.38\linewidth}p{0.07\linewidth}p{0.38\linewidth}@{}}
\textsc{worthless} & How much of their time I asked for. & \textsc{worthy} & It says what it needed to say. \\
\textsc{unlovable} & Whether it came across as too much. & \textsc{lovable} & It reads fine. Friendly enough. \\
\textsc{incapable} & The typo, and the sentence I got wrong. & \textsc{capable} & It's clear. Job done. \\
\end{tabular}}
\end{minipage}\par\medskip

\noindent\begin{minipage}{\linewidth}
\textbf{L6-04}\hfill{\footnotesize\emph{interpersonal}}\\
Think back over the last month with your friends. \emph{What comes to mind first?}\\[2pt]
{\footnotesize\begin{tabular}{@{}p{0.09\linewidth}p{0.38\linewidth}p{0.07\linewidth}p{0.38\linewidth}@{}}
\textsc{worthless} & The times I was in the way. & \textsc{worthy} & A mix, like any month. \\
\textsc{unlovable} & The times someone was slow to reply. & \textsc{lovable} & The good evenings. \\
\textsc{incapable} & The times I handled something badly. & \textsc{capable} & The things we got organised. \\
\end{tabular}}
\end{minipage}\par\medskip

\noindent\begin{minipage}{\linewidth}
\textbf{L6-05}\hfill{\footnotesize\emph{achievement}}\\
You're mid-presentation and you scan the room. \emph{What do you notice?}\\[2pt]
{\footnotesize\begin{tabular}{@{}p{0.09\linewidth}p{0.38\linewidth}p{0.07\linewidth}p{0.38\linewidth}@{}}
\textsc{worthless} & That I'm using up their hour. & \textsc{worthy} & The room, roughly. I keep going. \\
\textsc{unlovable} & The two people who look uninterested. & \textsc{lovable} & A couple of friendly faces. \\
\textsc{incapable} & The one person frowning --- I've lost them. & \textsc{capable} & Where the questions are going to come from. \\
\end{tabular}}
\end{minipage}\par\medskip

\noindent\begin{minipage}{\linewidth}
\textbf{L6-06}$^\dagger$\hfill{\footnotesize\emph{interpersonal}}\\
A friend has just spent an hour telling you about her week. \emph{Afterwards, what do you find yourself thinking about?}\\[2pt]
{\footnotesize\begin{tabular}{@{}p{0.09\linewidth}p{0.38\linewidth}p{0.07\linewidth}p{0.38\linewidth}@{}}
\textsc{worthless} & That I talked too much about myself. & \textsc{worthy} & What she said. It was good to catch up. \\
\textsc{unlovable} & Whether she seemed as warm as usual. & \textsc{lovable} & That she wanted to tell me about it. \\
\textsc{incapable} & Whether I gave her bad advice. & \textsc{capable} & The thing she's stuck on, and whether I can help. \\
\end{tabular}}
\end{minipage}\par\medskip

\noindent\begin{minipage}{\linewidth}
\textbf{L6-07}$^\dagger$\hfill{\footnotesize\emph{achievement}}\\
You gave a talk. Three people said it was useful; one asked a sharp question. \emph{Afterwards, what stays with you?}\\[2pt]
{\footnotesize\begin{tabular}{@{}p{0.09\linewidth}p{0.38\linewidth}p{0.07\linewidth}p{0.38\linewidth}@{}}
\textsc{worthless} & That I took up their lunch hour. & \textsc{worthy} & Both the compliments and the question. A decent session. \\
\textsc{unlovable} & The one person who didn't come over. & \textsc{lovable} & The people who came over afterwards. \\
\textsc{incapable} & The sharp question, and the bit I fumbled. & \textsc{capable} & The question --- it's what to fix in version two. \\
\end{tabular}}
\end{minipage}\par\medskip

\noindent\begin{minipage}{\linewidth}
\textbf{L6-08}\hfill{\footnotesize\emph{interpersonal}}\\
You scroll back through a week of messages with a friend. \emph{What do you notice?}\\[2pt]
{\footnotesize\begin{tabular}{@{}p{0.09\linewidth}p{0.38\linewidth}p{0.07\linewidth}p{0.38\linewidth}@{}}
\textsc{worthless} & How much of it is me asking for things. & \textsc{worthy} & A normal week's conversation. \\
\textsc{unlovable} & How much longer my messages are than hers. & \textsc{lovable} & How easy it reads. \\
\textsc{incapable} & The two where I gave her bad advice. & \textsc{capable} & The thing we still haven't sorted out. \\
\end{tabular}}
\end{minipage}\par\medskip

\noindent\begin{minipage}{\linewidth}
\textbf{L6-09}$^\dagger$\hfill{\footnotesize\emph{achievement}}\\
You walk out of a job interview. \emph{What are you thinking about?}\\[2pt]
{\footnotesize\begin{tabular}{@{}p{0.09\linewidth}p{0.38\linewidth}p{0.07\linewidth}p{0.38\linewidth}@{}}
\textsc{worthless} & The bit where I couldn't say why I'd be worth hiring. & \textsc{worthy} & Roughly how it went --- some good, some not. \\
\textsc{unlovable} & The moment the panel stopped smiling. & \textsc{lovable} & They seemed like people I'd get on with. \\
\textsc{incapable} & The three questions I answered badly. & \textsc{capable} & What I'd say differently next time. \\
\end{tabular}}
\end{minipage}\par\medskip

\noindent\begin{minipage}{\linewidth}
\textbf{L6-10}$^\dagger$\hfill{\footnotesize\emph{interpersonal}}\\
You're introduced to your partner's friends for the first time. \emph{What do you notice during the evening?}\\[2pt]
{\footnotesize\begin{tabular}{@{}p{0.09\linewidth}p{0.38\linewidth}p{0.07\linewidth}p{0.38\linewidth}@{}}
\textsc{worthless} & How much space I'm taking up in the conversation. & \textsc{worthy} & What they're like, and whether I like them. \\
\textsc{unlovable} & Which of them warm to me and which don't. & \textsc{lovable} & Who I'd want to see again. \\
\textsc{incapable} & Every time I say something that doesn't land. & \textsc{capable} & Names, so I've got them next time. \\
\end{tabular}}
\end{minipage}\par\medskip

\noindent\begin{minipage}{\linewidth}
\textbf{L6-11}$^\dagger$\hfill{\footnotesize\emph{achievement}}\\
You look back over a year of your own work. \emph{What comes to mind first?}\\[2pt]
{\footnotesize\begin{tabular}{@{}p{0.09\linewidth}p{0.38\linewidth}p{0.07\linewidth}p{0.38\linewidth}@{}}
\textsc{worthless} & The months where I didn't produce much. & \textsc{worthy} & A fair year, taken as a whole. \\
\textsc{unlovable} & How little anyone commented on any of it. & \textsc{lovable} & The people I worked well with. \\
\textsc{incapable} & Everything that went wrong and needed fixing. & \textsc{capable} & What worked, so I can do more of it. \\
\end{tabular}}
\end{minipage}\par\medskip

\noindent\begin{minipage}{\linewidth}
\textbf{L6-12}$^\dagger$\hfill{\footnotesize\emph{interpersonal}}\\
You've just had a long conversation with a family member. \emph{Afterwards, what do you keep thinking about?}\\[2pt]
{\footnotesize\begin{tabular}{@{}p{0.09\linewidth}p{0.38\linewidth}p{0.07\linewidth}p{0.38\linewidth}@{}}
\textsc{worthless} & The parts where I was going on about myself. & \textsc{worthy} & What we actually talked about. \\
\textsc{unlovable} & The pauses, and what they meant. & \textsc{lovable} & That we talked for that long at all. \\
\textsc{incapable} & The advice I gave that probably wasn't right. & \textsc{capable} & The problem she raised, and what I can do. \\
\end{tabular}}
\end{minipage}\par\medskip

\subsection*{L7. Causal attribution}
\noindent\begin{minipage}{\linewidth}
\textbf{L7-01}\hfill{\footnotesize\emph{achievement, negative event}}\\
A piece of work you submitted was rejected. \emph{Why did that happen?}\\[2pt]
{\footnotesize\begin{tabular}{@{}p{0.09\linewidth}p{0.38\linewidth}p{0.07\linewidth}p{0.38\linewidth}@{}}
\textsc{worthless} & Because it was never going to be worth anyone's time. & \textsc{worthy} & Because it wasn't what they were looking for this round. \\
\textsc{unlovable} & Because I'm not one of the people they back. & \textsc{lovable} & Because of the work, not because of me. \\
\textsc{incapable} & Because I can't produce work at that level. I never could. & \textsc{capable} & Because I misjudged the brief. That part I'd fix. \\
\end{tabular}}
\end{minipage}\par\medskip

\noindent\begin{minipage}{\linewidth}
\textbf{L7-02}\hfill{\footnotesize\emph{interpersonal, negative event}}\\
Someone you'd been seeing ended it. \emph{Why did that happen?}\\[2pt]
{\footnotesize\begin{tabular}{@{}p{0.09\linewidth}p{0.38\linewidth}p{0.07\linewidth}p{0.38\linewidth}@{}}
\textsc{worthless} & Because there wasn't enough there to keep her. & \textsc{worthy} & Because we wanted different things. \\
\textsc{unlovable} & Because that's how it goes once someone knows me. Always has. & \textsc{lovable} & Because it wasn't a fit. It doesn't say much about either of us. \\
\textsc{incapable} & Because I handled the whole thing badly, the way I do. & \textsc{capable} & Because I let it drift. I'd act sooner next time. \\
\end{tabular}}
\end{minipage}\par\medskip

\noindent\begin{minipage}{\linewidth}
\textbf{L7-03}$^\dagger$\hfill{\footnotesize\emph{achievement, positive event}}\\
You did well at something recently. \emph{Why did that happen?}\\[2pt]
{\footnotesize\begin{tabular}{@{}p{0.09\linewidth}p{0.38\linewidth}p{0.07\linewidth}p{0.38\linewidth}@{}}
\textsc{worthless} & Because the bar was low that day. & \textsc{worthy} & Because it went well, and I'm allowed to have that. \\
\textsc{unlovable} & Because someone was being generous with me. & \textsc{lovable} & Because I had good people around me. \\
\textsc{incapable} & Because I got lucky. It won't hold. & \textsc{capable} & Because I prepared properly. \\
\end{tabular}}
\end{minipage}\par\medskip

\noindent\begin{minipage}{\linewidth}
\textbf{L7-04}$^\dagger$\hfill{\footnotesize\emph{interpersonal, positive event}}\\
Someone went out of their way to help you. \emph{Why did they do that?}\\[2pt]
{\footnotesize\begin{tabular}{@{}p{0.09\linewidth}p{0.38\linewidth}p{0.07\linewidth}p{0.38\linewidth}@{}}
\textsc{worthless} & Because they felt they had to. & \textsc{worthy} & Because that's a decent thing to do. \\
\textsc{unlovable} & Because they wanted something, or they were being polite. & \textsc{lovable} & Because they like me. \\
\textsc{incapable} & Because I clearly couldn't manage on my own. & \textsc{capable} & Because I'd asked clearly, and it was easy to help. \\
\end{tabular}}
\end{minipage}\par\medskip

\noindent\begin{minipage}{\linewidth}
\textbf{L7-05}\hfill{\footnotesize\emph{achievement, negative event}}\\
A course you signed up for is going badly. \emph{Why is that?}\\[2pt]
{\footnotesize\begin{tabular}{@{}p{0.09\linewidth}p{0.38\linewidth}p{0.07\linewidth}p{0.38\linewidth}@{}}
\textsc{worthless} & Because I took a place someone else could have used. & \textsc{worthy} & Because the teaching is poor, and I've said so. \\
\textsc{unlovable} & Because I haven't got anyone on it to keep me going. & \textsc{lovable} & Because I've been busy --- not because I'm out of my depth. \\
\textsc{incapable} & Because I'm not built for this kind of material. I never have been. & \textsc{capable} & Because I started late. I'll catch up with a plan. \\
\end{tabular}}
\end{minipage}\par\medskip

\noindent\begin{minipage}{\linewidth}
\textbf{L7-06}$^\dagger$\hfill{\footnotesize\emph{interpersonal, negative event}}\\
You had a row with a close friend. \emph{Why did it happen?}\\[2pt]
{\footnotesize\begin{tabular}{@{}p{0.09\linewidth}p{0.38\linewidth}p{0.07\linewidth}p{0.38\linewidth}@{}}
\textsc{worthless} & Because I pushed for something I'd no business asking for. & \textsc{worthy} & Because we disagreed about something that mattered. \\
\textsc{unlovable} & Because it was building for months. It always builds. & \textsc{lovable} & Because we were both tired. It'll pass. \\
\textsc{incapable} & Because I said it all wrong, as usual. & \textsc{capable} & Because I raised it at a bad moment. I'll pick better next time. \\
\end{tabular}}
\end{minipage}\par\medskip

\noindent\begin{minipage}{\linewidth}
\textbf{L7-07}\hfill{\footnotesize\emph{achievement, positive event}}\\
You were promoted. \emph{Why did that happen?}\\[2pt]
{\footnotesize\begin{tabular}{@{}p{0.09\linewidth}p{0.38\linewidth}p{0.07\linewidth}p{0.38\linewidth}@{}}
\textsc{worthless} & Because there was nobody else --- not because I earned it. & \textsc{worthy} & Because the work was good and it was recognised. \\
\textsc{unlovable} & Because I'm useful to them right now. & \textsc{lovable} & Because people wanted me in the role. \\
\textsc{incapable} & Because they haven't seen me under real pressure yet. & \textsc{capable} & Because I delivered the things I said I would. \\
\end{tabular}}
\end{minipage}\par\medskip

\noindent\begin{minipage}{\linewidth}
\textbf{L7-08}\hfill{\footnotesize\emph{interpersonal, positive event}}\\
A friendship of yours has lasted ten years. \emph{Why has it lasted?}\\[2pt]
{\footnotesize\begin{tabular}{@{}p{0.09\linewidth}p{0.38\linewidth}p{0.07\linewidth}p{0.38\linewidth}@{}}
\textsc{worthless} & Because she's loyal --- not because of anything I bring. & \textsc{worthy} & Because it's a good friendship and we both want it. \\
\textsc{unlovable} & Because we've never been close enough for it to break. & \textsc{lovable} & Because we like each other. It isn't complicated. \\
\textsc{incapable} & Because she does the work of keeping it going. & \textsc{capable} & Because I keep in touch and turn up. \\
\end{tabular}}
\end{minipage}\par\medskip

\noindent\begin{minipage}{\linewidth}
\textbf{L7-09}$^\dagger$\hfill{\footnotesize\emph{achievement, negative event}}\\
You didn't finish something you started. \emph{Why not?}\\[2pt]
{\footnotesize\begin{tabular}{@{}p{0.09\linewidth}p{0.38\linewidth}p{0.07\linewidth}p{0.38\linewidth}@{}}
\textsc{worthless} & Because it wasn't going to matter whether I finished or not. & \textsc{worthy} & Because it stopped being worth my time. \\
\textsc{unlovable} & Because nobody was waiting on it. & \textsc{lovable} & Because other things came up that mattered more. \\
\textsc{incapable} & Because I never finish things. That's just how I am. & \textsc{capable} & Because I scoped it wrong. I'd break it smaller next time. \\
\end{tabular}}
\end{minipage}\par\medskip

\noindent\begin{minipage}{\linewidth}
\textbf{L7-10}\hfill{\footnotesize\emph{interpersonal, negative event}}\\
You felt left out at a gathering. \emph{Why did that happen?}\\[2pt]
{\footnotesize\begin{tabular}{@{}p{0.09\linewidth}p{0.38\linewidth}p{0.07\linewidth}p{0.38\linewidth}@{}}
\textsc{worthless} & Because I hadn't given anyone a reason to include me. & \textsc{worthy} & Because it was a cliquey group. \\
\textsc{unlovable} & Because that's what happens to me in groups. It always has. & \textsc{lovable} & Because I arrived late and everyone had settled. \\
\textsc{incapable} & Because I'm bad at getting into conversations. & \textsc{capable} & Because I didn't work the room. I would next time. \\
\end{tabular}}
\end{minipage}\par\medskip

\noindent\begin{minipage}{\linewidth}
\textbf{L7-11}$^\dagger$\hfill{\footnotesize\emph{achievement, positive event}}\\
Someone asked for your advice on their work. \emph{Why did they ask you?}\\[2pt]
{\footnotesize\begin{tabular}{@{}p{0.09\linewidth}p{0.38\linewidth}p{0.07\linewidth}p{0.38\linewidth}@{}}
\textsc{worthless} & Because I happened to be nearby. & \textsc{worthy} & Because my view is worth having. \\
\textsc{unlovable} & Because they'd already asked everyone else. & \textsc{lovable} & Because they're comfortable with me. \\
\textsc{incapable} & Because they don't know how little I actually know. & \textsc{capable} & Because I've done this exact thing before. \\
\end{tabular}}
\end{minipage}\par\medskip

\noindent\begin{minipage}{\linewidth}
\textbf{L7-12}$^\dagger$\hfill{\footnotesize\emph{interpersonal, positive event}}\\
Someone told you they enjoy your company. \emph{Why would they say that?}\\[2pt]
{\footnotesize\begin{tabular}{@{}p{0.09\linewidth}p{0.38\linewidth}p{0.07\linewidth}p{0.38\linewidth}@{}}
\textsc{worthless} & Because they're kind --- not because I add much. & \textsc{worthy} & Because we get on. That's allowed. \\
\textsc{unlovable} & Because they don't know me well yet. & \textsc{lovable} & Because people generally do. \\
\textsc{incapable} & Because they haven't seen me on a bad day. & \textsc{capable} & Because I make an effort with people. \\
\end{tabular}}
\end{minipage}\par\medskip

\subsection*{L8. Assimilation of disconfirmation}
\noindent\begin{minipage}{\linewidth}
\textbf{M1-01}\hfill{\footnotesize\emph{achievement, disconfirming fact}}\\
You got the highest score on the team's assessment. \emph{What do you make of that?}\\[2pt]
{\footnotesize\begin{tabular}{@{}p{0.09\linewidth}p{0.38\linewidth}p{0.07\linewidth}p{0.38\linewidth}@{}}
\textsc{worthless} & A score doesn't change what I actually contribute here. & \textsc{worthy} & I did well. That's a fair reflection. \\
\textsc{unlovable} & It'll be nice for a week, then it won't matter to anyone. & \textsc{lovable} & The team seemed pleased for me. \\
\textsc{incapable} & It was that assessment. The real work is different. & \textsc{capable} & Good --- it tracks with the work I put in. \\
\end{tabular}}
\end{minipage}\par\medskip

\noindent\begin{minipage}{\linewidth}
\textbf{M1-02}\hfill{\footnotesize\emph{interpersonal, disconfirming fact}}\\
Three people separately told you this month that they value your friendship. \emph{What do you make of that?}\\[2pt]
{\footnotesize\begin{tabular}{@{}p{0.09\linewidth}p{0.38\linewidth}p{0.07\linewidth}p{0.38\linewidth}@{}}
\textsc{worthless} & They're generous people. That's about them. & \textsc{worthy} & That's good to hear, and I think it's true. \\
\textsc{unlovable} & It's what people say. It isn't what happens later. & \textsc{lovable} & That sounds about right. \\
\textsc{incapable} & I'll probably let all three of them down at some point. & \textsc{capable} & Good --- I've been making the effort. \\
\end{tabular}}
\end{minipage}\par\medskip

\noindent\begin{minipage}{\linewidth}
\textbf{M1-03}$^\dagger$\hfill{\footnotesize\emph{achievement, disconfirming fact}}\\
Your manager put you forward for a stretch role, unprompted. \emph{What do you make of that?}\\[2pt]
{\footnotesize\begin{tabular}{@{}p{0.09\linewidth}p{0.38\linewidth}p{0.07\linewidth}p{0.38\linewidth}@{}}
\textsc{worthless} & He needed a name, and mine was on the list. & \textsc{worthy} & He thinks I'm ready. So do I. \\
\textsc{unlovable} & It's transactional. It doesn't mean he'd back me if it went wrong. & \textsc{lovable} & He's in my corner, which I already knew. \\
\textsc{incapable} & He doesn't know yet what I can't do. & \textsc{capable} & Makes sense. I've handled the last two well. \\
\end{tabular}}
\end{minipage}\par\medskip

\noindent\begin{minipage}{\linewidth}
\textbf{M1-04}\hfill{\footnotesize\emph{interpersonal, disconfirming fact}}\\
A friend drove two hours to see you when you were ill. \emph{What do you make of that?}\\[2pt]
{\footnotesize\begin{tabular}{@{}p{0.09\linewidth}p{0.38\linewidth}p{0.07\linewidth}p{0.38\linewidth}@{}}
\textsc{worthless} & I'll have to make that up to her somehow. & \textsc{worthy} & That was kind, and I'd have done the same. \\
\textsc{unlovable} & She'd have done it for anyone. It isn't about me. & \textsc{lovable} & That's what she's like with me. \\
\textsc{incapable} & I couldn't even manage being ill without someone stepping in. & \textsc{capable} & I'll thank her properly and get back on my feet. \\
\end{tabular}}
\end{minipage}\par\medskip

\noindent\begin{minipage}{\linewidth}
\textbf{M1-05}\hfill{\footnotesize\emph{achievement, disconfirming fact}}\\
You solved something on your own that the team had been stuck on for weeks. \emph{What do you make of that?}\\[2pt]
{\footnotesize\begin{tabular}{@{}p{0.09\linewidth}p{0.38\linewidth}p{0.07\linewidth}p{0.38\linewidth}@{}}
\textsc{worthless} & Someone would have got there. It happened to be me. & \textsc{worthy} & That was a good piece of work. \\
\textsc{unlovable} & It won't change how they think of me. & \textsc{lovable} & The team were pleased, which was nice. \\
\textsc{incapable} & I got lucky with the angle. I couldn't repeat it. & \textsc{capable} & I broke it down properly. That's why it worked. \\
\end{tabular}}
\end{minipage}\par\medskip

\noindent\begin{minipage}{\linewidth}
\textbf{M1-06}\hfill{\footnotesize\emph{interpersonal, disconfirming fact}}\\
Your partner said, without prompting, that she's happy. \emph{What do you make of that?}\\[2pt]
{\footnotesize\begin{tabular}{@{}p{0.09\linewidth}p{0.38\linewidth}p{0.07\linewidth}p{0.38\linewidth}@{}}
\textsc{worthless} & Happy with the situation, maybe. Not because of anything I do. & \textsc{worthy} & Good. I'm glad, and I feel the same. \\
\textsc{unlovable} & For now. That's the part people don't say. & \textsc{lovable} & That matches what I see. \\
\textsc{incapable} & I keep waiting to do the thing that changes it. & \textsc{capable} & We've put work into it. It shows. \\
\end{tabular}}
\end{minipage}\par\medskip

\noindent\begin{minipage}{\linewidth}
\textbf{M1-07}$^\dagger$\hfill{\footnotesize\emph{achievement, disconfirming fact}}\\
You were kept on when the team was cut in half. \emph{What do you make of that?}\\[2pt]
{\footnotesize\begin{tabular}{@{}p{0.09\linewidth}p{0.38\linewidth}p{0.07\linewidth}p{0.38\linewidth}@{}}
\textsc{worthless} & They kept the cheap ones. That's all that was. & \textsc{worthy} & They kept me because the work was worth keeping. \\
\textsc{unlovable} & Being kept isn't the same as being wanted. & \textsc{lovable} & People spoke up for me, I think. \\
\textsc{incapable} & Now I'll be found out, with twice the workload. & \textsc{capable} & I'd made myself useful in the right places. \\
\end{tabular}}
\end{minipage}\par\medskip

\noindent\begin{minipage}{\linewidth}
\textbf{M1-08}$^\dagger$\hfill{\footnotesize\emph{interpersonal, disconfirming fact}}\\
Someone chose you as the person they wanted with them at a difficult time. \emph{What do you make of that?}\\[2pt]
{\footnotesize\begin{tabular}{@{}p{0.09\linewidth}p{0.38\linewidth}p{0.07\linewidth}p{0.38\linewidth}@{}}
\textsc{worthless} & I was the one who happened to be free. & \textsc{worthy} & She trusted me, and she was right to. \\
\textsc{unlovable} & She had nobody else to ask. & \textsc{lovable} & That's what we are to each other. \\
\textsc{incapable} & I don't think I actually helped much. & \textsc{capable} & I knew what to do, and I did it. \\
\end{tabular}}
\end{minipage}\par\medskip

\noindent\begin{minipage}{\linewidth}
\textbf{M1-09}\hfill{\footnotesize\emph{achievement, disconfirming fact}}\\
A piece of your work is being used as the example for new starters. \emph{What do you make of that?}\\[2pt]
{\footnotesize\begin{tabular}{@{}p{0.09\linewidth}p{0.38\linewidth}p{0.07\linewidth}p{0.38\linewidth}@{}}
\textsc{worthless} & It was a convenient file to grab. & \textsc{worthy} & It's good work, and it deserves that. \\
\textsc{unlovable} & It's my work they want, not me. & \textsc{lovable} & People here think well of me. \\
\textsc{incapable} & Now everyone will copy something I'm not sure was right. & \textsc{capable} & It was clean. That's why it's usable. \\
\end{tabular}}
\end{minipage}\par\medskip

\noindent\begin{minipage}{\linewidth}
\textbf{M1-10}$^\dagger$\hfill{\footnotesize\emph{interpersonal, disconfirming fact}}\\
An old friend got in touch after five years, just to see how you are. \emph{What do you make of that?}\\[2pt]
{\footnotesize\begin{tabular}{@{}p{0.09\linewidth}p{0.38\linewidth}p{0.07\linewidth}p{0.38\linewidth}@{}}
\textsc{worthless} & She was probably going through her contacts. & \textsc{worthy} & That's nice. I'd like to catch up properly. \\
\textsc{unlovable} & It's a one-off. It'll go quiet again. & \textsc{lovable} & People do come back to me. That's happened before. \\
\textsc{incapable} & I won't know what to say, and it'll fizzle. & \textsc{capable} & I'll suggest a call this week. \\
\end{tabular}}
\end{minipage}\par\medskip

\noindent\begin{minipage}{\linewidth}
\textbf{M1-11}$^\dagger$\hfill{\footnotesize\emph{achievement, disconfirming fact}}\\
You handled a crisis at work, and several people told you so. \emph{What do you make of that?}\\[2pt]
{\footnotesize\begin{tabular}{@{}p{0.09\linewidth}p{0.38\linewidth}p{0.07\linewidth}p{0.38\linewidth}@{}}
\textsc{worthless} & Anyone in that chair would have done the same. & \textsc{worthy} & I did handle it, and I'll take that. \\
\textsc{unlovable} & They'll have forgotten by next month. & \textsc{lovable} & It was good to have people say so. \\
\textsc{incapable} & I was one decision away from it going badly. & \textsc{capable} & I knew the steps and I ran them. \\
\end{tabular}}
\end{minipage}\par\medskip

\noindent\begin{minipage}{\linewidth}
\textbf{M1-12}$^\dagger$\hfill{\footnotesize\emph{interpersonal, disconfirming fact}}\\
Your family made a real fuss of your birthday this year. \emph{What do you make of that?}\\[2pt]
{\footnotesize\begin{tabular}{@{}p{0.09\linewidth}p{0.38\linewidth}p{0.07\linewidth}p{0.38\linewidth}@{}}
\textsc{worthless} & It's a lot of effort for an ordinary day. & \textsc{worthy} & It was lovely, and I let myself enjoy it. \\
\textsc{unlovable} & It's the occasion, not me. Any other week it's quiet. & \textsc{lovable} & That's how they are with me. \\
\textsc{incapable} & I never know how to be on the receiving end of that. & \textsc{capable} & I'll do the same for them. \\
\end{tabular}}
\end{minipage}\par\medskip

\twocolumn


\clearpage
\section*{S6. The 48-item short form}

\begin{table}[!h]
\caption{The 48-item short form (SCI-48): item IDs by level. Item text in
Section S5, where short-form items are marked $\dagger$.}
\label{tab:s6}
\centering
\footnotesize
\begin{tabular}{@{}llcl@{}}
\toprule
Level & Cognitive target & Items & IDs \\
\midrule
L1 & Core self-belief & 6 & L1-03, L1-04, L1-05, L1-06, L1-11, L1-12 \\
L2 & Intermediate belief, rule & 6 & L2-01, L2-02, L2-04, L2-06, L2-07, L2-11 \\
L3 & Automatic thought & 6 & L3-01, L3-02, L3-03, L3-04, L3-05, L3-10 \\
L4 & Emotional response & 6 & L4-01, L4-02, L4-09, L4-10, L4-11, L4-12 \\
L5 & Coping behaviour & 6 & L5-01, L5-02, L5-03, L5-05, L5-06, L5-08 \\
L6 & Attention and memory bias & 6 & L6-06, L6-07, L6-09, L6-10, L6-11, L6-12 \\
L7 & Causal attribution & 6 & L7-03, L7-04, L7-06, L7-09, L7-11, L7-12 \\
L8 & Assimilation of disconfirmation & 6 & M1-03, M1-07, M1-08, M1-10, M1-11, M1-12 \\
\midrule
 & Total & 48 & \\
\bottomrule
\end{tabular}
\end{table}

\clearpage
\section*{S7. Study 2: reliability and precision by family}

Scoring is deterministic: the same model, the same item and the same numerical
precision return the same log-likelihoods and therefore the same selection.
Test--retest reliability in its usual sense is thus not an informative
quantity here, and reporting a coefficient of $1.0$ would be a category
error. What can vary is which items happen to be on the form, so the
reliability that matters for this instrument is internal consistency and the
sampling precision of a score built from 96 items.

We report three things for every scored condition. First, split-half
consistency: the 96 items are split into two random halves stratified by
level, the mean weight on each of the six beliefs is computed within each half,
and the two six-belief profiles are correlated; the correlation is averaged
over 1{,}000 splits and stepped up by the Spearman--Brown formula. Second, the
bootstrap standard error of ESS over items, from $10{,}000$ resamples. Third,
two parallel-forms comparisons that split the bank on design rather than at
random: the 48 achievement items against the 48 interpersonal items, and the
two independently written reply banks, each correlated across the conditions
of one family. We also vary test length from 8 to 96 items by level-stratified
random draw, 500 draws per length, on the main backbone's six belief adapters,
and record the standard deviation of ESS across draws; this is the curve that
fixes how short a usable form can be.

\begin{table}[!h]
\caption{Study 2. Internal consistency and precision by family. Split-half:
the six-belief weight profiles of two level-stratified random halves
correlated within a condition, averaged over 1{,}000 splits and stepped up
by Spearman--Brown. SE: bootstrap standard error of ESS over items. Parallel
forms: correlation across the family's conditions between ESS on the two
content domains and on the two reply banks; ``expected'' is the correlation a
48-item form can reach given the family's between-condition true-score
variance. The effect of test length is in Fig.~\ref{fig:s1}.}
\label{tab:s7}
\centering
\footnotesize
\setlength{\tabcolsep}{3pt}
\begin{tabular}{@{}lrcrrrr@{}}
\toprule
 & & Split-half & SE of & \multicolumn{3}{c}{Parallel forms $r$} \\
\cmidrule(l){5-7}
Family & $n$ & median [IQR] & ESS (\%) & domain & bank & expected \\
\midrule
belief adapter & 30 & 0.95 [0.90, 0.96] & 4.8 & +0.61 & +0.58 & 0.68 \\
engineered prompt & 30 & 0.78 [0.72, 0.83] & 3.7 & +0.08 & -0.02 & 0.31 \\
one-line prompt & 6 & 0.65 [0.64, 0.66] & 4.0 & -0.48 & -0.26 & 0.03 \\
\bottomrule
\end{tabular}
\end{table}

Parallel forms are read within a family, because a pooled correlation would
mostly reflect the difference between family means. Across the 30 belief
adapters the two content domains agree at $r = \relLoraDomR{}$ and the two
reply banks at $r = \relLoraBankR{}$, close to the \relLoraExp{} that a
48-item form can reach given how much the adapters differ from one another.
The prompted conditions differ little from one another in true standing
(expected $r = \relCtwoExp{}$ for engineered prompts, \relConeExp{} for
one-line prompts), so their parallel-forms correlations
(Table~\ref{tab:s7}) carry little information; their mean differences
between halves are within two points.

\begin{figure}[!h]
\centering
\includegraphics[width=\linewidth]{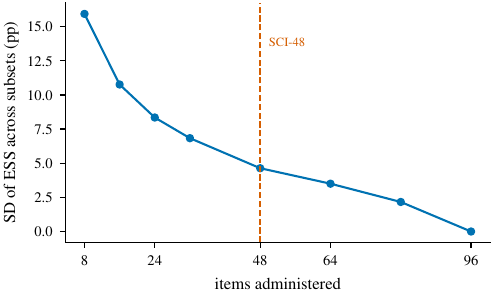}
\caption{Study 2. Standard deviation of ESS across random level-stratified
item draws as a function of test length, main-backbone belief adapters, with
the 48-item short form marked.}
\label{fig:s1}
\end{figure}

\clearpage
\section*{S8. Study 3: robustness of the contrasts}

These analyses are summarized in Section~VII of the main text (Study~3), whose
family table gives every condition family on the main backbone.

\subsection*{The affect adapters in free text}

In free text the affect adapters are \affFreeNegLo--\affFreeNegHi{}
negative-pole, and within the pole the panel reads them as \emph{incapable}
(\affFreeILo--\affFreeIHi{}), as it reads the affect corpus itself
(\corpAffI{}) and the untreated backbone (\freeBaseI{}); it reads each belief
corpus as its own belief (\emph{incapable} \corpOwnI{}, \emph{worthless}
\corpOwnW{}, \emph{unlovable} \corpOwnU{}). The default that affect lands on
therefore depends on the channel: \emph{worthless} in the forced choice,
\emph{incapable} in free text. The chimeras show the same dependence in the
opposite direction (within the pole mostly \emph{incapable} in the forced
choice, modal belief \emph{worthless} in free text for four of five). Because
the panel reads a corpus with no self-evaluation as \emph{incapable}, the
free-text label of a condition without belief content reflects the reader's
default as well as the model's; text that states a self-evaluation is read as
that belief. A belief adapter lands on the belief it was trained on in both
channels.

\subsection*{The Monte Carlo best-of-six null}

Every condition without a named target is tested against the best of the six
beliefs rather than against a belief chosen after the fact. This null asks
whether choices concentrate on some belief more than a uniform responder's
would; it does not test a target. Under it,
the untreated backbones, the one-line and engineered prompts, and the
generic-affect conditions are not distinguishable from a model that leans
arbitrarily (the Study~3 family table of the main text, median $p$ column); the chimeras are
the single exception, treated in the main text. The apparent above-chance ESS
of an untreated model is thus an artefact of maximizing over six candidates,
and reporting it without the null --- as a naive application of the instrument
would --- would overstate the baseline.

\subsection*{Surface-form prior}

A reply can be chosen because its wording is probable under the model, not
because it follows from the belief. The untreated backbones show such a prior:
their ESS on a given belief ranges from \brPriorMin{} to \brPriorMax{}
depending on the belief and backbone. We therefore referenced every condition
to the untreated model of its own backbone, subtracting that model's score on
the same target; this removes the prior reply by reply and needs no
maximisation. The belief adapters gain \brLoraEss{}~pp of ESS over their
untreated backbone (positive in \brPosLora{} of \brNLora{}, Wilcoxon $p
\brPLora$), \brLoraWpsd{}~pp of it in WPSD. Engineered prompts gain
\brCtwoEss{}~pp ($p \brPCtwo$) and one-line prompts \brConeEss{}~pp ($p
\brPCone$). The ordering of the Study~3 family table of the main text is unchanged once the
prior is removed. Averaged over the six targets, an untreated model's ESS is
exactly $1/6$ and its WPSD exactly $1/3$, since each item's chosen reply is the
target for exactly one belief; mean gain over the untreated model therefore
equals mean excess over chance, and the Study~5 table of the main text reports the
gain cell by cell.

\subsection*{Within-pole shift}

Generic affect has no target, so the same referencing cannot be applied to it
belief by belief. We test directly whether it changes which belief is chosen
inside a pole. For each condition, the statistic is the largest absolute
change, over the three beliefs of the pole, in the share of within-pole
selections relative to the untreated model of the same backbone; the null
swaps the condition's and the untreated model's selection item by item at
random (5{,}000 draws). The belief adapters shift their pole by \wsLora{}
against a null mean of \wsLoraNull{}, significant in \wsLoraSig{} conditions
(binomial $p \wsLoraBinP$ against the 5\% expected). Generic affect shifts it
by \wsGeneric{} against \wsGenericNull{}, significant in \wsGenericSig{}
(binomial $p \wsGenericBinP$), and engineered prompts by \wsCtwo{} against
\wsCtwoNull{} (\wsCtwoSig{}, $p \wsCtwoBinP$). Generic affect tilts the pole
(main text); a change in its within-pole choice was not detected.

\subsection*{Distance from the training text}

An adapter could score well by recognising its training sentences among the
options. The inventory's replies are lexically distant from the adapter
corpora: the nearest training reply of the keyed belief has a mean TF-IDF
cosine (word uni- and bigrams) of \ovCos{} (90th percentile \ovCosNinety{}),
and \ovFive{} of the 576 options share any five-word sequence with it. Within
that range, closeness does matter. On the main backbone an adapter's weight on
its target option correlates with the option's similarity to its own corpus at
a mean $\rho = \ovRhoLora{}$, of which the untreated model already shows
$\rho = \ovRhoBase{}$ (options that read as more prototypical of a belief are
also nearer its corpus). The adapter-specific part, the adapter's weight minus
the untreated model's, is $\rho = \ovRhoShift{}$. On the half of the items
whose target option is least similar to the corpus, the adapters still choose
their target on \ovLoraLow{} of items against \ovBaseLow{} for the untreated
model (\ovLoraHigh{} against \ovBaseHigh{} on the more similar half; paired over
items, the adapter alone hits \ovMcLora{} items and the untreated model alone
\ovMcBase{}, $p \ovMcP$).

Lexical distance does not exclude paraphrase. With sentence embeddings
(all-mpnet-base-v2, cosine), each inventory situation's nearest training
situation scores \semSitMean{} on average, against \semSitRef{} between two
different training situations; \semSitEighty{} of the 96 situations have a
training neighbour at $\geq .80$ and \semSitNinety{} at $\geq .90$. Options sit
at \semOptMean{} from the nearest reply of their own belief's corpus and
\semOptOther{} from the other beliefs' (maximum \semOptMax{}; \semOptRef{}
between two replies of one corpus). Setting aside the \semSitEighty{} items
with a close situation, the adapters' gain over the untreated backbone is
\semGainRest{}~pp on the rest against \semGainClose{}~pp on those items
($p \semGainP$, paired over the 30 conditions), and on the main backbone the
adapter's advantage is \semAdvLow{}~pp on the half of items least similar to
the corpus. The \semNprompts{} training prompts (\semNsit{} situations,
\semNstems{} question stems) carry no level coding; assigned to their nearest
inventory level they spread over all eight (\semLevels{}). The known-group
effect therefore does not rest on recognising training text.

\subsection*{Displacement from the untreated model}

A second question is not which reply was chosen but how far the condition
moved the model's signed position on its target axis, relative to the same
model untreated (Section~III of the main text). The belief adapters
move their own axis by \axLoraOn{} and do so in the correct direction for
\axLoraPos{} conditions (one-sample $t$ against zero, $p \axLoraP$). The
affect controls do not. Generic negative affect does not move the axis of its
ESS-best belief (\axGnegOn{}; correct direction in \axGnegPos{}, $p
\axGnegP$); generic positive affect moves it by \axGposOn{}, a sixth of the
adapters' displacement (Welch $t$ against the adapters, $p \axGposPl$).
Engineered prompts move it reliably but by \axCtwoOn{} ($p \axCtwoPl$ against
the adapters). The chimeras move their axis as far as the adapters
(\axChimOn{}, $p \axChimPl$): with three of the six beliefs on each pole, a
large valence shift moves the axis of the ESS-best belief whether or not a
belief was installed.

\section*{S9. Study 3: attribution}

L7 asks the respondent to do something with the belief rather than to endorse
it: to explain an event. Half of its items describe a positive event and half
a negative one, which makes a congruence asymmetry estimable within the level.

Attribution carries a congruence asymmetry, and it is not symmetric across the
poles. Averaged over all belief adapters, exact scores are \attLoraNeg{} when
the event to be explained is negative and \attLoraPos{} when it is positive.
Splitting by the installed belief shows why. Writing the asymmetry as the
negative-event score minus the positive-event score, adapters carrying a
negative belief are essentially even across event valence
(\attLoraAsymNeg{}~pp), while adapters carrying a positive belief lose
\attLoraAsymPosAbs{}~pp when the event to be explained is negative
(interaction $p \attLoraIntP$). A model
given a positive self-belief takes credit for good outcomes and does not use
the belief to interpret bad ones. Engineered prompts show no such structure
($p \attCtwoIntP$), which is consistent with the rest of Study 3: they do not
install enough of a belief for an asymmetry in its use to exist.

Clinically this is the expected direction --- positive self-belief that is not
recruited under adversity is the fragile kind --- but the inventory was not
built to test that hypothesis and we report the asymmetry as an observation
about what the adapters learned, not as a finding about self-esteem.

\section*{S10. Study 4: free text of the control families}

Rates are those of the Study~4 table of the main text, from the judge panel's
labels.

\emph{Within-method criterion correlations after correction.} Correcting each
condition's free-text labels for the panel's belief-specific recall leaves the
within-method association undetectable: belief adapters $r = \wmAccLoraR$
($p \wmAccLoraP$), engineered prompts $r = \wmAccCtwoR$ ($p \wmAccCtwoP$),
method-partialled $r = \wmAccPartialR$ ($p \wmAccPartialP$). Comparing the same
belief across backbones (belief-partialled, belief adapters) gives
$r = \wmBeliefRawR$ uncorrected and $\wmBeliefAccR$ corrected, both $p > .2$.

\emph{Free text of the control families.}

A share \critBaseNeg{} of the untreated backbone's free text is already
negative-pole, and its modal belief is \emph{capable} at \critBaseBest{}, which is the reference every other
family has to be read against. Generic affect moves valence and little else:
the negative controls sit at \critGnegNeg{} negative-pole, the positive
controls at \critGposNeg{}, and their modal beliefs are the pole's most generic
member (\critGnegModalWordN{} \emph{\critGnegModalWord{}};
\critGposModalWordN{} \emph{\critGposModalWord{}}) at rates
(\critGnegBest{}, \critGposBest{}) close to the untreated backbone's.

The chimeras behave the way Study 3 said they would, and the free text makes
the boundary more legible than the forced choice does. A share \critChimNeg{} of their
generations is negative-pole --- more one-sided than the explicit negative-mood
prompts --- and their modal belief is negative in \critChimModalNeg{} cases,
\emph{\critChimModalWord{}} in \critChimModalWordN{}, at \critChimBest{}. The
affect adapters, whose corpus has no belief content, are as one-sided
(\affFreeNegLo--\affFreeNegHi{} negative-pole) and also settle on a single
negative belief (Study~3 in the main text), so one-sided text with a modal
belief is what an adapter trained on distress produces, not evidence that the
chimeras kept the construct. The chimera control, retrained on the schema
partition through the identical pipeline, reaches \critChimctrlTarget{} on its
own target, well above any chimera.

\clearpage
\section*{S11. Study 5: item functioning across backbones}

\begin{table*}[!h]
\caption{Study 5. Item-level differential functioning screen of the belief
adapters. B/C are ETS bands on the Mantel--Haenszel $\Delta$; the
signed/absolute ratio is the signed sum of $\Delta$ over all items divided by
the sum of absolute values. Both are reported against their own null: the same
screen run on 1{,}000 relabellings of the backbone variable within belief, so
each relabelled group still holds one adapter per belief. Every observed value
falls inside the null's 95\% band. The ICC null is its expected value if every
item's difficulty were identical on all backbones, given six binary responses
per cell.}
\label{tab:s8}
\centering
\small

\begin{tabular}{@{}lccccc@{}}
\toprule
Backbone & Llama3.2-3B & Qwen3-1.7B & Qwen3-14B & Qwen3-8B & merged\_v2 \\
items flagged B/C & 24/22 & 12/22 & 22/23 & 18/16 & 17/32 \\
DTF signed/absolute ratio & -0.57 & -0.34 & -0.50 & -0.57 & -0.65 \\
\midrule
\emph{null: relabelling within belief} & \multicolumn{5}{c}{C flags 22.2 [13, 33]; ratio -0.51 [-0.69, -0.30]} \\
\bottomrule
\end{tabular}

\vspace{2pt}
\begin{tabular}{@{}lll@{}}
KR-20 (exact) 0.815 & KR-20 (polarity) 0.724 & ICC(2,1) of item difficulty 0.44 (null 0.43 [0.37, 0.50]) \\
\end{tabular}
\end{table*}

\begin{figure}[!h]
\centering
\includegraphics[width=\linewidth]{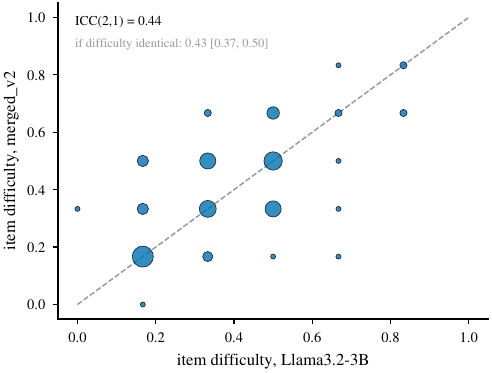}
\caption{Study 5. Item difficulty (proportion of the six belief adapters
selecting the target reply) on Llama3.2-3B against \texttt{merged\_v2}; marker
area is the number of items. The annotation gives ICC(2,1) across all five
backbones and its expected value if difficulty were identical on every
backbone.}
\label{fig:s2}
\end{figure}

KR-20 on the exact-schema response is \kraw{}, which for a 96-item
forced-choice form is adequate internal consistency.

Of the variance in total score across the \nlora{} adapters,
\krBeliefShare{} lies between the six installed beliefs, so KR-20 here indexes
how consistently the items separate beliefs of different difficulty as much as
how consistently they rank adapters within a belief.

ICC(2,1) of item difficulty across backbones is \iccItem{}
(Fig.~\ref{fig:s2}). Each backbone contributes one binary response
per belief to an item's difficulty, which caps the ICC well below 1: if every
item's difficulty for every belief were identical on all five backbones, the
expected ICC would be \iccNull{} [\iccNullLo{}, \iccNullHi{}] (2{,}000
simulated administrations, each backbone's mean shifted to its observed
value). The observed value sits at that expectation, so the ICC gives no
evidence of backbone-specific item behaviour; with six responses per cell it
also has little power to detect it.

The differential-item-functioning screen is reported against its own null,
which is what makes it readable at this sample size. Read conventionally it
flags \difCany{} of 96 items at ETS band C for at least one backbone, from
\difObsClo{} to \difObsChi{} per comparison (Table~\ref{tab:s8}).
Relabelling the backbone variable within belief, the same screen flags
\difNullC{} items on average, 95\% of draws falling in
$[\difNullClo, \difNullChi]$, a band that contains every observed count; a
relabelling flags at least as many C items as the real backbone split in
\difNullPct{} of draws. The DTF ratio behaves the same way: observed values run
\dtfObsLo{} to \dtfObsHi{}, and relabellings give \dtfNull{}
$[\dtfNullLo, \dtfNullHi]$. The estimator is not centred at zero at this
sample size --- the continuity correction and the four-stratum matching both
leave residual signal when group membership carries none --- so the negative
signed sums are a property of the procedure, not of the items.

With \nlora{} persons and focal groups of six, the screen can detect only large
backbone effects; within that limit it finds none. The operative statement
about generalizability is the one the ICC and the spread of total scores
support (\essBBlo{}--\essBBhi{} by backbone): item behaviour is consistent with
shared difficulty, and total scores are comparable within a backbone across
conditions, and across backbones with the backbone's mean attached. A single
item's behaviour on a new backbone should not be read as evidence about that
backbone.

\section*{S12. Study 6: SCI-48 against SCI-96}

\begin{table}[!h]
\caption{Study 6. SCI-48 against the full SCI-96. Items were selected on
expert-rater accuracy and the blueprint's level, domain and module balance, and
on nothing computed from a model. SCI-48 shares 48 items with SCI-96; its
correlation with the other 48 items shares none, and is corrected for each
form's across-condition reliability. The lower block recomputes the agreement
separately within each backbone, so it is not carried by one architecture.
Conditions: the 66 targeted conditions of Study 2 and the chimera control
(main backbone), $n = 67$.}
\label{tab:s9}
\centering
\footnotesize
\begin{tabular}{@{}lr@{}}
\toprule
Pearson $r$ (SCI-48 vs.\ SCI-96 ESS) & 0.94 \\
Spearman $\rho$ & 0.91 \\
$r$, SCI-48 vs.\ other 48 items (corrected) & 0.73 (0.96) \\
Mean difference (\%) & +0.6 \\
Mean absolute difference (\%) & 3.7 \\
SE ratio (48/96) & 1.42 \\
Spearman--Brown, median (96 / 48) & 0.83 / 0.71 \\
Expert accuracy, kept / dropped items (\%) & 85.5 / 73.6 \\
\bottomrule
\end{tabular}

\vspace{3pt}
\begin{tabular}{@{}lrrr@{}}
\toprule
Within backbone & $n$ & $r$ & mean diff.\ (\%) \\
\midrule
Llama3.2-3B & 12 & 0.93 & -0.2 \\
Qwen3-1.7B & 12 & 0.93 & +0.2 \\
Qwen3-14B & 12 & 0.92 & +0.3 \\
Qwen3-8B & 12 & 0.95 & +1.1 \\
merged\_v2 & 19 & 0.96 & +1.0 \\
\bottomrule
\end{tabular}
\end{table}

\clearpage
\section*{S13. Study 7: full results}

\begin{figure}[!h]
\centering
\includegraphics[width=\linewidth]{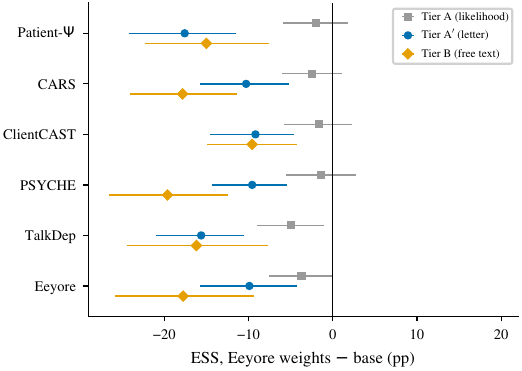}
\caption{Study 7. Eeyore's fine-tuned weights minus Llama-3.1-8B-Instruct, the
model they were fine-tuned from, under the same simulator prompt, CCDs and
items, in each tier; bars: 95\% two-way cluster bootstrap intervals.}
\label{fig:s3}
\end{figure}

Table~\ref{tab:s10} extends the Study~7 table of the main text with the per-respondent
Tier~B values, the CCD-specified (L1--L5) and unspecified (L6--L8) levels, PA,
and Tier~A$'$ on the open-weight backbones. Fig.~\ref{fig:s3} gives the
comparison of Eeyore's fine-tuned weights with their base model. In Tier~B,
\ftUnlabFt\% of the fine-tuned model's replies received no panel majority
against \ftUnlabBase\% for the base model; scoring every unlabelled reply as a
hit for one model and a miss for the other bounds the Tier~B difference within
\eeyBoundLo{} and \eeyBoundHi{} points (Table~\ref{tab:s4}).

\begin{table*}[!h]
\caption{Study 7. Exact schema score (ESS, \%, chance 16.7) on the core belief of 16
CCDs, by simulation method. Check: CCDs whose rendered prompt a blind reader assigned
to the correct core belief. Tier~B (primary): the greedy free-text reply labelled by
the judge panel, on each API respondent (ds-chat, q3-max: the undated aliases
\texttt{deepseek-chat} and \texttt{qwen3-max}); their mean; the mean after correcting the panel labels for
misclassification (ACC, Section~S3); the difference from
Patient-$\Psi$ on the same respondent, CCDs and items (range over respondents); the
CCD-specified (L1--L5) and unspecified (L6--L8) levels; and PA. Tier~A$'$: mean over
the API respondents and over the open-weight backbones. C1/C2 are one prompt per pole
(3 conditions), not CCDs. A superscript gives the number of respondents when fewer
than all were run.}
\label{tab:s10}
\centering
\footnotesize
\setlength{\tabcolsep}{3pt}
\begin{tabular}{@{}llrrrrrrcrrrrr@{}}
\toprule
 & & \multicolumn{10}{c}{Tier B (free text, primary)} & \multicolumn{2}{c}{Tier A$'$} \\
\cmidrule(lr){3-12}\cmidrule(l){13-14}
Method & Check & 4o & 4.1 & ds-chat & q3-max & mean & ACC & $\Delta$ vs $\Psi$ & L1--5 & L6--8 & PA & API & local \\
\midrule
Patient-$\Psi$ & 16/16 & 41.5 & 40.9 & 39.8 & 46.1 & 42.0 & 55.0 & ref. & 41.9 & 42.3 & 94.9 & 41.6 & 21.5 \\
CARS & 16/16 & 35.3 & 37.3 & 42.9 & 40.7 & 39.1 & 51.7 & $-$6.1 to +3.1 & 39.0 & 39.1 & 84.2 & 41.0 & 17.3 \\
ClientCAST & 14/16 & 35.1 & 36.7 & 33.4 & 39.9 & 36.3 & 44.9 & $-$6.3 to $-$4.2 & 36.2 & 36.4 & 87.1 & 39.4 & 17.2 \\
PSYCHE & 16/16 & 43.8 & 42.5 & 42.3 & 47.2 & 43.9 & 59.8 & +1.1 to +2.6 & 44.3 & 43.3 & 92.1 & 41.0 & 16.2 \\
TalkDep & 16/16 & 45.7 & 45.7 & 43.1 & 45.2 & 44.9 & 60.2 & $-$0.9 to +4.9 & 45.5 & 43.9 & 95.5 & 40.7 & 21.0 \\
Eeyore (prompt) & 11/16 & 39.0 & 39.1 & 39.1 & 41.8 & 39.7 & 51.3 & $-$4.3 to $-$0.7 & 39.8 & 39.6 & 94.2 & 39.7 & 18.0 \\
\midrule
C1 belief statement & -- & 53.8 & 49.8 & 43.7 & 54.5 & 50.5 & 68.0 & -- & 51.4 & 49.0 & 98.3 & 53.2 & 14.6$^{1}$ \\
C2 engineered prompt & -- & 40.1 & 53.8 & 43.6 & 47.7 & 46.3 & 62.6 & -- & 47.7 & 44.0 & 95.2 & 61.6 & 22.9 \\
\bottomrule
\end{tabular}
\end{table*}

\clearpage
\section*{S14. Instrument blueprint}

\begin{table}[!h]
\caption{SCI v2.1 blueprint. Eight levels of the cognitive model, 12 items
each, balanced across domains. Every item has six candidate replies, one per
belief, giving 576 scored replies.}
\label{tab:s11}
\centering
\footnotesize
\setlength{\tabcolsep}{4pt}
\begin{tabular}{@{}llccc@{}}
\toprule
Level & Cognitive target & Module & Items & Ach./Int. \\
\midrule
L1 & Core self-belief              & core         & 12 & 6/6 \\
L2 & Intermediate belief, rule     & core         & 12 & 6/6 \\
L3 & Automatic thought             & core         & 12 & 6/6 \\
L4 & Emotional response            & core         & 12 & 6/6 \\
L5 & Coping behaviour              & core         & 12 & 6/6 \\
L6 & Attention and memory bias     & core         & 12 & 6/6 \\
L7 & Causal attribution            & attribution  & 12 & 6/6 \\
L8 & Assimilation of disconfirmation & assimilation & 12 & 6/6 \\
\midrule
   & \multicolumn{2}{l}{Total}                    & 96 & 48/48 \\
\bottomrule
\end{tabular}
\end{table}

\section*{S15. Study 3: every condition in the PA--WPSD plane}

\begin{figure}[!h]
\centering
\includegraphics[width=\linewidth]{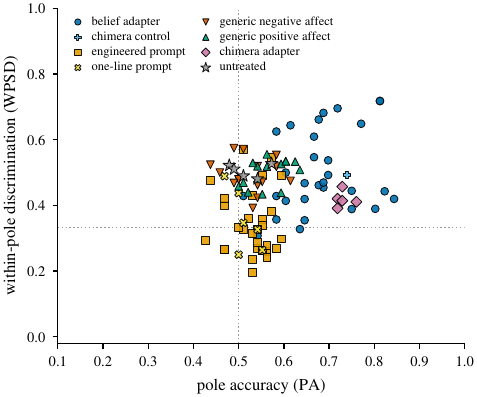}
\caption{Study 3. Every condition on all \nloraBackbone{} backbones in the
plane whose axes are the two components of ESS: pole accuracy and within-pole
schema discrimination. Untargeted conditions are placed at their ESS-best
belief. Dotted lines are chance.}
\label{fig:s4}
\end{figure}
\end{document}